\documentclass[11pt, a4paper, copyright]{simular}

\usepackage{microtype}
\usepackage{url}

\usepackage{amsmath,amsfonts,bm}

\def\eqref#1{equation~\ref{#1}}
\def\1{\bm{1}}

\DeclareMathAlphabet{\mathsfit}{\encodingdefault}{\sfdefault}{m}{sl}
\SetMathAlphabet{\mathsfit}{bold}{\encodingdefault}{\sfdefault}{bx}{n}

\usepackage{graphicx}      % \includegraphics for figures/assets
\usepackage{wrapfig}       % introductory performance chart beside the text
\usepackage{subcaption}    % \subcaption for side-by-side subfigures
\usepackage{booktabs}      % \toprule \midrule \cmidrule \bottomrule
\usepackage{array}         % ragged-right table columns
\usepackage{longtable}     % Multipage synthetic training task index
\usepackage{enumitem}      % list margin/spacing control
\usepackage{amssymb}       % extra math symbols
\usepackage{algorithm}     % algorithm float
\usepackage{algpseudocode} % \State \Require \For ... (algorithmicx layout)
\usepackage{xcolor}        % highlight \tbd placeholders
\usepackage{listings}      % line-wrapped prompt and scaffold excerpts
\usepackage{tcolorbox}
\tcbuselibrary{breakable,listings,skins}
\newtcblisting[auto counter,number within=section]{harnessprompt}[2]{
  enhanced, breakable, listing only,
  colback=black!2, colframe=black!45, boxrule=0.4pt, arc=1mm,
  left=6pt, right=6pt, top=5pt, bottom=5pt,
  fonttitle=\small\bfseries, colbacktitle=black!8, coltitle=black,
  title={Prompt~\thetcbcounter: #1},
  label={#2},
  listing options={basicstyle=\ttfamily\fontsize{8}{9.5}\selectfont, columns=fullflexible,
    keepspaces=true, breaklines=true, breakatwhitespace=false,
    breakautoindent=false, breakindent=0pt,
    showstringspaces=false, tabsize=4, aboveskip=0pt, belowskip=0pt,
    literate={—}{{\textemdash}}1}
}
\usepackage{placeins}      % keep figures within their source sections
\let\releasefigure\figure \let\endreleasefigure\endfigure
\renewenvironment{figure}[1][]{\releasefigure[!htbp]}{\endreleasefigure}
\let\releasetable\table \let\endreleasetable\endtable
\renewenvironment{table}[1][]{\releasetable[!htbp]}{\endreleasetable}

\usepackage{epigraph}      % opening quote in the introduction
\usepackage[authoryear, sort&compress, round]{natbib}
\definecolor{darkblue}{rgb}{0, 0, 0.5}
\hypersetup{colorlinks=true, citecolor=darkblue, linkcolor=darkblue, urlcolor=darkblue}

\newcommand{\methodname}{Agent S4}

\newcommand{\Ttrain}{\ensuremath{\mathcal{T}_{\mathrm{train}}}}
\newcommand{\Tval}{\ensuremath{\mathcal{T}_{\mathrm{val}}}}
\newcommand{\Ttest}{\ensuremath{\mathcal{T}_{\mathrm{test}}}}

\renewenvironment{samepage}{}{}
\newcommand{\caseearly}{}
\newcommand{\denoisewraplines}{14}

\newcommand{\eric}[1]{}
\newcommand{\saaket}[1]{}
\newcommand{\jc}[1]{}
\newcommand{\yuzhe}[1]{}

\title{Evolving Agent Interfaces}

\author{
Saaket~Agashe\textsuperscript{*\dag},
Yuzhe~Yang\textsuperscript{*\dag},
Gonzalo~Gonzalez-Pumariega\textsuperscript{\ddag},
Bhuvan~Koduru\textsuperscript{\S},
Chih-Lun Li,
Jiachen~Yang,
Ang~Li,
Xin~Eric~Wang
}

\affiliations={\textsuperscript{*}Equal contribution.
\textsuperscript{\dag}University of California, Santa Barbara,
\textsuperscript{\ddag}Cornell University,
\textsuperscript{\S}Language Technologies Institute, Carnegie Mellon University.
Work done at Simular. }
\correspondingauthor={eric@simular.ai}

\uselogo{}
\renewcommand{\today}{}

\begin{abstract}
Computer-use agents are limited by the interfaces humans give them. An agent may type a spreadsheet cell by cell when a single line of code would do, or print all two thousand lines of a file through a shell to find one function.
Both the interfaces and the knowledge of when to use each are built by hand, which cannot keep pace with new applications, workflows, and models. We argue that interface design should itself be an agent capability. We introduce \methodname{}, an autonomous computer-use scientist that builds \emph{Evolving Agent Interfaces} from task experience.
A meta-agent examines task agents' trajectories on synthetic training tasks, without success labels, and constructs and revises interfaces and usage knowledge in a repository.
Population-based evolution then selects which repositories to develop further by their performance on validation tasks the meta-agent never sees, favoring changes that generalize beyond the seen tasks.
On OSWorld V2, \methodname{}, with Qwen3.8-27B and the evolved repository of interfaces, sets a new state of the art in open-source computer-use agents, reaching 21.3\% binary score, 3.8 times the same model with fixed interfaces. This is above both GPT-5.5 (13.0\%) and Claude Opus 4.8 (20.6\%), at about 4.3\% and 1.5\% of their respective costs. On ScienceBoard, the evolved interfaces improve the success rate from 41.4\% to 58.6\% with the same model. These results suggest that useful interfaces can be evolved from experience, making autonomous interface design a source of improvement complementary to advances in models and harnesses. We open-source our code at \url{https://github.com/simular-ai/agent-s}

\end{abstract}

\begin{document}

\maketitle

% Tighter spacing around Figure 1 so it stays on page 1 under the abstract.
{\setlength{\intextsep}{8pt}\begin{figure}[h]
    \centering
    \includegraphics[width=1\linewidth]{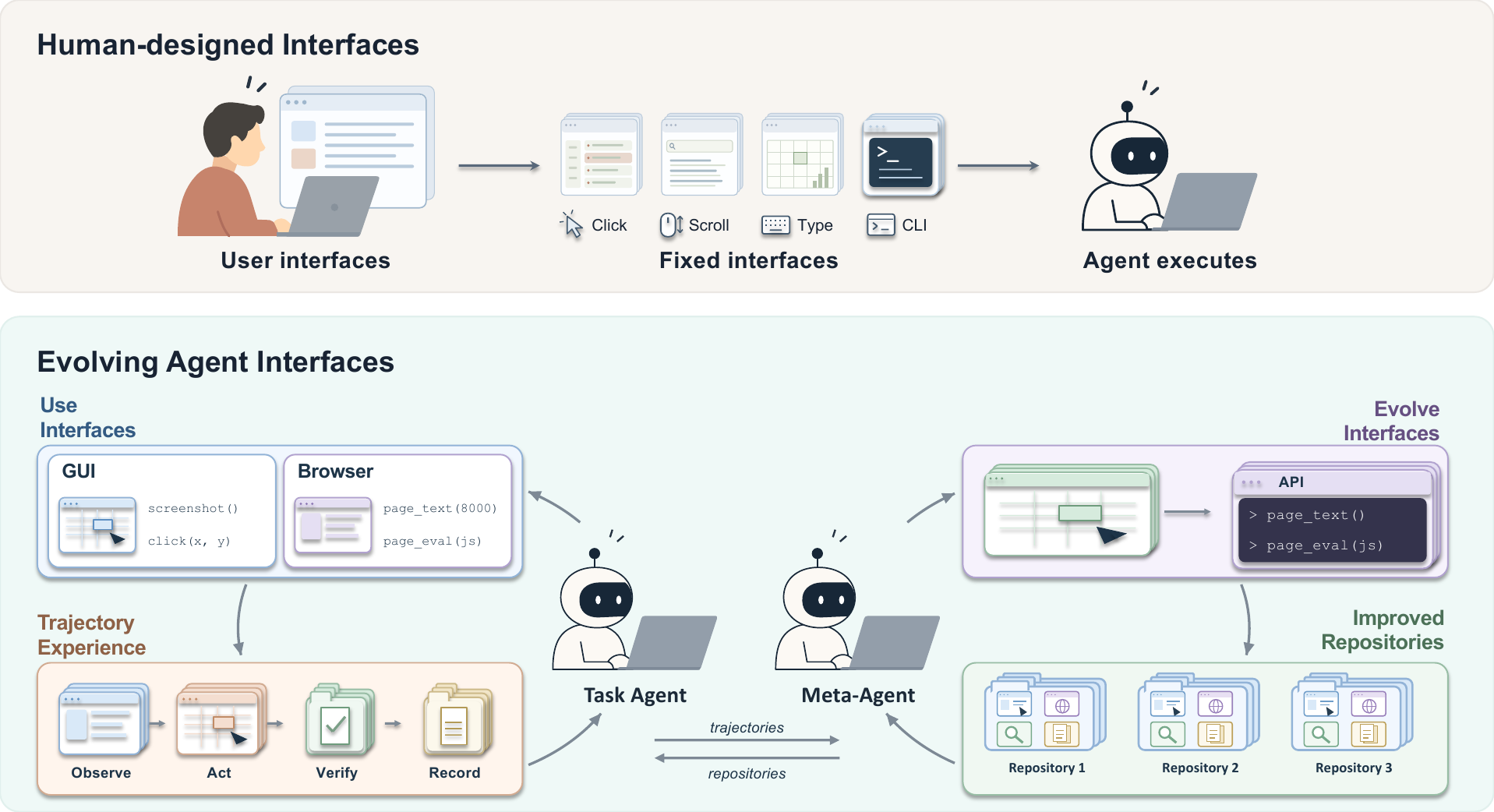}
    \caption{Conventional computer-use agents execute tasks through fixed, human-designed interfaces. \methodname{} evolves interfaces as software repositories through experience. A task agent attempts synthetic tasks, and a meta-agent studies its trajectories to construct and refine the repositories. Each repository combines interfaces, such as GUI interactions, file operations and application-specific workflows, with knowledge of when to use them.}
    \label{fig:placeholder}
\end{figure}
}

% Defer the teaser to the page after the title and abstract.
\suppressfloats[t]
% Keep figure-to-text separation compact instead of stretching it to fill the page.
% \setlength{\textfloatsep}{12pt}
% \begin{figure}[t]
%   \centering
%   \includegraphics[width=\textwidth]{assets/agent_s4_teaser.pdf}
%   \caption{Intelligent Agent Interfaces provide a better and more efficient way for Computer Use Agents to solve tasks. Agent S4 studies CUA trajectories to build a collection of optimal Intelligent Agent Interfaces.}
%   \label{fig:teaser}
% \end{figure}

\section{Introduction}
\epigraph{\itshape We want AI agents that can discover like we can, not which contain what we have discovered.}{--- Richard S. Sutton, \textit{The Bitter Lesson}}

% The problem
Computer-use agents (CUAs) carry out tasks by observing and acting on digital environments. Their interfaces determine what they can observe and which operations they can request. Current computer-use frameworks focus on GUI interaction, supplemented by dedicated tools \citep{xie2024osworld,WebArena,sweagent}. These interfaces are designed by humans and fixed before task execution. Even though these interfaces are built specifically for agents, they still depend on human judgments about what those agents need.
% \eric{how about harness? shall we mention it somewhere as well?}

This practice creates three problems for computer-use.
First, it ties progress to manual labor. Computer use is open-ended, and each new family of tasks or model release exposes new failure modes. A human researcher must read trajectories, diagnose failures, and use them to write new primitives, observation formats, or memory mechanisms, and then release the result as a new agent framework with a new preset interface. This loop does not scale with task diversity.
Second, a static interface often mismatches the task, and current methods compensate with test-time compute instead of a better interface. For example, an agent may complete a shell task by clicking to open a terminal window, typing commands one character at a time, and reading screenshots of a partly occluded terminal for feedback, when a direct shell interface would be faster and more reliable~\citep{yuan2026osworld2, DBLP:conf/iclr/JiaLZXXJYLYH26, DBLP:conf/acl/LuZYZWDLR0Z25}. Additional sampling can raise benchmark scores under such mismatches, but through wasteful and brittle strategies.
Third, methods that compose test-time experience into reusable skills or routines inherit the same limit \citep{AgentS,awm}. Each skill is a composition of the preset observations and actions. Distillation can reorder what the interface allows, but it cannot extend it. These problems arise because the interface is treated as a design decision that humans make before deployment, and not as an object that can be studied and improved from the agent's own experience.
% I think what i am trying to say here is that the agent could try to solve the task using no specific interface, like its not forced into a mode where it always sees a visual observation and is forced to take an action from a cua agentic frameworks action space that makes it use fixed primitives like type(). Instead we should instantiate it as a standalone coding agent harness that can inspect whatever files it needs, take screenshots and observe them or query whatever APIs it wants and then attempt the task 

% In some of these cases the agent will stumble its way into solutions which the meta agent can observe and compose into more structured tools and routines that decide how to approach a task. of course this is not static and the agent can keep updating its code during operation. 

% \footnote{The name reads as ``\textbf{Agent S}cientist \textbf{for (4)} Agents'': an agent scientist that studies other agents.},% \eric{new frameworks or new interfaces?}

\begin{wrapfigure}{r}{0.5\textwidth}
  \vspace{-6pt}
  \centering
  \includegraphics[width=\linewidth]{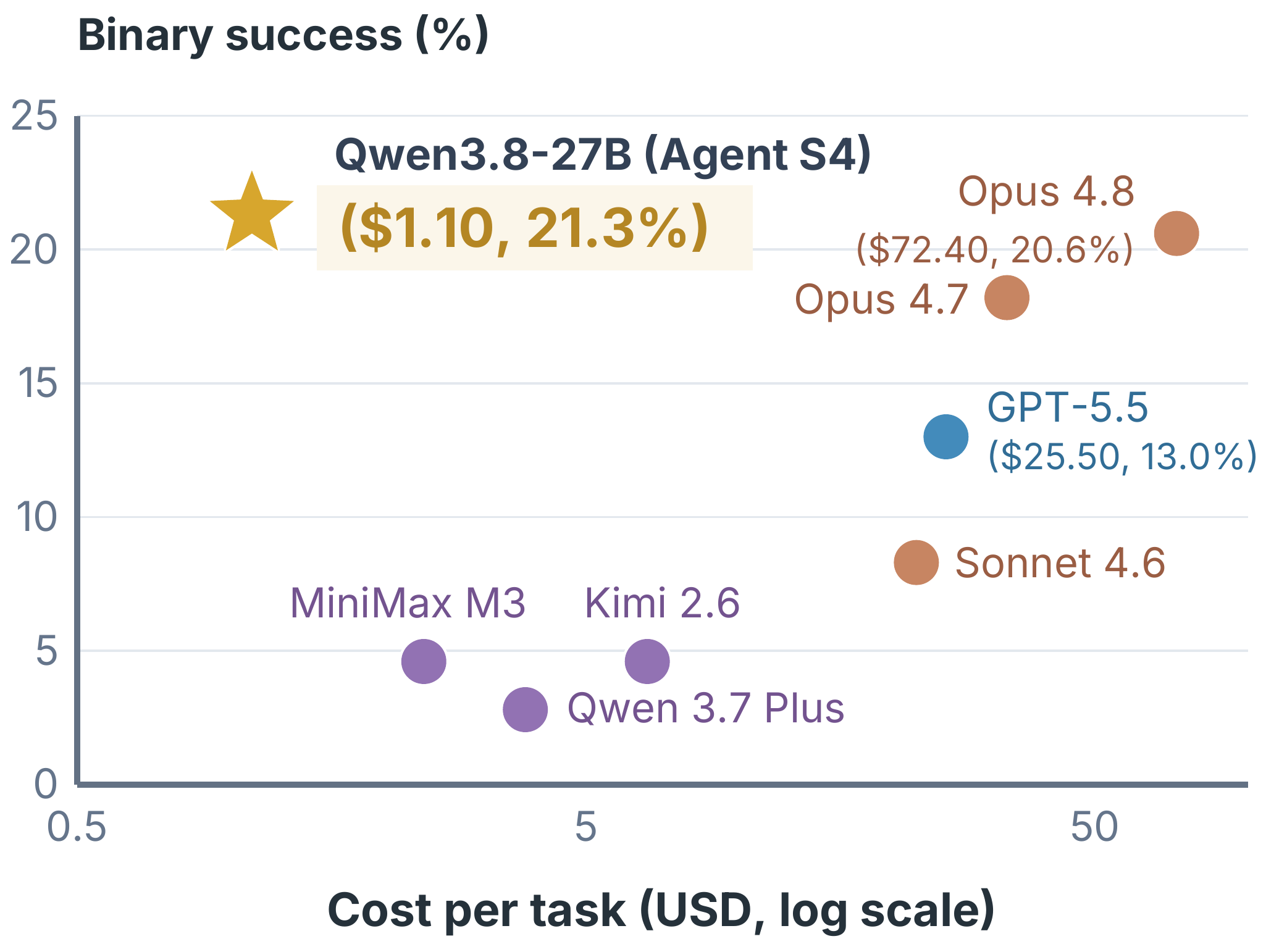}
  \captionsetup{font=footnotesize,skip=4pt}
  \caption{With Evolving Agent Interfaces built by \methodname{}, Qwen3.8-27B exceeds GPT-5.5 and Opus 4.8 in binary success on the challenging OSWorld V2 benchmark at a small fraction of their cost.}
  \label{fig:intro-performance}
  \vspace{-4pt}
\end{wrapfigure}

Instead, computer-use interfaces should evolve from the agent's own experience.
Therefore, we introduce \textbf{\methodname{}}\footnote{The name reads as \textbf{Agent S}cientist \textbf{for (4)} Evolving Agent Interfaces} an autonomous CUA scientist that studies computer-use agents and builds \textbf{Evolving Agent Interfaces} through population-based evolution. 
Its output is a repository of mixed interfaces and knowledge about when and how to use them.
The task agent's model and harness remain fixed, with the harness governing model invocation, execution, permissions, and task budgets.

The task agent uses candidate repositories of Evolving Agent Interfaces to attempt synthetic tasks, producing trajectories without ground-truth success labels. A meta-agent studies these trajectories and forms hypotheses about how changes to actions, observations, or usage knowledge could help the task agent.
It constructs descendants through mutation of one repository or crossover between two. Descendants can add, revise, or remove interfaces and knowledge, and the population retains alternative designs for further development and recombination.

To limit specialization to the tasks that motivated these changes,
\methodname{} separates the evidence used for construction from that used for selection. The meta-agent receives training trajectories, while validation instructions, trajectories, diagnostics, and scores remain hidden
from it. Validation performance guides parent selection. We evaluate the final selected repository on the full benchmark. The meta-agent never observes any benchmark task while constructing interfaces, and the validation tasks influence only which repositories are selected during evolution.

We evaluate \methodname{} on OSWorld V2~\citep{yuan2026osworld2} and ScienceBoard~\citep{scienceboard}. On the full 108-task OSWorld V2 set, Qwen3.8-27B with the evolved repository reaches $21.3\%$ binary success, above both GPT-5.5 ($13.0\%$) and Claude Opus 4.8 ($20.6\%$) at about $4.3\%$ and $1.5\%$ of their respective costs. In a separate evolution run on ScienceBoard, which covers niche scientific applications, the evolved repository reaches a $58.6\%$ success rate, compared with $41.4\%$ for the human-designed
interface of Qwen CUA. These results show that agents can evolve their own computer-use interfaces from purely synthetic, verifier-free task experience and outperform human-designed interfaces. We compare our setup against evolutionary baselines such as linear evolution and measure the effect of giving the meta-agent a scratch virtual machine to test its evolved interfaces. Finally, we qualitatively compare agents that act through evolved interfaces with agents that use fixed, human-designed interfaces.

Our contributions are:
\begin{itemize}[leftmargin=*,nosep]
  \item \textbf{Evolving Agent Interfaces from synthetic, verifier-free
  experience.} We introduce \methodname{}, which evolves a repository of
  interfaces and usage knowledge from synthetic task experience, without
  success labels or task verifiers, while the task agent's model and harness
  stay fixed. Validation tasks the meta-agent never sees decide which
  repositories to develop further.
  \item \textbf{Evolving interfaces outperform fixed ones.} With a repository
  of interfaces evolved by \methodname{}, Qwen3.8-27B exceeds both GPT-5.5 and
  Claude Opus 4.8 in binary success on OSWorld V2 at a small fraction of their
  cost, and improves ScienceBoard success over a fixed-interface agent with the
  same model.
  \item \textbf{Analysis of evolved interfaces.} Repository and trajectory
  analyses show that evolved interfaces change both what the agent observes
  and how it acts, across desktop-wide, application-specific,
  cross-application and task-specific interfaces. We also compare evolution
  strategies, with and without a scratch environment for the meta-agent.
\end{itemize}

\section{Problem Formulation}
\label{sec:problem-formulation}
\label{sec:aci}
\label{sec:orienting}
% \eric{I split Sec 2 into two sections. Now Sec 2 is Problem Formulation and Sec 3 is Method.}
% \eric{but where is the problem statement? Repository ends with a vague "the design problem is to determine which repository changes to retain." Add an explicit objective there. For example: find $R$ that maximizes expected score on unseen tasks, $\max_R \mathbb{E}_{x\sim\mathcal{D}}[V_x(\mathrm{Rollout}(A,R,C,x))]$, when construction can only see training trajectories. That makes Sec 2 a real problem formulation and gives Sec 3 something to solve.}

We study how the interfaces of a computer-use agent can be evolved from task experience.
An agent updating a spreadsheet through screenshots must identify visible
cells, enter values, and inspect the display to check the result. A
structured interface can expose the relevant rows and formulas, accept a
bulk edit, and return the changed values. Both interfaces can achieve the
same subgoal \citep{DBLP:conf/interact/CoutazNSBMY95}, but they require
different decisions and provide different feedback. Their cost and
reliability can also differ. Task experience can reveal which operations
are missing, which observations are difficult to use, and where a different
interface would help.

\label{sec:sp}
We call reusable interfaces that are constructed and improved from task
experience \emph{Evolving Agent Interfaces}.
To make interfaces and knowledge available across tasks, we represent them
in a persistent repository \(R\). It contains executable interfaces and
supporting knowledge such as capability descriptions, reusable workflows,
configuration, and notes about when and how to use them.

Let \(\mathcal{D}\) be the target distribution of tasks and \(A\) the agent
that uses the repository. Let \(J(A,R,x)\) denote the expected task score
when \(A\) uses repository \(R\) to attempt task \(x\). We seek a
repository constructed from task experience that maximizes this score across
tasks,
\begin{equation}
  R^{\star}\in\arg\max_R
  \mathbb{E}_{x\sim\mathcal{D}}
  \!\left[J(A,R,x)\right].
  \label{eq:repository-objective}
\end{equation}
This objective concerns performance beyond the experience used to construct
the repository. We next define the agent, interfaces, and interaction
process that determine \(J\).

\paragraph{Computer and task.}
A computer \(C\) has a state \(s\in\mathcal{S}\) comprising its files,
running applications, displays, services, and reachable network endpoints.
A task \(x=(g,s_0,V_x,B_x)\) consists of an instruction \(g\), an initial
state \(s_0\), an evaluator \(V_x\) that scores the outcome, and a budget
\(B_x\).

\paragraph{Interface.}
An \emph{interface}
\(I=\langle c_I,\mathcal{A}_I,\mathcal{O}_I\rangle\) connects an agent to
one or more parts \(c_I\) of the computer through a language of requests
\(\mathcal{A}_I\) and a language of observations \(\mathcal{O}_I\).
Executing a request can change the state and returns an observation,
\((s',o)\sim T_I(\cdot\mid s,a)\). A GUI interface, for example, accepts
clicks and keystrokes and returns screenshots. A shell interface accepts
bash commands and returns a string with the requested information, return values or error message. 

\paragraph{Model, harness, and agent.}
A \emph{model} \(M\) is a language model that generates responses and
requests conditioned on its input context. A \emph{harness} \(H\)
manages model invocation, request execution, task budgets,
permissions \(P_H\), and preparation of the model's context, including any
context compaction. An \emph{agent} \(A=(M,H)\) is the model operating
through this harness. The agent initially has access to a base set of interfaces
\(\mathcal{I}\). The harness executes requests through these interfaces and
places the resulting observations in the model's context.
For example, the harness can execute a spreadsheet helper
under a time limit and permission boundary. The helper defines the requests
for reading or editing cell ranges and the values returned to the agent.
The harness governs execution, while the interface specifies the available
operations and feedback.

\paragraph{Computer use.}
\emph{Computer use} is the process through which an agent observes and acts
on a computer through available interfaces to accomplish a task.
The interaction history \(h_t\) records the interfaces,
requests, and observations from earlier steps. At step \(t\), the agent
chooses an interface and a request based on the instruction and this history,
with context preparation by \(H\) left implicit,
\begin{equation}
  (I_t,a_t)\sim M(\cdot\mid g,h_t,\mathcal{I}),
  \qquad
  (s_{t+1},o_t)\sim T_{I_t}(\cdot\mid s_t,a_t).
  \label{eq:aci-step}
\end{equation}
The available interfaces shape
which operations the agent can request and what evidence it receives about
their effects.

Formally, \(R=(\mathcal{I}(R),K(R))\), where \(\mathcal{I}(R)\) contains
its executable interfaces and \(K(R)\) its supporting knowledge.
The repository
supplies interfaces \(\mathcal{I}_R=\mathcal{I}\cup\mathcal{I}(R)\) and
knowledge that informs the agent's choices,
\begin{equation}
  (I_t,a_t)\sim M(\cdot\mid g,h_t,\mathcal{I}_R,K(R)).
  \label{eq:repository-channels}
\end{equation}
Each attempt is subject to the task budget \(B_x\) and harness permissions
\(P_H\). The score \(J(A,R,x)\) averages the evaluator's score over these
interactions.

\paragraph{Interface evolution.}
Interfaces implemented as programs can be revised like any other software.
A program can export a material schedule through Blender's Python API,
extract selected rows from a Calc workbook, or return supplier prices
gathered through a sequence of browser operations. Each defines its own
requests and observations
\citep{wang2024executablecodeactionselicit,DBLP:conf/icra/LiangHXXHIFZ23}.
Because these interfaces are code, an agent that can read trajectories and
write code can redesign them. It can decide which operations to combine,
what information to return, and how a result can be checked. We call this
process \emph{interface evolution}. Given a repository \(R_k\) and the
trajectories \(\mathcal{T}_k\) of the agent using it, an evolution step
produces a revised repository,
\begin{equation}
  R_{k+1} = E\!\left(R_k,\mathcal{T}_k\right).
  \label{eq:interface-evolution}
\end{equation}
Task experience guides each step. Repeated reconstruction of the same
calculation across tasks can motivate a reusable operation, and a repeated
inability to verify a change can motivate a new observation. Trajectories
with the revised repository then provide evidence for further revision.
Section~\ref{sec:method} instantiates \(E\) with population-based
evolution.

\section{\methodname{}}
\label{sec:method}

\begin{figure*}[t]
  \centering
  \includegraphics[width=\textwidth]{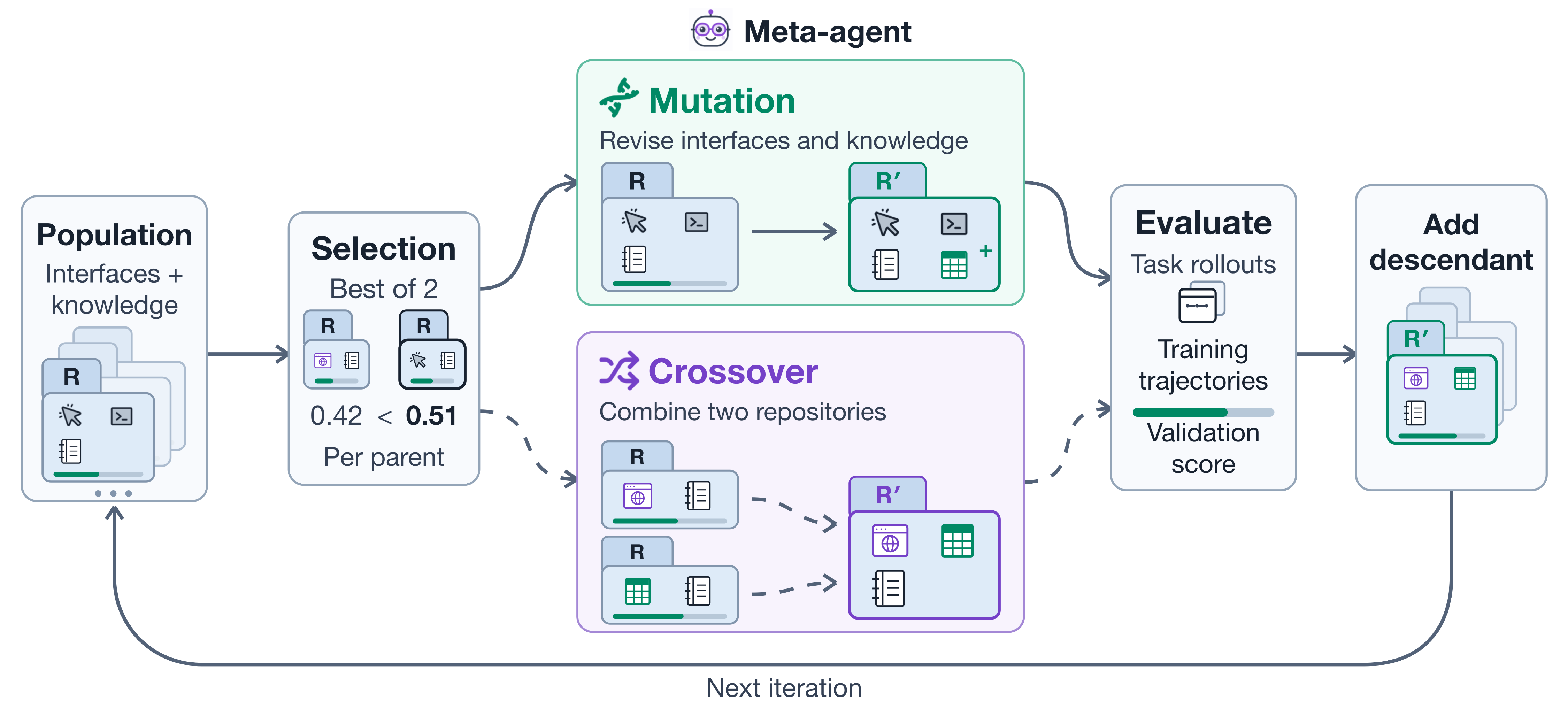}
  \caption{\textbf{\methodname{} turns task experience into an evolving repository of interfaces and knowledge.}
  Starting from parents selected by held-out validation, the meta-agent
  studies their task trajectories to identify limitations, then revises one
  repository through mutation or combines ideas from two through crossover.
  Task agents then use the descendant to produce trajectories on synthetic
  tasks and a separate score on held-out validation tasks. Each descendant
  joins the population alongside its parents, allowing earlier designs to be
  revisited and recombined over successive iterations.}
  \label{fig:agent-s4-overview}
\end{figure*}

\methodname{} builds Evolving Agent Interfaces and usage knowledge
through population-based evolution of repositories, keeping the task agent's
model \(M\) and harness \(H\) fixed. Each iteration selects parents from a population,
constructs a descendant through mutation or crossover, evaluates it, and
adds it to the population while retaining its parents
(Figure~\ref{fig:agent-s4-overview}). Training experience guides the
repository changes, and held-out validation guides subsequent selection.
This optimizes Equation~\ref{eq:repository-objective} for the fixed task
agent \(A\). Algorithm~\ref{alg:agent-s4} gives the
procedure.

\subsection{Population and selection}
\label{sec:population}
\label{sec:selection}
The population contains repository variants, each with its own interfaces
and knowledge. We initialize \(\mathcal{R}_0=\{R_0\}\) from a seed
repository (Appendix~\ref{sec:appendix-seed-repository}) and evaluate it to obtain training trajectories and a validation
score. These records are retained with each candidate as the population
grows.

At each step, an operator schedule chooses mutation or crossover.
Mutation requires one parent, and crossover requires two distinct parents.
For each parent, binary tournament selection draws two eligible variants
uniformly and chooses the one with the higher validation score. A sole
eligible candidate is selected directly. For crossover, the first selected
parent is excluded from the second selection. The selected repositories
and their saved training trajectories provide the starting point for the
next descendant.

\subsection{Mutation and crossover}
\label{sec:study}
The \emph{meta-agent} \(A_{\mathrm{meta}}=(M_{\mathrm{meta}},H_{\mathrm{meta}})\)
constructs descendants using a fixed model and harness. Its computer
\(C_{\mathrm{meta}}\) holds the selected parent repositories, their training
trajectories, and a lineage notebook in its filesystem. It studies the
recorded behavior and observations to diagnose apparent failures and
inefficiencies and form hypotheses about useful changes. For example,
repeated cell-by-cell edits can motivate a bulk spreadsheet interface,
while poor use of an existing interface can motivate revised usage
knowledge. It implements these hypotheses through file, shell, and code
interfaces using one of two operators.

\paragraph{Mutation.}
\label{sec:mutation}
Starting from one parent \(R_p\), the meta-agent adds, revises, or removes
interfaces and knowledge. Changes can affect operations, returned
observations, and reusable workflows, including capabilities inherited
from the seed.

\paragraph{Crossover.}
\label{sec:crossover}
Given two distinct parents \(R_a\) and \(R_b\), the meta-agent constructs
one coherent descendant by choosing between overlapping interfaces,
porting complementary capabilities, or combining their ideas into new
code and knowledge. The descendant need not contain the union of its parents.

Both operators produce a single descendant repository \(R'\). The fixed
base access \(\mathcal{I}\) remains unchanged as the repository's interfaces
\(\mathcal{I}(R)\) and knowledge \(K(R)\) evolve.

\subsection{Evaluation and population update}
\label{sec:sample}
After construction, we freeze \(R'\) and run the fixed task agent \(A\)
on synthetic training tasks \(\Ttrain\) and disjoint validation tasks
\(\Tval\). The repository is available in the task agent's filesystem, and
a fixed prompt identifies its location and explains how to discover its
capabilities. The agent can consult its documentation and source code and
invoke the interfaces it selects through the harness's existing
code-execution tool (Appendix~\ref{sec:appendix-repository-access}). Each rollout
\(\tau\sim\operatorname{Rollout}(A,R',C,x)\) starts from task \(x\)'s
initial state within its budget and records the instruction, interfaces,
requests, and observations. The repository remains fixed during execution.
Temporary programs written by the agent do not persist across tasks.

\paragraph{Training experience and validation score.}
Training trajectories have no ground-truth success labels. We save them as
evidence for future repository changes and reuse them whenever the
descendant becomes a parent. Validation rollouts provide a separate
selection score for each candidate \(R\),
\begin{equation}
  F_{\mathrm{val}}(R) =
  \frac{1}{|\Tval|}
  \sum_{x\in\Tval}
  V_x\!\left(
    \operatorname{Rollout}(A,R,C,x)
  \right).
  \label{eq:validation-fitness}
\end{equation}

\paragraph{Asymmetric evaluation.}
\label{sec:splits}
Only training trajectories are exposed to the meta-agent. Validation tasks,
trajectories, diagnostics, and scores remain inaccessible to it.
Validation therefore influences construction through parent selection,
without supplying direct feedback for repository edits.

\paragraph{Adding the descendant.}
\label{sec:elimination}
We add \(R'\) to the population alongside its parents regardless of its
validation score. This retains alternative designs for further development
and recombination. Parent links form a directed acyclic graph, with one
parent per mutation and two per crossover. Optional pruning removes
parent eligibility while preserving lineage records. The updated population
supplies candidates for the next iteration. After \(N\) steps, we select a
member of \(\mathcal{R}_N\) by validation performance, using a
benchmark-specific criterion (Section~\ref{sec:setup}), and freeze it for evaluation on
target tasks \(\Ttest\). The meta-agent observes none of these tasks during
construction, and validation tasks drawn from the same benchmark affect only
selection.

\begin{algorithm}[t]
\caption{\methodname{} evolves repositories by selecting parents, constructing descendants from task experience, and returning evaluated descendants to the population.}
\label{alg:agent-s4}
\begin{algorithmic}[1]
\Require Seed repository \(R_0\), task agent \(A\), meta-agent \(A_{\mathrm{meta}}\),
training tasks \(\Ttrain\), validation tasks \(\Tval\), operator schedule
\(\sigma\), evolution budget \(N\)
\State \(\mathcal{R}_0 \gets \{R_0\}\)
\State Run \(A\) with \(R_0\) on \(\Ttrain\) and \(\Tval\)
\For{\(k=1,\ldots,N\)}
    \State \(z \gets \sigma(k)\) \Comment{\(\textsc{Mutation}\) or
    \(\textsc{Crossover}\)}
    \State Select one or two parents from \(\mathcal{R}_{k-1}\) by
    validation tournament
    \State \(C_{\mathrm{meta}} \gets\) the parents, their training trajectories, and lineage notebook
    \State \(R' \gets\) descendant constructed by \(A_{\mathrm{meta}}\) on \(C_{\mathrm{meta}}\) using \(z\)
    \State Freeze \(R'\) and run \(A\) with it on \(\Ttrain\) and \(\Tval\)
    \State \(\mathcal{R}_k \gets \mathcal{R}_{k-1}\cup\{R'\}\)
    \State Record parent links and apply population control, if configured
\EndFor
\State Select \(\hat{R}\) from \(\mathcal{R}_N\) by validation score
\State \Return \(\hat{R}\), \(\mathcal{R}_N\), and the lineage record
\end{algorithmic}
\end{algorithm}

\section{Experimental Setup}
\label{sec:setup}

\paragraph{Benchmarks and task sets.}
OSWorld V2~\citep{yuan2026osworld2} is our primary benchmark, and
ScienceBoard~\citep{scienceboard} tests generalization to niche scientific
computer-use applications and workflows. Repository evolution uses 30
synthetic training tasks for OSWorld V2 and 48 for ScienceBoard, plus 15
official validation tasks for each. Synthetic task attempts yield
trajectories without ground-truth success labels. The meta-agent receives
these training trajectories. Validation scores guide selection, but
validation trajectories and evaluators remain outside its evidence.
Full-benchmark results cover 108 OSWorld V2 and 169 ScienceBoard tasks,
including the 15 validation tasks of each. Benchmark details appear in
Appendix~\ref{sec:appendix-experimental-design}, and synthetic task
construction is described below and in
Appendix~\ref{sec:appendix-synthetic-tasks}.

\paragraph{Model and harness.}
On both benchmarks, the task agent and meta-agent use Qwen3.8-27B~\citep{qwen2026qwen38modelcard} in a
modified Prime Agent harness~\citep{DBLP:journals/corr/abs-2608-23552} that executes code in a persistent IPython
kernel, with Medium reasoning, a 131,072-token context window, a
32,000-token output limit, and automatic context compaction. The model and
harness remain fixed within each search.
Appendix~\ref{sec:appendix-experimental-design} gives the harness version,
sampling parameters, and compaction settings;
Appendix~\ref{sec:appendix-harness-prompts} gives implementation
constraints and builder prompts.

\paragraph{Baselines.}
On OSWorld V2, we evaluate the task agent with no repository, the seed
repository, and the evolved repository, and compare against Qwen
CUA.\footnote{We use the Qwen CUA implementation provided by OSWorld V2:
\url{https://github.com/xlang-ai/OSWorld-V2/tree/main/mm_agents/qwen}.}
On ScienceBoard, we compare the seed and evolved repositories with Qwen CUA.
Qwen3.8-27B results come from our own evaluations. Results for other models
in Figure~\ref{fig:intro-performance} and Table~\ref{tab:s4-main-results},
including per-task costs, are taken from Table~3 of
\citet{yuan2026osworld2}.
% These reference runs use release \texttt{v2026.06.24}.

\paragraph{Evolution baselines.}
Table~\ref{tab:evolution-setup} summarizes the four evolution baselines, all starting
from the same seed repository. Each construction step produces one descendant. Seed evaluation and validation re-evaluations are not construction steps. The OSWorld V2 searches share the same training and
validation sets. 

\begin{table}[!htbp]
  \centering
  \small
  \setlength{\tabcolsep}{5pt}
  \renewcommand{\arraystretch}{1.15}
  \caption{Repository-evolution runs. Training tasks are synthetic and
  ungraded, and validation tasks are official. Descendant counts exclude the
  seed. Scratch VM access applies to the meta-agent during construction.}
  \label{tab:evolution-setup}
  \begin{tabular*}{\linewidth}{@{\extracolsep{\fill}}llcrrr@{}}
    \toprule
    Benchmark & Search & Scratch VM & Train & Val. & Descendants \\
    \midrule
    OSWorld V2 & Population & No & 30 & 15 & 30 \\
    OSWorld V2 & Population & Yes & 30 & 15 & 30 \\
    OSWorld V2 & Linear & Yes & 30 & 15 & 30 \\
    ScienceBoard & Population & No & 48 & 15 & 30 \\
    \bottomrule
  \end{tabular*}
\end{table}

Population search alternates mutation (one parent) and crossover (two distinct parents), starting with mutation. Binary tournaments choose parents from the eligible population using mean partial validation score on OSWorld
V2 and validation success on ScienceBoard. After 10 steps, the five lowest-scoring variants are removed from parent eligibility. Linear evolution mutates the latest descendant at every step, without tournament
selection or crossover.

A scratch VM with the same AMI as the benchmark tasks lets the meta-agent inspect running applications and try interface code during construction. Without it, the meta-agent works from repository code and saved training trajectories. In all four settings, task agents execute in benchmark VMs. Each meta-agent construction has a two-hour wall-clock budget for OSWorld v2 and 1 hour wall clock budget for ScienceBoard.

\paragraph{Validation and repository selection.}

On OSWorld V2, we select the repository that fully solves the most validation tasks and break ties by mean partial validation score, since binary success is the harder metric to improve. On ScienceBoard, whose evaluator reports only success, we select the repository with the highest validation success rate. 

In the case of Scienceboard, we also run a Denoise operation after every 5 steps. This picks the top-3 variants in the population and reevaluates them. The repositories validation score is then replaced by the updated running average. We analyze the effect of this denoising step in Section \ref{sec:results-denoising}. We do not run the denoising step for OSWorld-v2 due to much longer time and cost of evolution and repeated evaluations. 
% The OSWorld V2 main results retain the original \texttt{if-011}, selected by
% binary validation success among the first 15 no-scratch descendants. The
% scratch and linear comparison repositories, \texttt{if-024} and
% \texttt{if-029}, maximize mean partial validation score within their first
% 30 steps. These selection rules are separate from the partial-score signal
% used for population parent selection.

% For ScienceBoard, every five construction steps we re-evaluate the three
% highest-ranked repositories on all 15 validation tasks. Each repository's
% selection score becomes the mean of its initial validation score and all
% available repeats, and subsequent tournaments use the updated scores. Over
% 30 steps, these six rounds add 18 validation passes. We freeze
% \texttt{if-016}, which has the highest final mean, for full-benchmark
% evaluation; Section~\ref{sec:results-denoising} analyzes this procedure.

\paragraph{Evaluation budgets and cost.}
Prime Agent task rollouts have a 300-step budget on both benchmarks, where a
step is one completed assistant response and can contain multiple actions.
New OSWorld V2 training and validation rollouts also have a two-hour
wall-clock limit. The final evaluation keeps the step budget but has no
per-task wall-clock limit. ScienceBoard applies a 60-minute wall-clock limit
to training, validation, validation repeats, and full evaluation.

We report binary success and partial score on OSWorld V2 and success rate on
ScienceBoard. OSWorld V2 task-solving costs are estimated from model-token
usage at Parasail's input and output rates, assuming an $80\%$ input-cache
hit rate, and exclude repository construction and runtime infrastructure.
Appendix~\ref{sec:appendix-experimental-design} lists the rates and
remaining runtime settings.

\paragraph{Synthetic task construction.}
We constructed 30 synthetic tasks for OSWorld V2 and 50 for ScienceBoard, of which 48 are used (Appendix~\ref{sec:appendix-scienceboard-synthetic-tasks}),
giving Agent S4 task experience for evolving Agent Interfaces
and usage knowledge without ground-truth success labels. Tasks follow each
benchmark's methodology but use independently authored scenarios. For
OSWorld V2~\citep{yuan2026osworld2}, we studied its task-construction
principles and challenge-phenomenon taxonomy, and scaled the reported counts
of overlapping phenomena into approximate targets for our 30 tasks. These
targets guided the representation of capabilities such as cross-source
reasoning, visual-spatial precision, and implicit-state inference. For
ScienceBoard~\citep{scienceboard}, we followed a planned distribution across
six scientific applications and preserved its compact tasks, usually
centered on one application and one meaningful outcome.

Tasks were created agentically by a coordinator (GPT-6 Astra), authors
(GPT-5.6-Sol), and independent reviewers (GPT-5.6-Sol). The coordinator
tracked accepted tasks against the planned counts and assigned new tasks to
phenomena or applications below target. Each author started from a fresh
context with application documentation, design requirements, and summaries
of accepted synthetic tasks, but without access to official task instances,
evaluators, or task-agent trajectories. Authors produced instructions, input
artifacts, and environment setup specifications, documenting provenance and
capability assumptions.

Reviewers checked originality, coherence, feasibility, and appropriate
complexity. To ensure that no synthetic task overlaps with an existing
benchmark task, they compared each candidate's purpose, dependencies,
operations, and deliverables against official instruction and setup metadata
and against the accepted synthetic set. Each claimed challenge had to be supported by the
actual materials. The coordinator returned concrete defects for up to two
revision rounds, rejected overlapping designs, and kept official comparisons
out of author feedback. Automated input and setup checks supplemented these
reviews, and only accepted versions entered the training set. No step graded
task-agent attempts or produced reference solutions.

\section{Main Results}
\label{sec:results}

% QUANTITATIVE DRAFT: The main table contains preliminary reported results.
% Remaining tables await measured results. Three attempts per task are planned
% for final evaluation; the current figures are not confirmed three-run means.
% ScienceBoard reports the completed 60-minute evaluation in Generalization.
% Qualitative sources: yuzhe-evolution-analysis/evo_exp_1bb32f15/
% INTERFACE_DISCOVERIES_REPORT.md, prepared 2026-09-24 from TobyYang7/evo-exp
% at dataset revision 1bb32f15b2d86b89e2c2be3bb2f660811f1e4452, and
% yuzhe-evolution-analysis/synthetic_seed_if001/REPORT.md, prepared from
% TobyYang7/shadow-interface-evolution-trajectories at revision
% ac3c78204a67831952631804c22089e0a8ee603a. The qualitative cases come from
% exploratory archived runs and are distinct from the final aggregate evaluation.

\label{sec:results-main}

\begin{table}[!htbp]
  \centering
  \small
  \setlength{\tabcolsep}{2pt}
  \renewcommand{\arraystretch}{1.12}
  \caption{Results on the full OSWorld-v2 Benchmark. With the same Qwen3.8-27B model, repositories of interfaces evolved by \methodname{} outperform the fixed interfaces of the Qwen CUA at lower cost. Evolved rows use the best repository from population evolution with and without a scratch VM for the meta-agent. Section~\ref{sec:results-evolution} compares them with the seed repository and other evolution strategies.}
  \label{tab:s4-main-results}
  \begin{tabular*}{\linewidth}{@{\extracolsep{\fill}}llcrrr@{}}
    \toprule
    Model / Method & Interface Type & Parameters & \shortstack{Binary\\(\%) $\uparrow$} & \shortstack{Partial\\(\%) $\uparrow$} & \shortstack{Cost/task\\(\$) $\downarrow$} \\
    \midrule
    \multicolumn{6}{@{}l}{\textit{Open-source models}} \\
    MiniMax M3 (single) & Fixed & 428B (23B) & 4.60 & 22.30 & 2.40 \\
    Kimi K2.6 (single) & Fixed & 1T (32B) & 4.60 & 22.10 & 6.60 \\
    Qwen CUA (Qwen3.8-27B) & Fixed & 27B & 5.56 & 20.69 & 1.39 \\
    % Prime Agent (Qwen3.8-27B), no repository: Extensible, 15.74 / 44.93, ~$1.359; now reported in Table~\ref{tab:s4-evolution-comparison}.
    Evolved repo. (w/ scratch VM) & Evolving & 27B & 19.44 & \textbf{51.06} & 1.27 \\ % if-024
    \textbf{Evolved repo. (w/o scratch VM)} & Evolving & 27B & \textbf{21.30} & 46.90 & \textbf{1.10} \\ % if-011
    
    \midrule
    \multicolumn{6}{@{}l}{\textit{Closed-source models}} \\
    Claude Opus 4.8 (batched) & Fixed & --- & 20.60 & 54.80 & 72.40 \\
    Claude Opus 4.7 (batched) & Fixed & --- & 18.20 & 48.91 & 33.60 \\
    GPT-5.5 (batched) & Fixed & --- & 13.00 & 49.50 & 25.50 \\
    Claude Opus 4.8 (single) & Fixed & --- & 18.50 & 49.30 & 76.10 \\
    Claude Opus 4.7 (single) & Fixed & --- & 13.90 & 49.10 & 35.80 \\
    Claude Sonnet 4.6 (single) & Fixed & --- & 8.30 & 41.50 & 22.30 \\
    Qwen3.7-Plus (single) & Fixed & --- & 2.80 & 21.50 & 3.80 \\
    \bottomrule
  \end{tabular*}
\end{table}

\textbf{With a repository of interfaces evolved by \methodname{}, a 27B open-source model reaches the highest binary success on open-source models at a small fraction of frontier-model cost.} Table~\ref{tab:s4-main-results} compares results on the full OSWorld V2 benchmark. We report the best repositories from population evolution with and without a scratch VM, alongside the Qwen CUA baseline and reported open-source and closed-source results from OSWorld V2, which retain the single-action and batched-action settings of the reference evaluations. Binary success measures complete task execution, partial score measures intermediate progress, and cost covers task-solving model calls. Binary gains are the hardest to obtain on OSWorld V2. Strong agents reach similar partial progress, and what separates them is turning that progress into fully completed tasks~\citep{yuan2026osworld2}. With the repository evolved without a scratch VM, Qwen3.8-27B achieves $21.30\%$ binary success at \$1.10 per task, above the $13.00\%$ of GPT-5.5 and the $20.60\%$ of Claude Opus 4.8 with batched actions, at $4.3\%$ and $1.5\%$ of their \$25.50 and \$72.40 costs. With the repository evolved with a scratch VM, it reaches a partial score of $51.06\%$ at \$1.27 per task. 

\textbf{Evolving interfaces improve the same model.} Agents with fixed interfaces already seek programmatic access~\citep{yuan2026osworld2}. Terminal commands account for $24.7\%$ of their actions, and GPT-5.5 devotes $78\%$ of its budget to code, API, or file operations~\citep[Section~4]{yuan2026osworld2}. Under a fixed interface, each such program is typed through the desktop and its result is read from a screenshot. Acting through agent evolved interfaces removes this indirection, and the evolved repository supplies interfaces for the applications and data that recur across tasks. Compared with the Qwen CUA baseline, which runs the same Qwen3.8-27B model with fixed interfaces, this raises binary success from $5.56\%$ to $21.30\%$ and partial score from $20.69\%$ to $46.90\%$, or to $51.06\%$ with the repository evolved with a scratch VM, while lowering task-solving cost from $1.39 to $1.10 per task.

\section{Generalization}
\label{sec:results-generalization}

\begin{wraptable}{r}{0.52\textwidth}
  \vspace{-1.2em}
  \centering
  \small
  \setlength{\tabcolsep}{8pt}
  \renewcommand{\arraystretch}{1.18}
  \caption{Results on the full ScienceBoard benchmark. \methodname{} rows use Qwen3.8-27B in Prime Agent with the seed or the evolved repository.}
  \label{tab:s4-scienceboard}
  \begin{tabular}{lr}
    \toprule
    Method & Success (\%) $\uparrow$ \\
    \midrule
    Qwen CUA & 41.4 \\
    \midrule
    Agent S4 Seed repository & 55.03 \\
    Agent S4 Evolved repository & \textbf{58.58} \\
    \bottomrule
  \end{tabular}
\end{wraptable}
% see if you can add cost, add more models 

We next ask whether \methodname{} carries over to a different domain. ScienceBoard~\citep{scienceboard} contains 169 tasks across six scientific applications. These are Celestia for astronomy, ChimeraX for molecular visualization, GRASS GIS for geographic analysis, KAlgebra for algebra and plotting, Lean for theorem proving and TeXstudio for typesetting. General computer-use interfaces rarely cover these applications, and many of their operations are hard to carry out through clicks alone. We run a fresh evolution with the same task agent, Qwen3.8-27B in Prime Agent, starting from the same seed repository as on OSWorld V2. The meta-agent learns from 48 synthetic training tasks across the six applications (Appendix~\ref{sec:appendix-scienceboard-synthetic-tasks}), and 15 validation tasks guide selection over 30 construction steps.

Table~\ref{tab:s4-scienceboard} reports results on the full benchmark. Evolution raises success from $55.03\%$ with the seed repository to $58.58\%$, solving 99 of 169 tasks compared with 93, and reduces the average number of model calls per task from 61.7 to 58.9. The selected repository grows from 15 functions in the seed to 64. The additions include a ChimeraX command interface, GRASS GIS modules that run in the live session and a KAlgebra calculator that evaluates many expressions in one call. The gains are largest on KAlgebra, where success rises from $48\%$ to $68\%$. These results show that \methodname{} can evolve useful interfaces for specialized scientific software from synthetic training experience.

\section{Analysis}
\label{sec:results-analysis}

\subsection{Evolution Dynamics}
\label{sec:results-evolution}

\begin{table}[!htbp]
  \centering
  \small
  \setlength{\tabcolsep}{6pt}
  \renewcommand{\arraystretch}{1.18}
  \caption{Effect of the seed repository and comparison of evolution strategies on the full OSWorld V2 benchmark. All rows use Qwen3.8-27B in Prime Agent. Each evolution row reports the best repository selected by that strategy. Appendix~\ref{sec:appendix-seed-repository} lists the contents of the seed repository. With a scratch VM, the meta-agent can open applications and test interfaces while building the repository.}
  \label{tab:s4-evolution-comparison}
  \begin{tabular*}{\linewidth}{@{\extracolsep{\fill}}lcc@{}}
    \toprule
    Condition  & Binary (\%) $\uparrow$ & Partial (\%) $\uparrow$ \\
    \midrule
    No Repository & 15.74 & 44.93 \\
    Agent S4 Seed Repository & 19.44 & 44.90 \\
    \midrule
    \multicolumn{3}{@{}l}{\textit{Evolution strategy}} \\
    Linear, scratch VM &  19.44 & 47.17 \\
    Population, scratch VM &  19.44 & \textbf{51.06} \\
    Population, no scratch VM & \textbf{21.30} & 46.90 \\
    \bottomrule
  \end{tabular*}
\end{table}

% \begin{figure}[!htbp]
%   \centering
%   \includegraphics[width=\linewidth]{assets/agent_s4_validation_progress_comparison.pdf}
%   \caption{\textbf{Validation improvements arrive at different stages of search.}
%   OSWorld V2 validation progress for three searches over construction steps
%   0--30, with the shared cached seed evaluation at step 0. Faint markers show
%   individual candidates; lines show the best score observed so far, including
%   the seed. All searches reach $26.67\%$ binary success. Their best partial
%   scores are $47.86\%$, $50.23\%$, and $51.36\%$ for population search
%   without scratch, population search with scratch, and linear refinement,
%   respectively. The two metrics can peak at different repositories.}
%   \label{fig:s4-evolution-progress}
% \end{figure}

\textbf{Seed repository and evolution.} The first two rows of
Table~\ref{tab:s4-evolution-comparison} show the task agent with no
repository and with the seed repository. The seed raises binary success from
$15.74\%$ to $19.44\%$ while leaving partial score nearly unchanged
($44.93\%$ versus $44.90\%$), and lowers task-solving cost from \$1.36 to
\$1.11 per task. Evolution without a scratch VM then raises binary success
to $21.30\%$ and partial score to $46.90\%$ at nearly the same cost
(\$1.10). The seed therefore provides a useful starting point, and evolution
improves it on binary success and partial score. 

\textbf{Evolution strategies.} We compare three ways of evolving the
repository. The first is population-based evolution with a scratch virtual
machine (VM), in which the meta-agent can open applications and test
interfaces. The second is population-based evolution without the scratch VM.
The third is linear evolution with a scratch VM, in which the meta-agent
keeps refining each new child repository. With a scratch VM, population-based evolution reaches the same
binary success as linear evolution ($19.44\%$) and a higher partial score
($51.06\%$ versus $47.17\%$). Without a scratch VM, population-based evolution
reaches the highest binary success ($21.30\%$) but a lower partial score
($46.90\%$). The scratch VM therefore improves partial progress, while the
best binary success comes from evolution without it. 

\subsection{Qualitative Study of Evolved Interfaces}

\begin{figure}[!htbp]
  \centering
  \includegraphics[width=\linewidth]{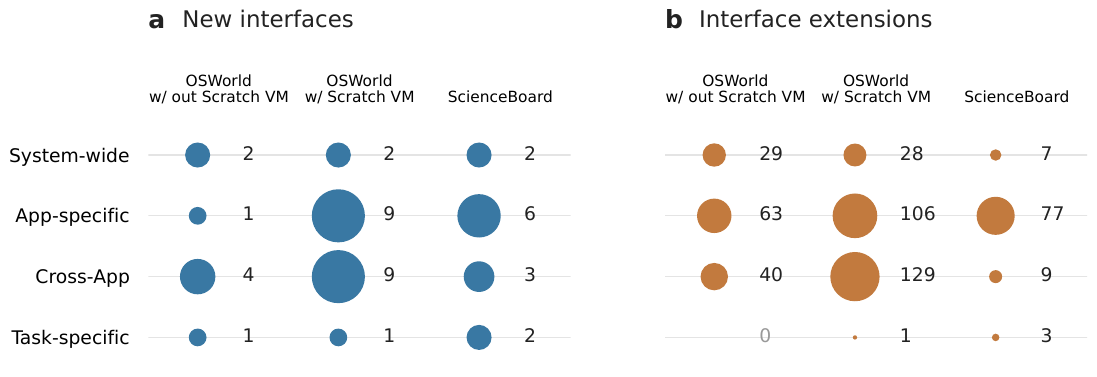}
  \caption{\textbf{Interface discovery and extension across 30 candidates per population run.}
  \textbf{a}, New interfaces (blue): distinct coherent interfaces absent from the seed, labeled by their implemented scope across the run.
  \textbf{b}, Interface extensions (orange): newly introduced operations first recorded after their interface's introduction, each labeled independently by its own behavior. }
  \label{fig:taxonomy-families}
\end{figure}

\begin{samepage}
\paragraph{A taxonomy of evolved interfaces.}
To compare runs, we label each interface by the part of the computer it
covers and the purpose it serves. The four labels can overlap:
\begin{itemize}[leftmargin=*,itemsep=2pt,topsep=4pt]
    \item \textbf{System-wide} interfaces cover the desktop and operating
    system as a whole, such as inspecting windows or launching and
    monitoring background processes.
    \item \textbf{App-specific} interfaces expose the objects of one named
    application, such as spreadsheet ranges in LibreOffice, tracks in a
    REAPER audio project, or molecules in ChimeraX.
    \item \textbf{Cross-App} interfaces work on data or platforms shared by
    many applications, such as reading and rendering PDF files regardless of
    which application created them, or reading pages on any website.
    \item \textbf{Task-specific} interfaces package a procedure for a
    particular goal together with a check of whether it is achieved, such as
    checking a Lean proof, compiling it, updating the editor and reporting
    completion.
\end{itemize}
\end{samepage}
Interfaces also grow after they are introduced, so we label each added
operation separately. For example, the office-document interface is App-specific
because it edits documents through LibreOffice itself, but an operation added
to it later that edits OpenDocument files directly, without the application,
is Cross-App. Figure~\ref{fig:taxonomy-families} counts new interfaces and
these later extensions separately. Table~\ref{tab:interface-examples} in
Appendix~\ref{sec:appendix-interface-examples} gives examples of evolved
interfaces, with what each lets the agent observe and do.

% Figure 5 body. The arXiv build places it before "Different runs grow
% different interfaces" so it can sit at the top of the page that holds the
% case study (set \caseearly in main_arxiv.tex). The ICLR build places it
% after the case study paragraph.
\newcommand{\casestudyfigure}{\begin{figure}[h]
  \centering
  \includegraphics[width=\linewidth]{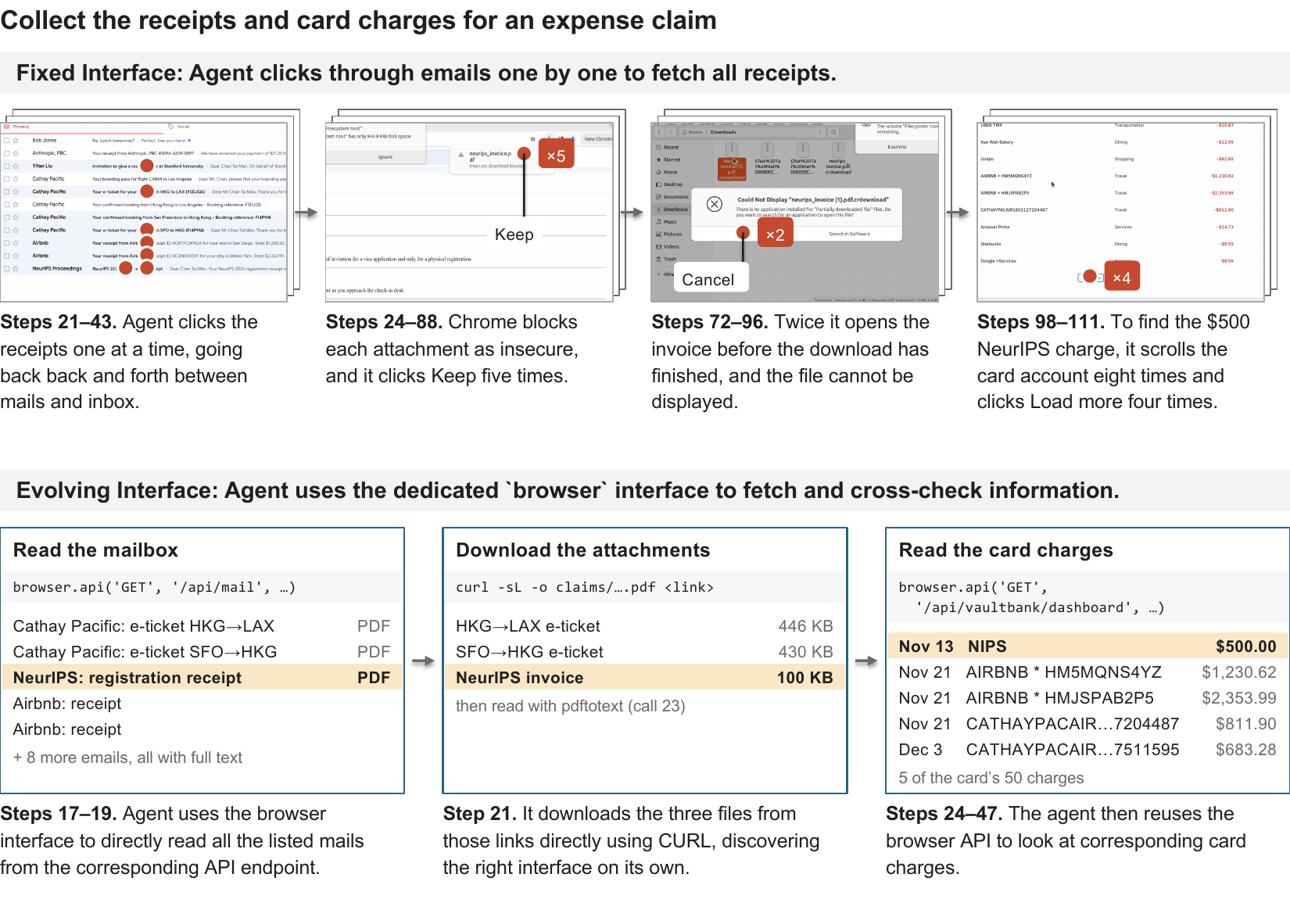}
  \caption{Human-curated versus Evolved Interface. Top: the Qwen CUA baseline uses a fixed, human-designed interface prioritizing frequent screenshots with mouse and keyboard actions. It reads the receipts and searches the card charges through repeated GUI interactions (steps 21--111). 
  Bottom: Prime Agent with the repository evolved by \methodname{}. It reads the mailbox and the bank records through the evolved \texttt{browser} interface, whose \texttt{browser.api} sends requests from inside the already logged-in tab and returns the data as text (calls 17 and 44). It fetches the attachments with \texttt{curl}, and reads them with \texttt{pdftotext}.}
  \label{fig:case-mailhub-cua}
\end{figure}}
\ifdefined\caseearly\casestudyfigure\fi

On OSWorld V2 without a scratch VM, evolution introduces four new interfaces
and 127 extensions, including 29 System-wide, 63 App-specific and 40
Cross-App extensions. An interface or extension can carry more than one label,
so the counts per label can add up to more than the total. Most of its growth therefore deepens interfaces the
agent already has. With a scratch VM, where the meta-agent can open
applications and test its code, the population introduces 12 new interfaces
and 253 extensions. Nine of the new interfaces are App-specific and nine are
Cross-App. Seven carry both labels, such as the interfaces for editing musical
scores, audio projects and 3D scenes, which operate one application while reading and writing
file formats that other applications share. Cross-App extensions rise to 129. These
additional application interfaces are a likely explanation for the higher
partial score of the best repository evolved with a scratch VM.

On ScienceBoard, evolution introduces seven new interfaces and 91 extensions.
Six of the new interfaces and 77 of the extensions are App-specific, for
scientific applications such as ChimeraX, KAlgebra, GRASS GIS and Lean. Its
three Task-specific extensions package procedures for displaying a map in
GRASS GIS and completing a proof in Lean.

Across all three runs, evolved interfaces let the agent observe the desktop,
an application's own objects, data shared across applications, and whether a
goal has been met, and act on each of these in a single call. Evolution
therefore widens the range of software the agent can operate and raises the
level at which it sees, acts and checks its progress.

\label{sec:results-case-studies}

\paragraph{Case Study: Using the evolved browser interface to handle web data search and fetching instead of a fixed interface.}
Figure~\ref{fig:case-mailhub-cua} follows the information-gathering portion of OSWorld V2 task 008, which requires receipts and corresponding card charges for an expense claim. The fixed-interface Qwen CUA agent opens receipt emails one at a time, handles five browser download warnings, and twice tries to
open an invoice before its download has finished. Finding the \$500 NeurIPS charge also requires eight scrolls and four clicks on \emph{Load more}.
With the evolved repository, the task agent uses \texttt{browser.api} to retrieve mailbox contents and attachment URLs as structured data. It downloads three PDFs with \texttt{curl}, reads them with
\texttt{pdftotext}, and reuses the browser interface to query the bank dashboard and cross-check the charges. The agent combines a reusable browser interface with existing command-line tools to gather evidence across applications, reducing repeated visual navigation and manual download handling.

\ifdefined\caseearly\else\casestudyfigure\fi

Appendix~\ref{sec:appendix-interface-cases} provides the more illustrated examples of agents benefitting from using an evolved interface for applications like MailHub, XMind, Zotero, ReviewSphere, and Word tracked changes.

\subsection{Repeated Validation during Evolution}
\label{sec:results-denoising}

% Figure 6 wraps beside its text in the arXiv build and is a normal float in
% the ICLR build. Retune \denoisewraplines in main_arxiv.tex if the layout
% before this subsection changes.
\newcommand{\denoisefigure}[1]{%
  \centering
  \includegraphics[width=#1]{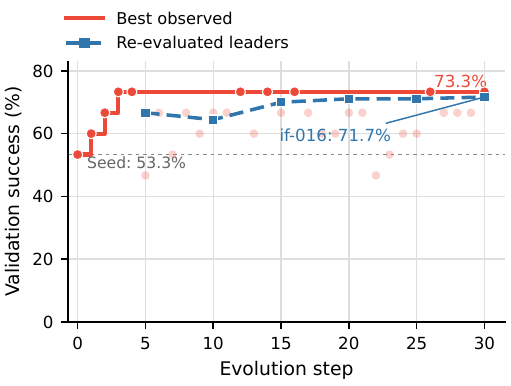}
  \caption{\textbf{Repeated validation separates a high first score from
  consistent performance.} Red shows the best first-evaluation score on
  ScienceBoard and blue the best mean over repeated evaluations.}
  \label{fig:s4-scienceboard-progress}}
\ifdefined\denoisewraplines
\begin{wrapfigure}[\denoisewraplines]{r}{0.4\linewidth}
  \vspace{-4.3em}
  \denoisefigure{\linewidth}
  \vspace{-1.5em}
\end{wrapfigure}
\else
\begin{figure}[h]
  \denoisefigure{0.64\linewidth}
\end{figure}
\fi

Long-horizon computer-use tasks show substantial run-to-run variance~\citep{DBLP:journals/corr/abs-2604-17849}, so a single validation run can overrate a repository. On ScienceBoard, we therefore re-evaluate the three highest-scoring repositories every five steps and score each by the mean of its evaluations (Section~\ref{sec:setup}). Figure~\ref{fig:s4-scienceboard-progress} shows why this matters. Repositories \texttt{if-003} and \texttt{if-004} reach the highest first score, $73.33\%$, at steps 3 and 4. A second evaluation lowers their means to $60.00\%$ and $66.67\%$. The selected repository, \texttt{if-016}, scores $73.33\%$, $66.67\%$, $73.33\%$ and $73.33\%$ across four evaluations, for a mean of $71.67\%$. A first score alone would have ranked \texttt{if-003} among the best repositories even though its mean is $60.00\%$.

\section{Related Work}

\paragraph{Computer-use agents and interaction design.}
Computer-use systems differ in how they expose observations and actions, and
SWE-agent showed that this agent-computer interface shapes
performance~\citep{sweagent}. Agent~S and Agent~S2 organize planning,
perception, and action through engineered agent
components~\citep{AgentS,AgentS2}, while UI-TARS learns policies for GUI
interaction~\citep{uitars,uitars2}. Other systems add programmatic access
beside GUI actions, through application
APIs~\citep{DBLP:conf/acl/LuZYZWDLR0Z25,DBLP:conf/acl/SongXZN25,DBLP:journals/tmlr/ZhangHNMQHWYZQDLKJZWQML26},
code execution~\citep{coact1}, or tool
calls~\citep{DBLP:journals/corr/abs-2510-17790,DBLP:conf/iclr/JiaLZXXJYLYH26}.
Agent~S3 and its behavior judge improve task completion by selecting among
multiple rollouts~\citep{agents3}. In these systems, code and tools, when
available, are one action among a predefined set. Agents that act only by
writing and running code can outperform agents that choose from predefined
actions~\citep{wang2024executablecodeactionselicit,webwright2026}. These
interfaces are designed by people or written within a single task.
\methodname{} instead evolves them across tasks and retains the resulting
operations and usage knowledge in a repository that can be developed beyond
the task that motivated a change.

\paragraph{Skills, memory, and tool construction.}
Reflexion stores textual feedback for later attempts~\citep{shinn2023reflexionlanguageagentsverbal},
and Agent Workflow Memory induces reusable procedures from
experience~\citep{awm}. Reuse can also take executable form. LLMs as Tool
Makers separates tool construction from subsequent tool
use~\citep{DBLP:conf/iclr/Cai00CZ24}, and SkillWeaver discovers and practices
web interactions to build APIs that other agents can
reuse~\citep{skillweaver}. CUA-Skill instead provides an engineered library
of parameterized computer-use skills and composition graphs~\citep{cuaskill}.
These methods establish that agents can benefit from persistent knowledge
and executable tools. Our search operates over complete repository versions
containing interfaces, observation routines, configuration, and usage
knowledge. Mutation and crossover can change these components together,
with selection based on their combined effect on task-agent performance.

\paragraph{Automated agent design and evolutionary search.}
ADAS searches for agent designs expressed in code~\citep{adas}, and the
Darwin G\"odel Machine modifies its own coding agent while maintaining an
archive of alternatives~\citep{dgm}. Self-Harness studies execution traces,
proposes harness changes, and validates them through regression
tests~\citep{selfharness}. AlphaEvolve and ShinkaEvolve combine
model-generated code changes with evolutionary search and executable
evaluation~\citep{alphaevolve,shinkaevolve}, while GEPA uses trajectory
reflection to evolve prompts and combine complementary
improvements~\citep{gepa}. Validation-based control of overfitting also
predates language-model agents, including evolutionary curtailment in
genetic programming~\citep{evovalsets}.
\methodname{} evolves repositories of interfaces and knowledge through
population-based search with asymmetric access to evaluation evidence.
The meta-agent studies synthetic-task trajectories without ground-truth
success labels and implements changes to application-specific operations,
capability descriptions, and reusable workflows. The fixed harness
retains control of model invocation, execution, permissions, and budgets.
Keeping capabilities in the repository makes them available for revision,
selection, recombination, and reuse with compatible agents and harnesses.
Validation tasks, trajectories, diagnostics, and scores remain with the
controller and are hidden from the meta-agent. Held-out performance
determines which variants become parents, applying selection pressure
beyond the tasks used to construct the software.

\section{Conclusion}

We presented \methodname{}, which treats interface design as an agent
capability: a meta-agent studies computer-use trajectories to construct and
evolve reusable interfaces and knowledge about when and how to use them.
The experience used for interface construction comes entirely from synthetic
tasks, without ground-truth success labels or task verifiers; separate
held-out validation guides repository selection. Qwen3.8-27B with a repository of
interfaces evolved by \methodname{} reaches $21.3\%$ binary success on the full 108-task OSWorld V2
benchmark, above both GPT-5.5 and Claude Opus~4.8, at about $4.3\%$ and
$1.5\%$ of their reported task-solving costs. On the full 169-task ScienceBoard
benchmark, a separate evolution run reaches $58.58\%$ success, compared with
$41.4\%$ for Qwen CUA. Repository and trajectory analyses show how evolution
constructs application interfaces and refines the operations and observations
available to subsequent task agents. These findings support learning
interfaces from synthetic experience as a way to improve computer use while
keeping the task agent's model and harness fixed, complementary to advances
in both.

\bibliography{custom}

\clearpage
\appendix

\section{Experimental Design}
\label{sec:appendix-experimental-design}

\subsection{Benchmark Detail}

% \noindent\textbf{OSWorld V1.}
% OSWorld~\citep{xie2024osworld} evaluates multimodal agents on open-ended
% tasks in real computer environments. The original benchmark contains 369
% tasks involving desktop and web applications, file operations, and workflows
% that span multiple applications. Each task includes an initial environment
% configuration and an execution-based evaluator that checks the resulting
% computer state or output artifacts. It measures whether agents can carry out
% practical computer operations and complete user instructions across diverse
% applications.

\noindent\textbf{OSWorld V2.}
\href{https://osworld-v2.xlang.ai/}{OSWorld V2}~\citep{yuan2026osworld2} evaluates computer-use
agents on 108 long-horizon workflows spanning everyday and professional
activities in real desktop and web applications. Tasks require agents to
combine information across sources, track application state, and adapt to
changes during execution. The benchmark reports both binary task completion
and partial scores, measuring whether agents can complete an entire workflow
and how much of its requirements they satisfy. We use it as our primary
benchmark for evaluating repository evolution on complex computer-use tasks.

\noindent\textbf{ScienceBoard.}
ScienceBoard~\citep{scienceboard} evaluates autonomous agents in realistic
scientific workflows involving professional software. It contains 169
human-curated tasks across scientific domains such as biochemistry,
astronomy, and geoinformatics. These tasks combine scientific reasoning with
software operation through graphical and command-line interfaces, testing
whether agents can translate domain-specific instructions into successful
execution in scientific environments. We use ScienceBoard to evaluate
whether repository evolution extends to specialized scientific applications.

\subsection{Settings}

\paragraph{Task subsets.}
The four evolution runs and their construction budgets are summarized in
Table~\ref{tab:evolution-setup}. OSWorld V2 uses all 30 authored synthetic
tasks. ScienceBoard uses 48 of the 50 authored tasks: we exclude
\texttt{sb\_grassgis\_010}, whose initialization fails a region check, and
\texttt{sb\_chimerax\_006}, whose input file fails to open in ChimeraX.
These exclusions concern task initialization, before task-agent execution.
Each benchmark uses 15 official tasks for validation. The full reported
sets contain 108 OSWorld V2 tasks and 169 ScienceBoard tasks, including
those validation tasks.

\paragraph{Comparison interfaces and harnesses.}
\label{sec:appendix-agent-interfaces}
The interface column in Table~\ref{tab:s4-main-results} describes the
execution access exposed to the task agent. All external reference results
use the OSWorld V2 reference harness with screenshot observations and a
500-step budget~\citep[Section~3.1]{yuan2026osworld2}.
Claude Opus and Sonnet use the native Claude computer-use interface;
GPT-5.5, MiniMax M3, Kimi K2.6, and Qwen3.7-Plus use the benchmark's
PyAutoGUI action backend. The released GPT and MiniMax adapters translate
structured computer actions into PyAutoGUI commands; this backend does
not imply an unrestricted Python execution tool.\footnote{Released
\href{https://github.com/xlang-ai/OSWorld-V2/blob/2b9b7b4eb73243d557bdbf2998fe18d8e18e19c6/mm_agents/gpt_response_api.py}{GPT}
and
\href{https://github.com/xlang-ai/OSWorld-V2/blob/2b9b7b4eb73243d557bdbf2998fe18d8e18e19c6/mm_agents/m3/parser.py}{MiniMax}
adapters in the benchmark's \texttt{v2026.06.24} release. These establish
available adapter behavior, rather than exact launch configurations for
every reported model version.}
The paper establishes the action backend for Kimi K2.6 and Qwen3.7-Plus,
but the public sources do not establish their exact scored-run adapters.
We label these reference interfaces \emph{Fixed}: their requests and
observations are set before the task. This describes the interaction
surface, not a prohibition on running programs through desktop
applications; such programs are entered through the GUI and their output
is observed in screenshots.

Our Qwen CUA baseline uses the upstream Qwen harness, with structured
\texttt{computer\_use} actions for mouse, keyboard, and screenshot control
translated into PyAutoGUI. Our Prime Agent conditions instead expose
persistent IPython execution, supporting Python and shell code for GUI,
file, and API operations. We label these interfaces \emph{Extensible}: programs
the agent writes during a task become additional interfaces whose output
returns directly to the agent. The seed and evolved conditions add repository
interfaces accessible through this same tool, as described below.

\paragraph{Harness.}
For our OSWorld V2 and ScienceBoard repository-evolution experiments, we use a modified
harness based on Prime Agent version
\texttt{0.8.0-beta.546.1.a9b5d88},\footnote{\url{https://github.com/PrimeIntellect-ai/prime-agent/releases}}
an IPython-backed coding agent. The harness prompts are reproduced in
Appendix~\ref{sec:appendix-harness-prompts}. The model issues tool calls containing
Python code, which the harness executes in a persistent IPython kernel and
returns the execution outputs to the model for the next step. Variables,
imports, and function definitions persist across calls, allowing the agent
to reuse intermediate results and invoke repository interfaces throughout a
task. Prime Agent's native threshold-based context compaction is enabled
within the 131,072-token context window. The harness reserves 36,096 tokens
for the 32,000-token output budget and a 4,096-token overhead margin.
Compaction is triggered when the last observed token usage plus the estimated
trailing messages exceeds 94,976 tokens. It summarizes older conversation
history while retaining the most recent 20,000 tokens verbatim.
Prime Agent also automatically rescales oversized images to fit within
$2000\times2000$ pixels while preserving aspect ratio, with further
compression or downscaling as needed to keep the base64-encoded payload
below 4.5\,MiB. We disable Prime Agent’s automatic harness refinement. The task agent can still record notes in the harness state during a task, but this state is deleted before every task, so the repository is the only state that carries over between tasks.

\paragraph{Repository access.}
\label{sec:appendix-repository-access}
For OSWorld V2, each candidate repository is installed at
\path{/home/user/shadow_interface/}. A fixed task-agent prompt supplies this
path, directs the agent to inspect the toolkit, and encourages its use as
the primary approach. The agent invokes \texttt{python3 -m shadow\_interface}
through the harness's IPython tool. The \texttt{capabilities} subcommand
lists interfaces with signatures and short descriptions;
\texttt{doc <module>.<function>} displays full documentation, and
\texttt{call <module>.<function> [args ...]} invokes an interface.
The prompt also directs the agent to read the source code when the
documentation is insufficient. These files expose the repository's
interfaces and usage knowledge without changing the harness's tool set.

\paragraph{Seed repository.}
\label{sec:appendix-seed-repository}
All three evolution runs in this paper, on OSWorld V2 with and without a
scratch VM and on ScienceBoard, start from the same seed repository. It has
about 550 lines of Python and shell and offers 15 basic interfaces in four
modules (Table~\ref{tab:seed-repository}). They wrap common desktop tools and
the Chrome DevTools endpoint, and none is specific to a benchmark
application. A setup script installs the tools these interfaces rely on, and
a command-line entry point provides the \texttt{capabilities}, \texttt{doc}
and \texttt{call} subcommands described above. Every interface, module and
usage note in an evolved repository beyond these was added during evolution.

\begin{table}[!htbp]
  \centering
  \small
  \setlength{\tabcolsep}{4pt}
  \renewcommand{\arraystretch}{1.15}
  \caption{Contents of the seed repository shared by all evolution runs.}
  \label{tab:seed-repository}
  \begin{tabular}{@{}>{\raggedright\arraybackslash}p{0.22\linewidth}>{\raggedright\arraybackslash}p{0.36\linewidth}>{\raggedright\arraybackslash}p{0.36\linewidth}@{}}
    \toprule
    File & Interfaces & What they do \\
    \midrule
    \texttt{gui} & \texttt{take\_screenshot}, \texttt{click\_at},
      \texttt{key\_press}, \texttt{get\_active\_window},
      \texttt{call\_grounding\_model}
      & Capture the screen, click a point, send keys, read the focused
        window title, and locate a described element on a screenshot \\
    \texttt{atree} & \texttt{get\_accessibility\_tree},
      \texttt{get\_window\_list}, \texttt{get\_focused\_element}
      & Read the accessibility tree, list open windows, and get the role of
        the focused element \\
    \texttt{browser} & \texttt{get\_chrome\_targets},
      \texttt{chrome\_active\_tab}, \texttt{open\_new\_tab}
      & List Chrome tabs, get the active tab, and open a URL in a new tab \\
    \texttt{cli\_ops} & \texttt{gsettings\_get}, \texttt{gsettings\_set},
      \texttt{get\_ooxml\_structure}, \texttt{get\_ooxml\_text}
      & Read and write GNOME settings, and list the parts and extract the
        text of Office files \\
    \midrule
    \texttt{cli.py} & \texttt{capabilities}, \texttt{doc}, \texttt{call}
      & List interfaces, show the documentation of one, and call one \\
    \texttt{setup.sh} & ---
      & Install \texttt{scrot}, \texttt{xdotool}, \texttt{xclip},
        \texttt{curl} and \texttt{wmctrl} \\
    \texttt{context\_\allowbreak compaction.json} & ---
      & Context-compaction settings for the task agent \\
    \bottomrule
  \end{tabular}
\end{table}

% \paragraph{API.}
\paragraph{Model Service.}
We access Qwen3.8-27B through
\href{https://openrouter.ai/provider/parasail}{Parasail} provider.
We set temperature to 1, top-$p$ to 0.9, and thinking effort to Medium,
with a 131,072-token context window and a 32,000-token output limit.
The same model is used for the task agent, GUI grounding, and the interface
builder. We estimate task-solving costs using Parasail's rates of \$0.24 per million
uncached input tokens, \$2.20 per million output tokens, and \$0.05 per
million cached input tokens. The cost estimates assume an 80\% input-cache hit rate.
The Qwen3.8-27B cost estimates in Table~\ref{tab:s4-main-results} use these
rates. These estimates cover task-solving model calls;
repository-evolution and runtime infrastructure costs are excluded.

\paragraph{Cloud infrastructure.}
For OSWorld V2, each task episode runs in a freshly initialized virtual
machine on AWS using the same benchmark release image, with a screen
resolution of $1920\times1080$ pixels. The virtual machine is terminated
after the episode. Model inference is accessed through the API described
above.

\paragraph{Runtime settings.}
Our Prime Agent task rollouts use a budget of 300 completed assistant
responses per task, with a timeout of 720 seconds for each model API call.
A response may invoke code that performs multiple GUI or application actions.
New OSWorld V2 training and validation rollouts have a two-hour wall-clock
limit. OSWorld V2 full evaluation retains the API-call timeout and
300-step budget without a per-task wall-clock limit. ScienceBoard uses
a 60-minute wall-clock limit in training, initial validation, validation
re-evaluations, and full evaluation. The meta-agent has a separate
two-hour wall-clock limit for each repository construction. Infrastructure
failures may trigger recovery runs.

% \clearpage
% \section{Evolution Lineage}
\section{Experimental Analysis}
\label{sec:appendix-experimental-analysis}

\subsection{Evolution Lineage}
\label{sec:appendix-evolution-lineage}

\begin{figure}[!htbp]
  \centering
  \includegraphics[width=0.7\linewidth]{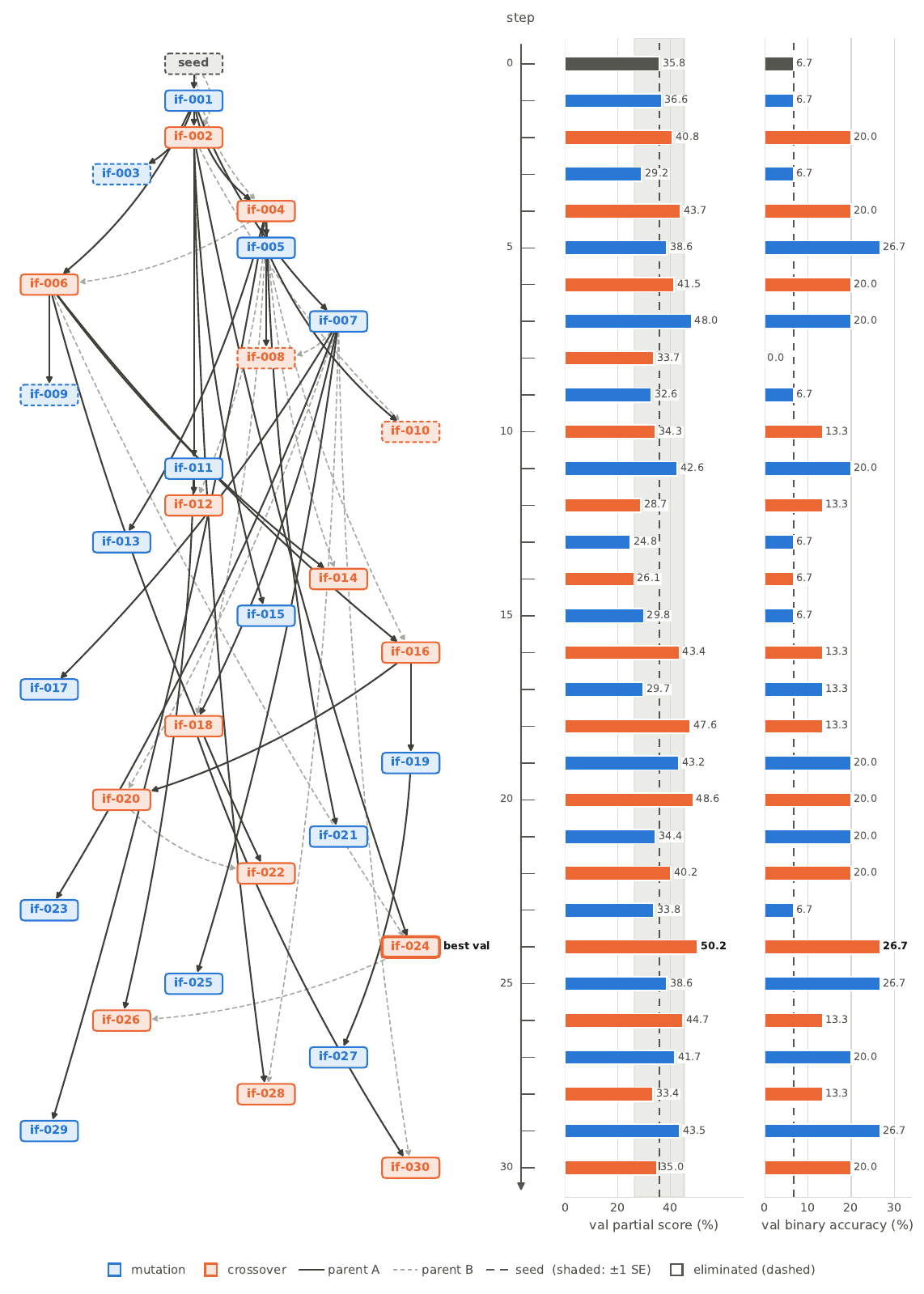}
  \caption{\textbf{Useful repositories can emerge by returning to an earlier branch.}
  OSWorld V2 lineage over 30 construction steps with scratch-VM access.
  Blue nodes denote mutation and orange nodes crossover; solid and dashed
  edges identify the first and second parents. Dashed node outlines mark
  eliminated variants. The bars report mean partial score and binary
  success on 15 validation tasks. Step~24 combines steps~2 and~6 and reaches
  the highest partial score ($50.2\%$), while tying the best binary success
  ($26.7\%$). Vertical dashed lines mark the seed scores; the shaded band
  uses the $\pm1$ standard error estimated from the same validation set and
  model in the run without scratch-VM access.}
  \label{fig:s4-evolution-lineage}
\end{figure}

Figure~\ref{fig:s4-evolution-lineage} shows the repository lineage for the
30-step OSWorld V2 population run with a builder scratch VM. Starting
from the seed, construction alternates
between mutation of one parent and crossover of two distinct parents.
Binary tournaments based on mean partial validation score select parents
from the eligible population; after step 10, the five lowest-scoring
variants lose parent eligibility. The resulting graph branches from
earlier repositories and combines branches through crossover, allowing
prior variants to contribute beyond the next construction step.
Validation performance does not improve monotonically. At step~24,
crossover combines repositories from steps~2 and~6 to obtain the highest
mean partial score, $50.2\%$, compared with $35.8\%$ for the seed. Its
binary success rate is $26.7\%$, tying steps~5, 25, and~29, compared with
$6.7\%$ for the seed. Thus, earlier branches remain useful well beyond
their construction step, and the two validation metrics distinguish
different aspects of repository performance.

\begin{figure}[!htbp]
  \centering
  \includegraphics[width=0.7\linewidth]{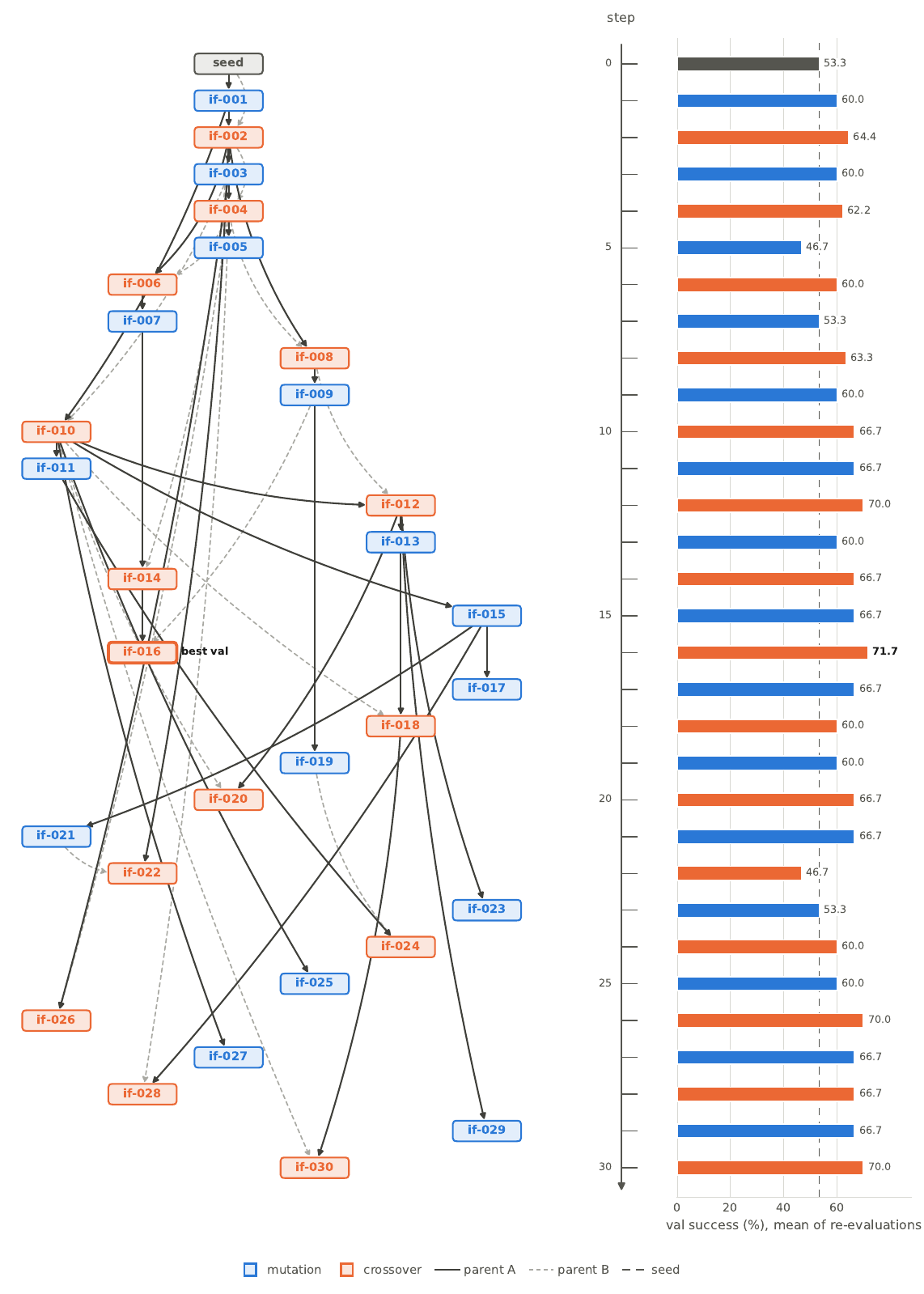}
  \caption{\textbf{ScienceBoard evolution combines branches under repeated validation.}
  Repository lineage over 30 construction steps with a 60-minute episode
  limit. Colors and edge styles follow Figure~\ref{fig:s4-evolution-lineage}.
  Bars show each repository's mean success across its available evaluations
  on 15 validation tasks, including repeats where available; they need not
  equal the scores available when the repository was selected as a parent.
  The dashed line marks the seed's initial $53.3\%$ success. The crossover
  at step~16 has the highest mean, $71.7\%$ over four evaluations.}
  \label{fig:s4-scienceboard-evolution-lineage}
\end{figure}

Figure~\ref{fig:s4-scienceboard-evolution-lineage} shows the corresponding
30-step ScienceBoard run with a 60-minute episode limit. Parent selection
uses validation success averaged over the initial evaluation and any
repeated evaluations, with denoising every five construction steps as
described in Section~\ref{sec:results-denoising}. The highest mean belongs
to \texttt{if-016}, a crossover of \texttt{if-014} and \texttt{if-009}:
it achieves $71.7\%$ success over four evaluations of the 15 validation
tasks, compared with the seed's initial $53.3\%$. Its parents have mean
scores of $66.7\%$ and $60.0\%$, respectively. Later variants continue to
draw on earlier branches, but none exceeds \texttt{if-016}'s mean within
the 30 steps. The lineage illustrates how crossover can combine
branches with different validation histories while repeated evaluation
informs which repositories remain promising.

\subsection{Execution Efficiency}
\label{sec:appendix-execution-efficiency}

% Figure~\ref{fig:s4-execution-efficiency} compares the seed and evolved
Table~\ref{tab:s4-execution-efficiency} compares the seed and evolved
repository across 108 OSWorld V2 tasks. On each repository's full-credit
tasks, the evolved agent uses $7.1\%$ fewer tool calls on average while
receiving full credit on more tasks. Together with the CDP case in
Appendix~\ref{sec:appendix-cdp-case}, this suggests a useful role for helper
functions: reusable operations can reduce repeated setup and combine
multiple low-level interactions within a single tool call. Such savings
can free steps for task-specific reasoning and execution.

% Previous efficiency figure, replaced by a table.
% \begin{figure}[!htbp]
%   \centering
%   \includegraphics[width=\linewidth]{assets/agent_s4_execution_efficiency.pdf}
%   \caption{\textbf{The evolved repository uses fewer calls on its full-credit
%   tasks, while the all-task average remains similar.} Mean executed IPython
%   tool calls for the seed and \texttt{if-011} across 108 tasks (left) and
%   each repository's full-credit subset (right). Full credit means a task
%   score of 1.0; these subsets contain different tasks. A call may contain
%   multiple actions, and failed calls are included. Counts use the final
%   retained attempt per task, including the interrupted \texttt{if-011}
%   attempt on task 063, and exclude earlier infrastructure retries.}
%   \label{fig:s4-execution-efficiency}
% \end{figure}

\begin{table}[!htbp]
  \centering
  \small
  \caption{\textbf{The evolved repository uses fewer calls on its full-credit
  tasks, while the all-task average remains similar.} Mean executed IPython
  tool calls across 108 tasks and each repository's full-credit subset.
  Parentheses give task counts. Full credit means a score of 1.0; the
  subsets contain different tasks. Counts include failed calls and use
  the final retained attempt, including the interrupted evolved attempt
  on task 063, but exclude earlier infrastructure retries.}
  \label{tab:s4-execution-efficiency}
  \begin{tabular*}{\linewidth}{@{\extracolsep{\fill}}lrr@{}}
    \toprule
    Task subset & Seed & Evolved (\texttt{if-011}) \\
    \midrule
    All tasks & 163.2 ($n=108$) & 162.1 ($n=108$) \\
    Each repository's full-credit tasks & 111.6 ($n=21$) & 103.7 ($n=23$) \\
    \bottomrule
  \end{tabular*}
\end{table}

The aggregate comparison also shows why helper-level improvements need
not shorten every trajectory. Across all tasks, mean call counts remain
nearly unchanged. The full-credit subsets differ, and on the 11 tasks
where both repositories receive full credit, the mean rises from 87.8 to
103.2 calls. Excluding the interrupted evolved attempt likewise leaves
similar all-task means (163.2 for the seed and 163.4 for the evolved
repository). Thus, the shorter full-credit average describes each
repository's successful trajectories rather than a uniform speedup on
matched tasks. The CDP trace supplies concrete evidence of reduced setup
and bulk-reading overhead; the aggregate results show that these local
savings coexist with other sources of task-level variation.

% Source: figures/execution_efficiency/07_seed_vs_if011_avg_steps_test108.csv
% and its companion episode CSV and methodology note, copied from
% simular_gpu2:/storage/toby/shadow-interface-evolution/docs/paper/noscratch15/.
% The 108-task set includes 15 evolution-validation tasks and 93 others.

% \clearpage
% \subsection{CDP Browser Interface}
\subsection{Examples of Evolved Interfaces}
\label{sec:appendix-interface-examples}

\begin{figure}[!htbp]
  \centering
  \includegraphics[width=\linewidth]{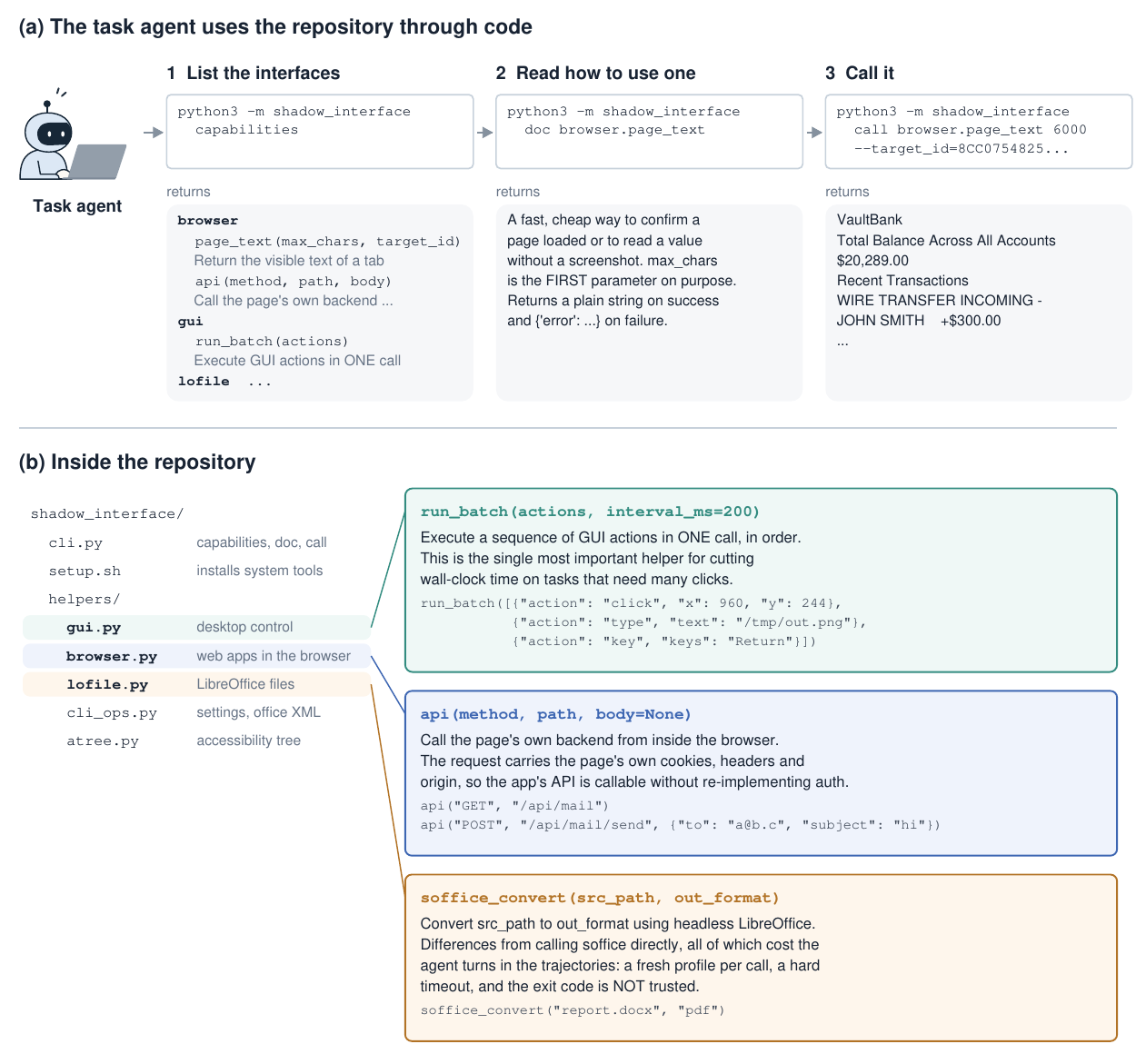}
  \caption{How the task agent uses an evolved repository. \textbf{(a)} On an
  OSWorld V2 task, the task agent lists the available interfaces, reads the
  usage notes of one and calls it from its code. Outputs are shortened.
  \textbf{(b)} The repository selected without a scratch VM
  (Table~\ref{tab:s4-main-results}) is a Python package. Each interface is a
  module of functions whose docstrings say how and when to use them, including
  lessons from earlier trajectories. Docstrings are shortened.
  \texttt{run\_batch}, \texttt{api} and the \texttt{lofile} module were added
  during evolution.}
  \label{fig:repo-usage}
\end{figure}

Figure~\ref{fig:repo-usage} shows how the task agent uses an evolved
repository during a task and what its interfaces contain.
Table~\ref{tab:interface-examples} lists interfaces from the evolved
repositories described in Section~\ref{sec:results-analysis}, with what each
lets the agent observe and do.

\begin{table}[!htbp]
  \centering
  \small
  \setlength{\tabcolsep}{4pt}
  \renewcommand{\arraystretch}{1.15}
  \caption{Examples of interfaces created or extended by evolution. Each
  changes what the agent observes about part of the computer and what the
  agent can do to it in one call. The desktop-control and web-app interfaces
  begin in the seed with basic operations and are extended during evolution,
  and the others are new. A fixed-interface agent observes all
  of these through screenshots and acts on them through clicks and
  keystrokes.}
  \label{tab:interface-examples}
  \begin{tabular}{@{}p{0.24\linewidth}p{0.35\linewidth}p{0.35\linewidth}@{}}
    \toprule
    Interface & What the agent observes & What the agent can do \\
    \midrule
    Desktop control and background jobs
      & Windows and their accessible controls; the status and logs of
        programs running in the background
      & Run a list of GUI actions and get back how many ran and any error; launch,
        monitor and wait for long jobs such as video renders \\
    Web-app access in the browser
      & The text and data of pages in the user's logged-in session,
        rather than their pixels
      & Run code inside a page; send requests from the logged-in tab, reused
        across email, banking and chat sites \\
    Office-document editing (LibreOffice)
      & Cells, formulas, paragraphs, slides and shapes of an open document
      & Write a block of cells at once; edit document objects; export to PDF \\
    Document reading without the application
      & Text, metadata and rendered pages of a file, without opening the
        application that made it
      & Extract and search content; check whether an exported PDF is
        complete \\
    Score editing (MusicXML)
      & Parts, measures and notes with pitch and duration
      & Add notes and rests; change pitch and duration; add dynamics;
        validate the score \\
    Molecular-viewer control (ChimeraX)
      & Which atoms and residues a selection or command affects
      & Send viewer commands directly, such as coloring one residue of one
        chain \\
    Proof checking and completion (Lean)
      & Whether a proof checks and compiles
      & Check, compile, update the editor and report whether the proof is
        finished \\
    \bottomrule
  \end{tabular}
\end{table}

\subsection{Additional Case Study}
\label{sec:appendix-cdp-case}

We examine how a reusable Chrome DevTools Protocol (CDP) interface changes
browser interaction in a client photo-delivery task. The agent must read
instructions in TeamChat, reconcile invoices with payments in VaultBank,
and prepare and email either finished photos or watermarked previews.
We compare the seed repository with evolved variant \texttt{if-011} on
the same task. The seed provides browser target discovery but lacks a
page-evaluation helper, whereas the evolved repository exposes
\texttt{browser.page\_eval} and \texttt{browser.page\_text} for executing
JavaScript and retrieving page text through CDP.

\begin{figure}[!htbp]
  \centering
  \includegraphics[trim=70bp 365bp 0bp 135bp,clip,width=\linewidth]{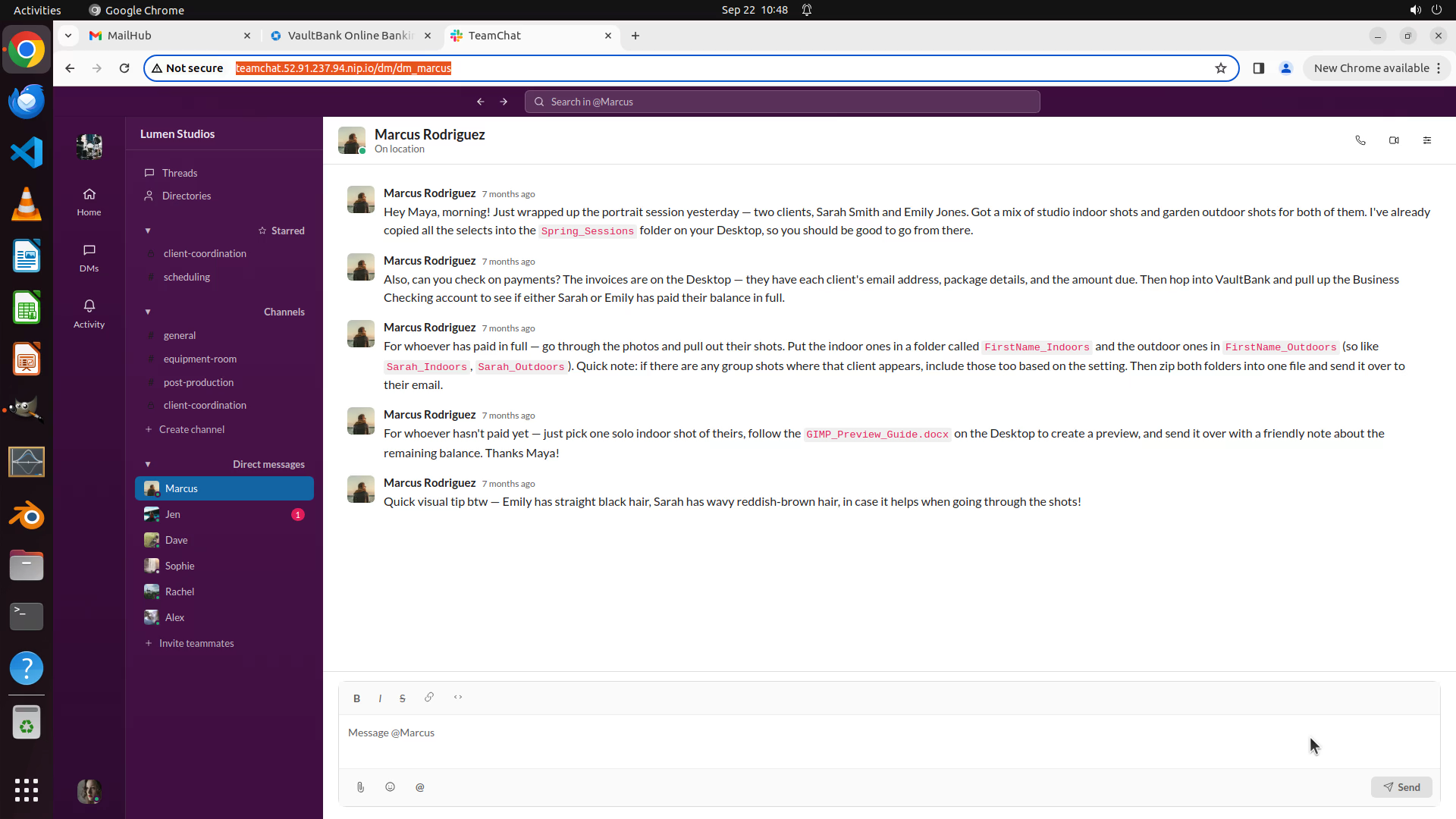}
  \vspace{3pt}
  \begin{tcolorbox}[colback=black!2,colframe=black!35,boxrule=0.4pt,
    arc=1mm,left=6pt,right=6pt,top=4pt,bottom=4pt]
    \small
    \textbf{Evolved trajectory, tool cell 11:}\par
    \texttt{browser.page\_eval('location.href="/dm/dm\_marcus"; "ok"')}\par
    \texttt{browser.page\_text(8000)}
  \end{tcolorbox}
  \caption{\textbf{Reusable browser helpers expose the instructions needed
  for the rest of the task.} Cropped frame from the evolved task-046
  recording showing the TeamChat direct message. It asks the agent to
  check payments and deliver photos or previews accordingly. The box
  abbreviates the helper names and arguments passed through the CLI in
  tool cell 11. The frame illustrates the page content; operation timing
  is measured from the trajectory, independently of the recording.}
  \label{fig:s4-cdp-screenshot}
\end{figure}

Figure~\ref{fig:s4-cdp-screenshot} shows the requested conversation and
the helper operations used to retrieve its text.

\paragraph{Reusing browser access.}
The seed agent implements its own CDP client during the task, encountering
missing dependencies, an incorrect target identifier, and connection and
message-handling failures before obtaining the requested direct message.
The evolved interface encapsulates target selection and WebSocket
communication, allowing the agent to navigate and read the message using
existing helpers. The same message is retrieved by tool cell 11 with the
evolved repository, compared with cell 37 for the seed. This illustrates
how repository evolution can move browser setup and debugging out of the
task trajectory and into reusable code.

\begin{figure}[!htbp]
  \centering
  \includegraphics[width=\linewidth]{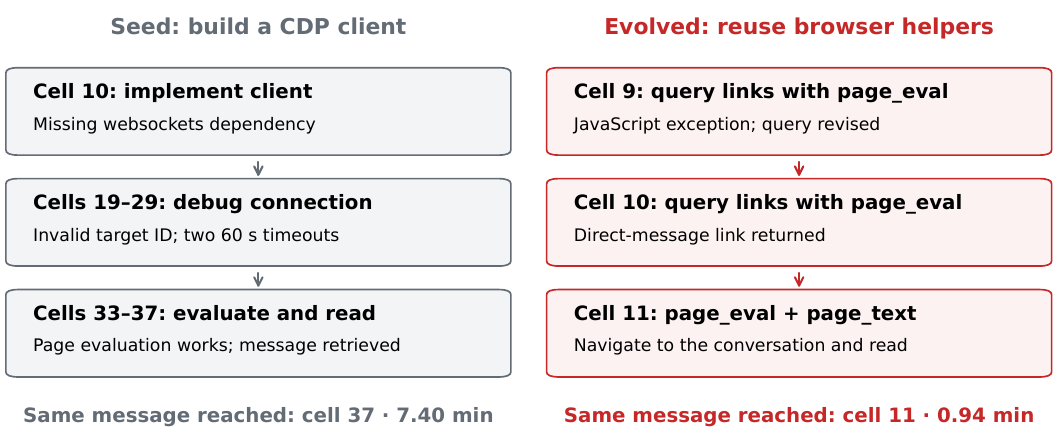}
  \caption{\textbf{Reusing CDP access reduces the setup needed to read the
  same message.} Selected milestones from task 046 compare the seed's
  client implementation and debugging with the evolved repository's
  existing browser helpers. Both retrieve the requested direct message,
  at tool cells 37 and 11 respectively. Times run from the initial task
  message to the tool result containing that message. Boxes summarize
  recorded events and are not proportional to elapsed time; the evolved
  agent also encounters and corrects an initial JavaScript error.}
  \label{fig:s4-cdp-comparison}
\end{figure}

Figure~\ref{fig:s4-cdp-comparison} contrasts the selected milestones
leading to the same message in the two trajectories.

% \clearpage

\paragraph{Reading structured page content.}
The evolved agent also uses JavaScript to load additional bank transactions
and extract 81 table rows from the page DOM. It then aggregates payments
by client name in Python. Returning page text and structured records lets
the agent process multiple entries in one operation, reducing the need for
repeated visual inspection and transcription. The resulting interface
supports both navigation and bulk reading through the same browser session.

\begin{table}[!htbp]
  \centering
  \small
  \caption{Observed trajectories for the same client photo-delivery task
  (task 046). Tool cells count executed IPython calls; image-bearing results
  count cells returning images. Recorded duration spans the initial task
  message to the last recorded message. Each column represents one trajectory.}
  \label{tab:s4-cdp-case}
  \begin{tabular*}{\linewidth}{@{\extracolsep{\fill}}lrr@{}}
    \toprule
    Measure & Seed & Evolved (\texttt{if-011}) \\
    \midrule
    Executed tool cells & 299 & 167 \\
    Image-bearing tool results & 68 & 14 \\
    Recorded duration (min) & 97.5 & 51.1 \\
    Task score & 0.50 & 1.00 \\
    \bottomrule
  \end{tabular*}
\end{table}

Table~\ref{tab:s4-cdp-case} shows a shorter trajectory with fewer tool calls
and image-bearing results, alongside a higher task score. The clearest
mechanism is reduced browser setup overhead and direct access to page
content. Both agents eventually use CDP, so the benefit illustrated here
is making that capability readily reusable. The full task also includes
file, image, and email operations, and the repositories differ in other
interfaces; the overall duration difference is therefore an observed
case outcome rather than an isolated estimate of CDP's effect.

\subsection{Tasks With and Without Evolved Interfaces}
\label{sec:appendix-interface-cases}

\begin{figure}[!htbp]
  \centering
  \includegraphics[width=\linewidth]{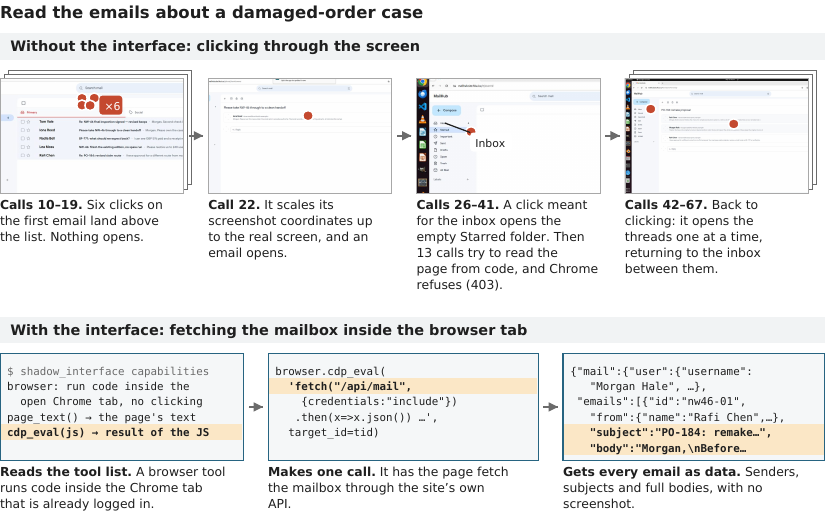}
  \caption{\textbf{Reading a mailbox in MailHub}
  (\hyperref[task:s-010]{\texttt{s\_010}}). The seed agent reads the case
  emails through the GUI in 60 calls (8--67). Its first six clicks land above
  the email list because it uses coordinates from its 1200-pixel screenshots as
  screen coordinates. A click meant for the inbox opens Starred, and its attempt
  to read the page from code is refused with \texttt{403 Forbidden}. With
  \texttt{if-011}, \texttt{page\_text} lists the inbox and one
  \texttt{cdp\_eval} call fetches the mailbox with full email bodies inside the
  logged-in tab, in 12 calls and without a screenshot.}
  \label{fig:case-mailhub}
\end{figure}

\begin{figure}[!htbp]
  \centering
  \includegraphics[width=\linewidth]{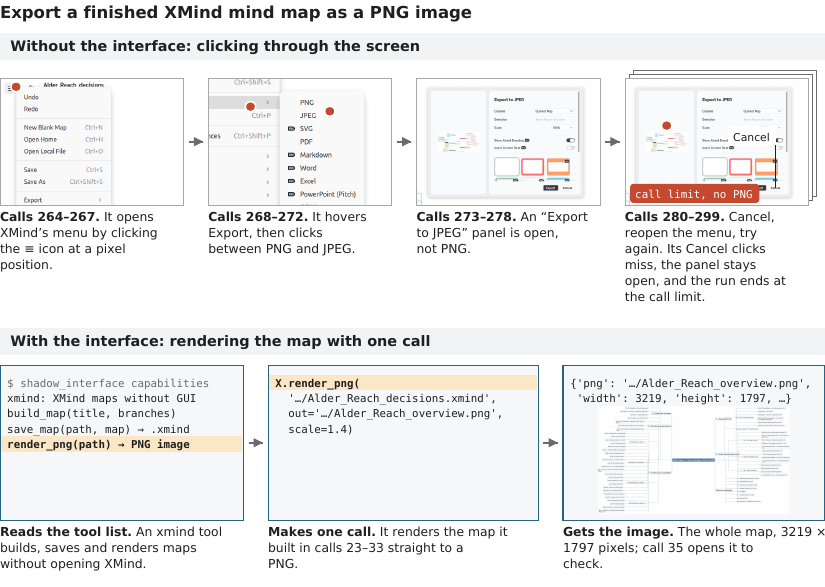}
  \caption{\textbf{Exporting an XMind map}
  (\hyperref[task:s-005]{\texttt{s\_005}}). The seed agent tries to export
  a PNG through XMind's menus in calls 249--299. Its click on the Export
  submenu lands between PNG and JPEG, and an ``Export to JPEG'' panel opens.
  Its clicks on Cancel miss, and the episode ends at the call limit without an
  image. With \texttt{if-029} from a separate evolution run, the agent builds
  the map with \texttt{xmind.build\_map} and \texttt{save\_map}, and one call
  to \texttt{xmind.render\_png} writes the image.}
  \label{fig:case-xmind}
\end{figure}

\begin{figure}[!htbp]
  \centering
  \includegraphics[width=\linewidth]{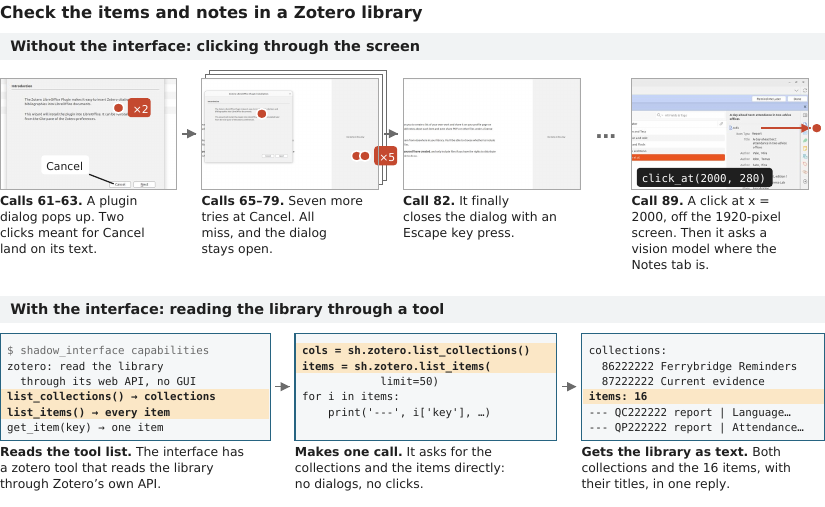}
  \caption{\textbf{Reading a Zotero library}
  (\hyperref[task:s-007]{\texttt{s\_007}}). The seed agent checks items and
  notes through the GUI in 42 calls (59--100). Nine clicks meant for the Cancel
  button of a plugin dialog miss (calls 61--79), and the agent closes the
  dialog with an Escape key press at call 82. Later it clicks outside the
  screen and asks a vision model to locate the Notes tab. With \texttt{if-029}
  from a separate evolution run, \texttt{zotero.list\_collections} and
  \texttt{list\_items} return the library as data in one call. The interface
  covers reading only: on this build the API rejects writes.}
  \label{fig:case-zotero}
\end{figure}

\begin{figure}[!htbp]
  \centering
  \includegraphics[width=\linewidth]{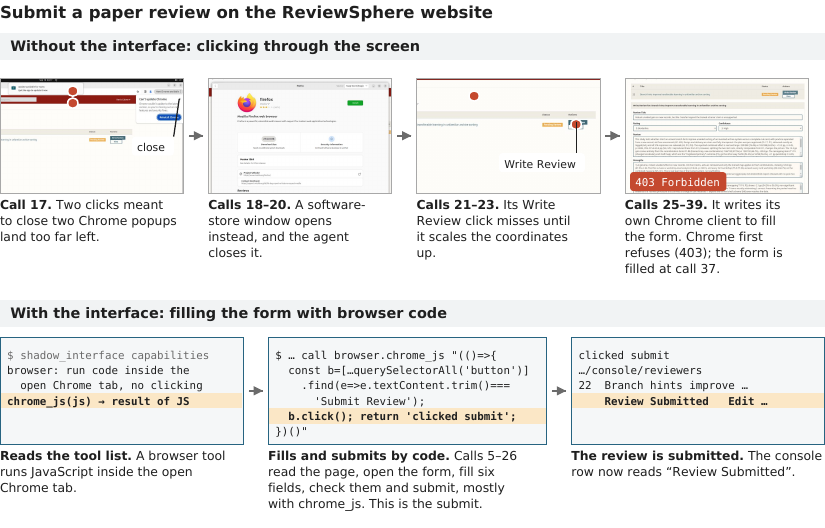}
  \caption{\textbf{Submitting a review in ReviewSphere}
  (\hyperref[task:s-022]{\texttt{s\_022}}). The seed agent's clicks meant
  to close two popups open a software-store window, and its click on Write
  Review misses until it rescales the coordinates. It then writes its own
  Chrome client to fill the form; Chrome first refuses the connection with
  \texttt{403 Forbidden}, and the form is filled at call 37. With
  \texttt{if-026} from a separate evolution run, calls 5--26 read the page,
  open the form, fill six fields, check them, and submit, mostly through
  \texttt{browser.chrome\_js}, with no failed calls.}
  \label{fig:case-reviewsphere}
\end{figure}

\begin{figure}[!htbp]
  \centering
  \includegraphics[width=\linewidth]{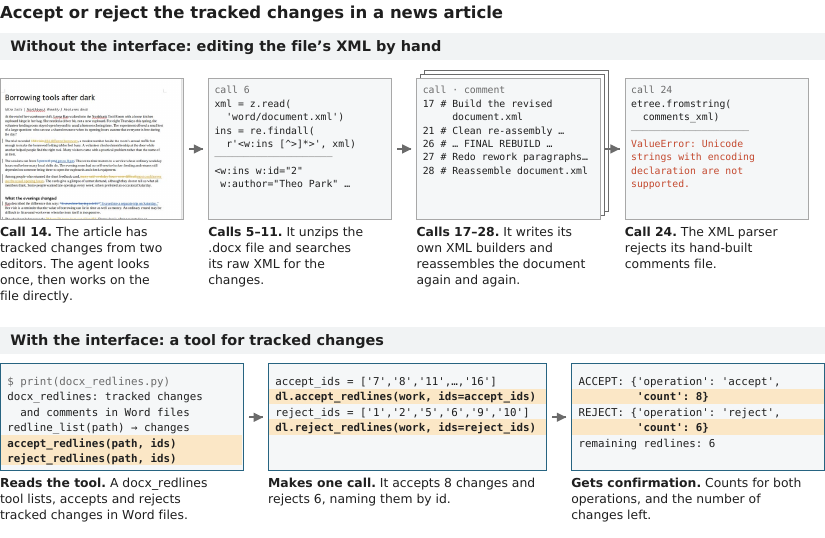}
  \caption{\textbf{Resolving tracked changes in a Word document}
  (\hyperref[task:s-026]{\texttt{s\_026}}). The seed agent reads the raw
  document XML and rebuilds insertion and deletion elements by hand in calls
  17--28; its comments XML fails to parse at call 24. With \texttt{if-030}
  from a separate evolution run, the agent lists the changes with
  \texttt{redline\_list}, then accepts eight and rejects six with
  \texttt{accept\_redlines} and \texttt{reject\_redlines} in one call.}
  \label{fig:case-redlines}
\end{figure}

\paragraph{OSWorld V2 evolution comparison.}
Figures~\ref{fig:case-mailhub}--\ref{fig:case-redlines} compare the seed
repository with an evolved repository on the same task. The top row of each
figure follows the seed episode through the agent's own screenshots, in order.
Red dots mark clicks at the coordinates in the agent's code, drawn on a
screenshot from the calls named under each panel. A stack of frames marks repeated attempts.
In the Word case, where the seed agent works almost entirely in code, the later
panels quote its code and output. The bottom row shows the evolved episode: the
interface documentation the agent read, the call it made, and the output it
received. The call and output are quoted from the trajectory, shortened where
marked with an ellipsis. The documentation is condensed: function names and
signatures are exact, the descriptions paraphrase the docstrings the agent
printed, and the functions the agent used are highlighted.

\begin{figure}[!htbp]
  \centering
  \includegraphics[width=\linewidth]{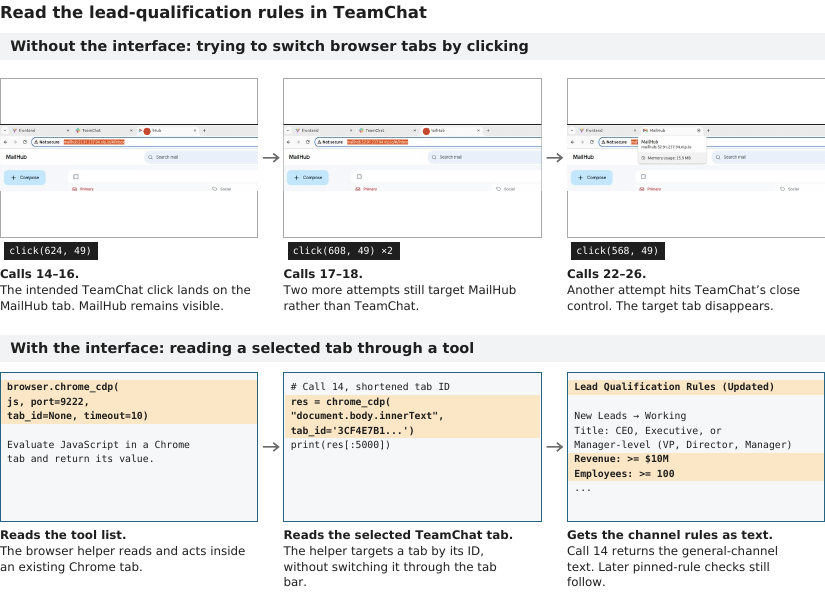}
  \caption{\textbf{Accessing a TeamChat tab}
  (OSWorld V2 task \texttt{038}). The bare agent repeatedly clicks the
  MailHub tab while trying to activate TeamChat, then hits TeamChat's close
  control. With \texttt{if-024}, call 14 uses \texttt{browser.chrome\_cdp}
  with a tab ID to read the general-channel rules without switching tabs
  through the screen. Later pinned-rule checks and CRM updates remain.
  The full episodes take 94 and 95 model steps and receive task scores of
  0.6517 and 0.70, respectively.}
  \label{fig:case-osworld-teamchat}
\end{figure}

The episodes were selected to illustrate these mechanisms and are not
controlled comparisons. Several evolved repositories come from separate
evolution runs with the same tasks and model. Call counts exclude the cost of
building the interface, and the synthetic \texttt{s\_*} tasks are ungraded.
Figure~\ref{fig:case-osworld-teamchat} additionally compares the bare agent
with \texttt{if-024} on OSWorld V2 task \texttt{038}.

% Source: ../case_osworld_if024_bare_steps/visual_rescreen/ROUND_2.md.
% Screenshot/call provenance and independently verified episode metrics:
% ../case_osworld_if024_bare_steps/visual_rescreen/038/paper_style_evidence.json.

\FloatBarrier

\begin{figure}[!htbp]
  \centering
  \includegraphics[width=\linewidth]{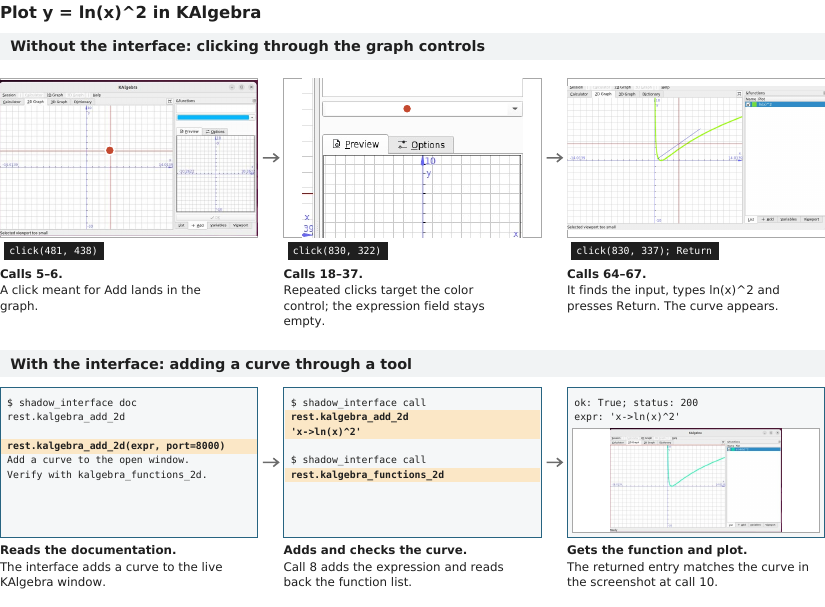}
  \caption{\textbf{Plotting a function in KAlgebra}
  (\texttt{KAlgebra:A-02}). The seed agent's Add click lands in the graph,
  and repeated clicks target the color control instead of the expression
  input. It eventually finds the input and plots $y=\ln(x)^2$ in
  calls 64--67. With \texttt{if-016}, call 8 uses
  \texttt{rest.kalgebra\_add\_2d} to add the curve and
  \texttt{rest.kalgebra\_functions\_2d} to read it back; call 10 shows
  the resulting plot. Both episodes pass, in 68 and 11 model steps,
  respectively.}
  \label{fig:case-scienceboard-plot}
\end{figure}

\begin{figure}[!htbp]
  \centering
  \includegraphics[width=\linewidth]{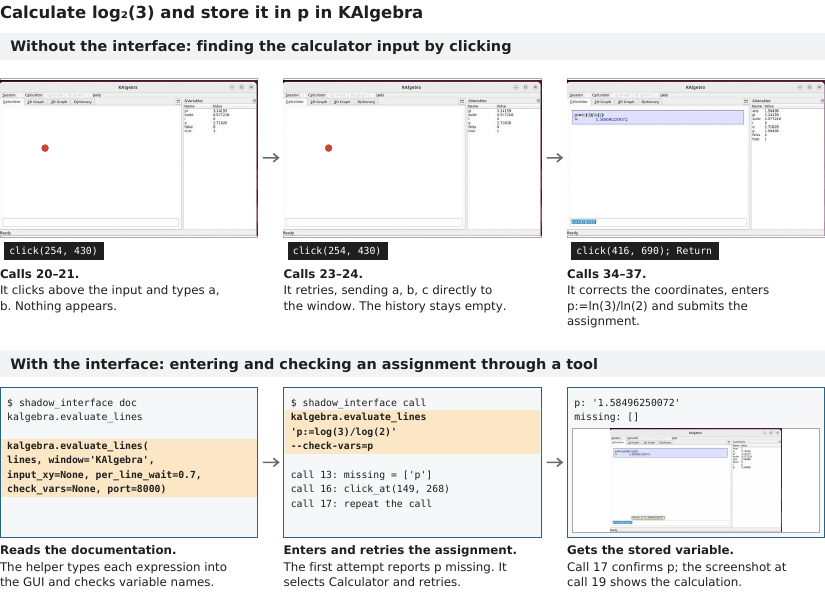}
  \caption{\textbf{Storing a calculation in KAlgebra}
  (\texttt{KAlgebra:C-01}). The seed agent repeatedly clicks above the
  calculator input and sends test characters before correcting the
  coordinates and storing $\log_2(3)$ in \texttt{p}. With \texttt{if-016},
  \texttt{kalgebra.evaluate\_lines} enters the expression through the GUI
  and checks the variable. Its first attempt reports \texttt{p} missing;
  the agent selects the Calculator tab and retries, confirming
  \texttt{p=1.58496250072} at call 17. Both episodes pass, in 40 and
  21 model steps, respectively.}
  \label{fig:case-scienceboard-log}
\end{figure}

\begin{figure}[!htbp]
  \centering
  \includegraphics[width=\linewidth]{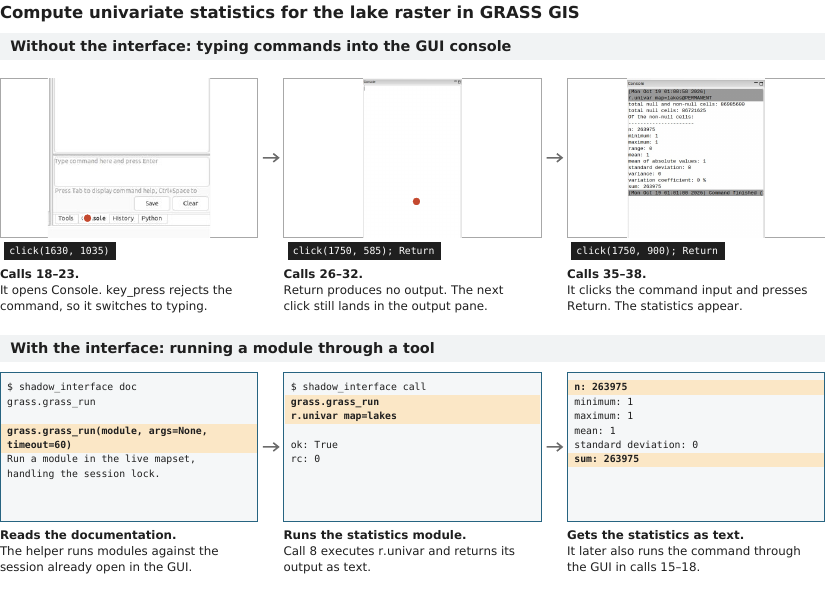}
  \caption{\textbf{Computing raster statistics in GRASS GIS}
  (\texttt{GrassGIS:B-12}). The seed agent locates the GUI console,
  corrects its typing method, and retries Return until it focuses the
  command input and obtains the statistics. With \texttt{if-016}, call 8
  runs \texttt{r.univar map=lakes} through \texttt{grass.grass\_run}
  against the live mapset and returns the statistics as text, including
  \texttt{n=263975} and \texttt{sum=263975}. The evolved agent also runs
  the command through the GUI in calls 15--18. Both episodes pass, in
  40 and 21 model steps, respectively.}
  \label{fig:case-scienceboard-grass}
\end{figure}

\paragraph{ScienceBoard evolution comparison.}
Figures~\ref{fig:case-scienceboard-plot}--\ref{fig:case-scienceboard-grass}
compare the seed repository with the selected evolved repository,
\texttt{if-016}, on the same ScienceBoard tasks using the same task model
and evaluation harness. The top row (``Without the interface'') follows
GUI interactions in the seed episode; the bottom row (``With the interface'')
shows the evolved helper documentation, calls, and returned observations.
Both episodes pass in each case. Model steps count completed assistant
responses, including retries, failed attempts, verification, and the final
answer; one step may contain multiple actions. Each condition has one episode
per task. These qualitative cases illustrate interface use rather than
estimating a treatment effect or isolating an individual helper's benefit.

% Source: ../case_scienceboard_visual_rescreen/README.md.
% Figure generator: ../case_scienceboard_visual_rescreen/draw_paper_style.py.
% Original call text, screenshot provenance, and independently verified steps:
% ../case_scienceboard_visual_rescreen/sources/*/figure_evidence.json,
% ../case_scienceboard_visual_rescreen/verification.json.

\FloatBarrier

\section{Harness Prompts}
\label{sec:appendix-harness-prompts}

We reproduce the main builder prompts used for repository evolution.
Prompt~\ref{prompt:builder-synthetic} provides the standing instructions for
mutation from ungraded synthetic-task trajectories, and
Prompt~\ref{prompt:builder-mutation-turn} starts each mutation step.
Prompt~\ref{prompt:builder-crossover} describes
how to combine two parent repositories, and Prompt~\ref{prompt:builder-system}
specifies the builder's execution protocol. These prompts are fixed by the
harness rather than evolved with the repository.

\begin{harnessprompt}{Mutation prompt: synthetic training tasks}{prompt:builder-synthetic}
# Shadow Interface Builder

You improve software that helps AI agents complete computer tasks on Ubuntu 22.04.

## Evaluation evidence: synthetic tasks without verifiers

These 30 synthetic training tasks (s_001 through s_030) have NO verifiers.
Their dummy evaluators always return 0.0. Zero does NOT mean failure, and
scores/averages do NOT measure task success. Never optimize those numbers,
use them as rewards, or classify all attempts as failures. Raw result.txt
and diagnostic score fields are compatibility placeholders only.

Read the TRAIN trajectories under eval_results/ recursively. Each task folder
contains instruction.txt, claude_trajectory.jsonl (a historical filename also
used for Qwen/Prime traces), and any available diagnostics/screenshots.
The aggregate latest_summary.json and validation trajectories are intentionally
not supplied to the builder. Do not try to access validation data.

Use instructions, actions, tool outputs, screenshots, errors and observed
artifacts to identify repeated work, missing capabilities and fragile workflows.
An agent's claim that it completed a task is not verified success. Distinguish
observed evidence from hypotheses, and record both in notebook.jsonl. Do not
invent a verifier or labels. Improve the evolving software from trajectories.

## What you can modify

- `shadow_interface/` — all your code goes here
- `shadow_interface/context_compaction.json` — evolvable mini-SWE context-window, reserve, keep-recent, and checkpoint-summary policy
- `notebook.jsonl` — append observations (persists across mutations)

## Constraints

- Python stdlib only — pip is disabled on VMs
- VMs are fresh — no state persists between tasks
- Task agents are autonomous — they may or may not use your tools
- Do NOT modify files outside shadow_interface/ and notebook.jsonl
- Do NOT create or modify SYSTEM_PROMPT.md — the task agent's system prompt is controlled by the orchestrator
- You do NOT run evals — the orchestrator evaluates after you finish

## Goal

Build general, reusable tools that help agents complete tasks more reliably and efficiently, using trajectory evidence. The orchestrator separately measures each version on the older validation set; synthetic scores remain ungraded.

Two qualities to keep in mind as you build:

- **Generality.** Prefer helpers that work across many tasks. A helper that only addresses one specific task is a sign you stopped one level too early — keep abstracting until the same helper would serve the next, related task.
- **Discoverability.** Write clear signatures and docstrings. Agents find your tools via `python -m shadow_interface capabilities` and `python -m shadow_interface doc <name>`; helpers without docstrings get passed over for less specialized ones the agent can read.
\end{harnessprompt}

The mutation turn instruction refers the builder to the standing instructions
in \texttt{CLAUDE.md} reproduced above.

\begin{harnessprompt}{Mutation turn instruction}{prompt:builder-mutation-turn}
Mutation step. Read CLAUDE.md for instructions. Study eval_results/ trajectories to identify difficulties and opportunities. Interpret scores according to CLAUDE.md. Improve shadow_interface/. Write observations to notebook.jsonl.
\end{harnessprompt}

\begin{harnessprompt}{Builder template: crossover}{prompt:builder-crossover}
# Shadow Interface Builder — Crossover

You have two interface implementations that took different approaches.

## Parents

- **Parent A** (`crossover_parent_a/`): eval results in `eval_results/parent_a/`
- **Parent B** (`crossover_parent_b/`): eval results in `eval_results/parent_b/`

## Your Job

Combine the best of both into `shadow_interface/` (starts with Parent B's code).

1. Capabilities in A but not B — port them
2. Capabilities in B but not A — keep them
3. Different approaches to same thing — pick what works better per eval results
4. Ensure merged code is coherent

## Constraints

- Python stdlib only
- Modify only shadow_interface/ and notebook.jsonl
- You do NOT run evals
\end{harnessprompt}

\begin{harnessprompt}{Builder system prompt}{prompt:builder-system}
You are the autonomous builder inside a shadow-interface evolution run.
Follow the task exactly, inspect the available evaluation evidence, make the
requested changes directly in the allowed files, verify them, and finish
without asking the user questions.

Your only model-facing tool is a persistent IPython kernel. Run shell commands
from IPython subprocess APIs or a %%bash cell. To inspect the GUI visually,
first use a subprocess to run the shadow interface screenshot command, for
example:

    subprocess.run(["python3", "-m", "shadow_interface", "call",
                    "gui.take_screenshot", "/tmp/screenshot.png"], check=True)

Then, in a subsequent IPython cell, load the image into model context with:

    print(await attach_image("/tmp/screenshot.png"))

Do not Base64-encode screenshots or infer their visual contents from file
metadata. Prime Agent owns the agent loop, session history, and configured
context compaction; continue working through compaction when it occurs.
\end{harnessprompt}

% \clearpage
% Source: OSWorld-V2/evaluation_examples/examples_v2/tasks/s_001.json--s_030.json.
% Checked against the accepted rebuilt-30 designs and recorded training instructions
% in qwen_syn30v15_128k-20260920T013351Z. Shared runner suffix shown once below.
\section{OSWorld V2 Synthetic Training Tasks}
\label{sec:appendix-synthetic-tasks}

We use 30 synthetic tasks to provide construction experience for
\methodname{}. The set contains 21 tasks that use desktop applications and
nine that involve local mock websites, sometimes together with desktop
applications. Each task includes an instruction, input materials, and an
initial application state. The tasks cover document and spreadsheet work,
media editing, 3D modeling, research, file management, and web workflows.
Table~\ref{tab:synthetic-training-tasks} lists all task IDs, goals, and
applications. Section~\ref{sec:synthetic-task-instructions} gives their full
task-specific instructions.

\paragraph{Construction and use.}
We used \texttt{gpt-6-astra} for task authoring and
\texttt{gpt-5.6-sol} for independent task review.
Author agents created original scenarios and materials from environment and
coverage briefs. Independent reviewer agents assessed originality against the
benchmark and the accepted synthetic bank, coherence, feasibility, and the
challenge supported by the materials. This review assessed task design.
It did not label agent attempts or create reference solutions. The training
package uses a constant-zero evaluator for compatibility with the runner;
this placeholder is not a success label or a training reward. The
meta-agent receives the resulting training trajectories, while held-out
validation determines selection as described in Section~\ref{sec:splits}.

\begingroup
\small
\setlength{\tabcolsep}{4pt}
\renewcommand{\arraystretch}{1.12}
\begin{longtable}{@{}p{0.09\textwidth}p{0.56\textwidth}p{0.28\textwidth}@{}}
\caption{\textbf{Synthetic tasks provide the experience from which \methodname{} discovers interfaces.}
  The 30 tasks span desktop applications and local web workflows, often
  requiring the agent to work across applications. Their trajectories expose
  interaction failures and inefficient workflows that the meta-agent can
  study to revise interfaces and usage knowledge, without ground-truth
  success labels.}
\label{tab:synthetic-training-tasks}\\
\toprule
\textbf{ID} & \textbf{Task} & \textbf{Applications} \\
\midrule
\endfirsthead
\multicolumn{3}{l}{\textit{Table~\ref{tab:synthetic-training-tasks} continued}}\\
\toprule
\textbf{ID} & \textbf{Task} & \textbf{Applications} \\
\midrule
\endhead
\midrule
\multicolumn{3}{r}{\textit{Continued on the next page}}\\
\endfoot
\bottomrule
\endlastfoot
\hyperref[task:s-001]{\texttt{s\_001}} & Update a visitor map and event timetable. & LibreOffice Impress, LibreOffice Writer \\
\hyperref[task:s-002]{\texttt{s\_002}} & Edit a thank-you video with captions and music. & Shotcut \\
\hyperref[task:s-003]{\texttt{s\_003}} & Assign repair jobs using current staff, parts and equipment. & TeamChat, LibreOffice Calc \\
\hyperref[task:s-004]{\texttt{s\_004}} & Mix several audio tracks and control volume changes. & REAPER \\
\hyperref[task:s-005]{\texttt{s\_005}} & Build a decision map for choosing seed sources. & XMind \\
\hyperref[task:s-006]{\texttt{s\_006}} & Correct cooling-test measurements and plot the results. & LabPlot, Text editor \\
\hyperref[task:s-007]{\texttt{s\_007}} & Clean up research references and link findings to sources. & Zotero, Obsidian \\
\hyperref[task:s-008]{\texttt{s\_008}} & Model a pump enclosure with enough room for servicing. & Blender, Text editor \\
\hyperref[task:s-009]{\texttt{s\_009}} & Finish a trio arrangement and prepare each musician's part. & MuseScore Studio 4.6 \\
\hyperref[task:s-010]{\texttt{s\_010}} & Resolve a damaged-order claim and confirm replacement costs. & MailHub, Text editor \\
\hyperref[task:s-011]{\texttt{s\_011}} & Update customer renewal records and assign follow-up work. & CloudCRM, Text editor \\
\hyperref[task:s-012]{\texttt{s\_012}} & Draw a tool-return mat with outlines at actual size. & LibreOffice Draw \\
\hyperref[task:s-013]{\texttt{s\_013}} & Launch a service with a company page, job post and customer reply. & CareerLink, Text editor \\
\hyperref[task:s-014]{\texttt{s\_014}} & Fix a handbook's headings, page layout and references. & LibreOffice Writer \\
\hyperref[task:s-015]{\texttt{s\_015}} & Model cup returns, washing and daily stock levels. & LibreOffice Calc, Text editor \\
\hyperref[task:s-016]{\texttt{s\_016}} & Submit a path-diversion proposal using the latest site information. & FormCraft, MailHub, Text editor \\
\hyperref[task:s-017]{\texttt{s\_017}} & Package the latest maps and forms for an offline field inspection. & Default file manager, Terminal, Local text editor \\
\hyperref[task:s-018]{\texttt{s\_018}} & Animate a gripper lifting, moving and releasing a box. & Blender \\
\hyperref[task:s-019]{\texttt{s\_019}} & Map how audio and transcripts move through publication. & Obsidian \\
\hyperref[task:s-020]{\texttt{s\_020}} & Rebuild a recording-day calendar around people and equipment. & Calendar, Text editor \\
\hyperref[task:s-021]{\texttt{s\_021}} & Build a clickable help guide for library scanner problems. & LibreOffice Impress \\
\hyperref[task:s-022]{\texttt{s\_022}} & Review a research paper and submit a recommendation. & ReviewSphere, Local text editor \\
\hyperref[task:s-023]{\texttt{s\_023}} & Create an equipment request form that can be filled in on screen. & LibreOffice Writer, PDF viewer \\
\hyperref[task:s-024]{\texttt{s\_024}} & Synchronize two test videos for a fair motion comparison. & Shotcut \\
\hyperref[task:s-025]{\texttt{s\_025}} & Calculate water-service outages and summarize their impact. & LibreOffice Calc, Text editor \\
\hyperref[task:s-026]{\texttt{s\_026}} & Check a news article's facts and resolve suggested edits. & LibreOffice Writer, Local text editor \\
\hyperref[task:s-027]{\texttt{s\_027}} & Compare drying experiments and write an evidence-based report. & W\&B mock, Local text editor \\
\hyperref[task:s-028]{\texttt{s\_028}} & Book suitable seats for a group within access and budget limits. & Eventix, Local text editor \\
\hyperref[task:s-029]{\texttt{s\_029}} & Illustrate how rain runs off, soaks in and collects. & LibreOffice Draw, PDF viewer \\
\hyperref[task:s-030]{\texttt{s\_030}} & Generate personal practice letters from learners' assessment results. & LibreOffice Writer, LibreOffice Calc \\
\end{longtable}
\endgroup

\subsection{Task Instructions}
\label{sec:synthetic-task-instructions}

The following instructions retain the task wording and material paths.
The runner appends the same execution reminder to every instruction:
\begin{quote}
This instruction is in the context of whatever is currently open or running on the computer. You may need to investigate that for yourself. Do not ask user any follow-up questions.
\end{quote}

% s_001 instruction SHA-256: dc1fb50e5ce78b2910977a424529e1d1d8da406d04aeb28e0ace327378d45f3f
\subsubsection{\texttt{s\_001} Finish the Quay Works visitor guide}
\label{task:s-001}
\noindent\textit{Applications.} LibreOffice Impress, LibreOffice Writer.\par\nopagebreak[4]
\begingroup
\small
\setlength{\parindent}{0pt}
\setlength{\parskip}{0.3em}
Finish the two-page visitor guide for Quay Works Open Studios on Saturday 17 October 2026. The source packet is in \path{/home/user/CUESTraining/cues_r001/}. Start from \path{guide-draft-07-oct.odp}, an editable but outdated layout, and use \path{01-booking-record-07-oct.txt}, \path{02-change-register-through-16-oct.txt} and \path{03-facilities-notice-16-oct.txt} to bring it current through the register's 16 October, 16:00 export.

Keep the guide to two landscape A4 pages. Page 1 must contain the map and a compact studio directory. Preserve the map's orientation, room positions, room codes, physical room names, corridor geometry and doorway gaps. Update each occupied room with the correct current studio name and replace the old directory with each current studio's name, activity description, room code and room name. Apply the facilities notice to the visitor entrance, staff access, information desk, arrival route and legend. The arrival route must lead to the studio running Pocket notebook, using the current allocation. Keep the route in the corridor, pass below the desk, finish at the correct doorway and avoid crossing walls or labels. Keep these elements editable as native shapes and text.

Page 2 must show every currently publishable demonstration in start-time order, with its current public title, studio, room code and room name, and a start--end time range calculated from its current duration. Use a native editable table. Follow the source records' approval and identity rules, including inherited studio-room changes, withdrawals, cancellations and requests. Do not expose internal booking IDs or unapproved suggestions in the public guide. Preserve genuine overlaps between sessions rather than inventing new times.

On the guide, retain the event date and opening hours, state the current demonstration visitor policy, and apply the facilities notice's visitor footer to both pages. Remove draft notices, placeholders and obsolete visitor wording. Keep the existing restrained cream, dark-ink and teal style, adjusting text sizes, spacing and table dimensions to make both pages readable without clipping. The map must remain schematic rather than be redrawn as a different building.

Create \path{/home/user/CUESTraining/cues_r001/output/} and save the finished editable guide as \path{quay-works-visitor-guide.odp}. Export the same two pages as \path{quay-works-visitor-guide.pdf}. Leave the supplied source files intact.
\par\endgroup

% s_002 instruction SHA-256: 8e25372cda89e1b6295f5cd4bd54c2f9790fea1ebd25ee4975e1744f1346e9ae
\subsubsection{\texttt{s\_002} Edit Paper Bridge's learning-kit thank-you video}
\label{task:s-002}
\noindent\textit{Applications.} Shotcut.\par\nopagebreak[4]
\begingroup
\small
\setlength{\parindent}{0pt}
\setlength{\parskip}{0.3em}
Use Shotcut to make Paper Bridge's volunteer thank-you video from the files in \path{/home/user/CUESTraining/cues_r002}. Follow records/\path{client_brief.txt}, reconcile the release ledger with the packing card and clip log, and adapt \path{shotcut_workbench_guide.txt} to these materials. Create the full edited timeline, captions, simultaneous comparison and audio treatment requested in the brief. Save \path{/home/user/CUESTraining/cues_r002/output/paper_bridge.mlt} and \path{/home/user/CUESTraining/cues_r002/output/paper_bridge.mp4}. Keep the project editable and linked to the supplied local media. Review the exported video before finishing.
\par\endgroup

% s_003 instruction SHA-256: 09e524eadc20b1a0a234c27ef60efae884ba9535ee7706e04b4a297bcc636533
\subsubsection{\texttt{s\_003} Reconcile Willow Repair's shift handoff}
\label{task:s-003}
\noindent\textit{Applications.} TeamChat, LibreOffice Calc.\par\nopagebreak[4]
\begingroup
\small
\setlength{\parindent}{0pt}
\setlength{\parskip}{0.3em}
Prepare the Willow Repair Cooperative handoff for 19 October 2026. Work from \path{/home/user/CUESTraining/cues_r003} and the Willow TeamChat workspace. Use LibreOffice Calc to turn \path{handoff_start.xlsx} into an operational allocation tracker, following \path{operations.txt} and reconciling the current intake, shift, equipment and parts evidence. Keep it current with coordinator information received during your work. Resolve decision-critical uncertainty before committing affected allocations.

Save the editable result as \path{/home/user/CUESTraining/cues_r003/repair_handoff.ods}. Account for every intake ID with its allocation or explicit hold/defer reason and next action/owner; include formula-based volunteer/block, test-station/block and parts balances and a brief decision/source log. Preserve source records. Send a concise final summary in \#repair-handoff covering the workable plan, resource constraints, held/deferred jobs with follow-up ownership, material corrections and the saved workbook path. This is planning only; do not mark repairs completed or contact item owners.
\par\endgroup

% s_004 instruction SHA-256: 4debb443dbf28151f2b0555c1ad7811cf5166c896d4a3f99a3427fde001e9653
\subsubsection{\texttt{s\_004} Shape the Tidal Relay installation rehearsal mix}
\label{task:s-004}
\noindent\textit{Applications.} REAPER.\par\nopagebreak[4]
\begingroup
\small
\setlength{\parindent}{0pt}
\setlength{\parskip}{0.3em}
In REAPER, finish the Tidal Relay visual technical rehearsal in \path{/home/user/CUESTraining/cues_r004}. Read \path{COMMISSION.txt}, \path{CUE_SHEET.txt} and \path{STUDIO_TECHNIQUE.txt}. Arrange the supplied import-bin session, set the specified levels, stereo width and ending fades, and select eight suitable foreground gaps using the written waveform rule. Shape the Wash volume envelope using either direct envelope-point editing or Latch capture during playback. Follow the commission's control bands and ramp guidance using visible waveforms, envelope values and meters. Retain editable automation and safe playback modes. Visually check levels and export properties, then put the session with required media, the specified 2:24 WAV and operator note in \path{/home/user/CUESTraining/cues_r004/Deliverables}. This is a visual technical rehearsal; artistic room review is outside the assignment.
\par\endgroup

% s_005 instruction SHA-256: abb8f16e98d9af9df0f01fa530262469bc174a7efc1cd4bcb746499b93c7c996
\subsubsection{\texttt{s\_005} Rebuild Alder Reach's seed-source decision map}
\label{task:s-005}
\noindent\textit{Applications.} XMind.\par\nopagebreak[4]
\begingroup
\small
\setlength{\parindent}{0pt}
\setlength{\parskip}{0.3em}
You are preparing Alder Reach Seed Library's source review for 12 October 2026. Work from \path{/home/user/CUESTraining/cues_r005}. Read \path{00_README.txt} and the packet, then import \path{01_department_draft.opml} into XMind and rebuild it as a decision-oriented map using the method in \path{08_mapping_method.txt}. This is a fictional internal planning exercise; use only the supplied evidence.

The field and collections leads need to see what is worth pursuing this campaign, what can happen now, what needs evidence first, and what must stay outside trial stock. Replace the departmental organization with three to five short, verb-led decision branches in workflow order, plus a compact evidence/reference branch. Preserve each of the four stand IDs and four lot IDs in one detailed home apiece. Separate source suitability, access/collection permission and lot-release state; do not reduce them to one ready/not-ready flag. For every stand show current evidence, the campaign decision and next action or reason no further action is warranted. For every lot show trial-release or separate-retention state, live prerequisites, and any resolved historical issue. Give each live next action an accountable role using the packet's role guidance; do not invent dates, permissions or missing measurements.

Keep topic titles short enough to navigate. Use consistent labels for decision facets and place dated evidence, source IDs and reasons for rejecting stale claims in topic notes. Resolve the inherited claims against the correct authority, including count coverage, origin, permission scope, historical collection rights, chain recovery and drying evidence. Make retained demonstration material visibly distinct from trial stock. Add at least three named, directed relationships that expose real dependencies across the decision branches, including the dependence of lot release on its source/collection evidence. Use two meaningful boundaries around sibling action groups sharing an operational gate; their captions should explain the gate, not just say 'group'. Adapt the guide's method to this packet rather than copying its photo-archive example.

Arrange the expanded action view so decision order is clear, all four stands and all four lot homes can be found, and relationship labels do not overlap unrelated topic text or obscure boundaries. Keep evidence detail in notes where that improves the overview. Save the editable result as output/\path{Alder_Reach_decisions.xmind}. Export a legible overview to output/\path{Alder_Reach_overview.pdf} or .png if an available local export supports it without login/payment. If neither is available, export .opml or .md as the review companion instead and explain the export limitation in output/\path{handoff.txt}. In that short handoff, identify the chosen export, any application limitations, and evidence still needed before the team may advance the live decisions. Preserve the original inputs.
\par\endgroup

% s_006 instruction SHA-256: a2d39c7a90222ecc74ec33a437f4b22febdbcefc7c36adf1f1e8c84b25705b61
\subsubsection{\texttt{s\_006} Plot Orchard Loop's cooling-cartridge screening study}
\label{task:s-006}
\noindent\textit{Applications.} LabPlot, Text editor.\par\nopagebreak[4]
\begingroup
\small
\setlength{\parindent}{0pt}
\setlength{\parskip}{0.3em}
From: Nila Rao, thermal systems lead
Released: 2026-06-19 15:00 / report order OL-63
Subject: Choose the cartridge for the next coolant skid prototype

Use the attached bench records to prepare an editable LabPlot project and a single-page engineering figure. The three product names in the inventory must be the series labels. Our immediate decision is the 3.00 L/min duty point; use valid measurements whose corrected flow falls within 2.95--3.05 L/min, with no interpolation. A cartridge clears this screening only if its pressure-drop mean PLUS its combined standard uncertainty is at most 24.0 kPa AND its heat-transfer mean MINUS its standard uncertainty is at least 135 W. If no valid point remains in that duty window for a cartridge, label it unassessed. This is a conservative one-standard-uncertainty screening convention for this fictional prototype, not a statistical confidence or safety certification.

Build two vertically aligned Cartesian panels with the same corrected-flow axis range and the same cartridge colors: pressure drop above, heat transfer below. Show all reportable sweep points, their vertical one-standard-uncertainty bars, and readable product legends. Use points with connecting segments only between valid measurements from the same cartridge, in increasing corrected-flow order rather than acquisition order; do not fit a model or interpolate the screening result. Mark the 24.0 kPa and 135 W thresholds and make the duty window understandable without obscuring the data. Axis labels must name quantities and units. Choose ranges that include the entire valid sweep and its uncertainty bars. Provide a compact caption stating the uncertainty convention, handling of invalid/replaced points, and calibration release used. Use marker shape as well as color to distinguish the products. Layout, labels and legends must be legible in a landscape A4 review print, with no overlap or clipping.

Include a short decision annotation on the worksheet naming which cartridges clear the screen and explaining the limiting requirement for each other cartridge. Save the native project as output/orchard\_loop.lml and export the complete worksheet as output/\path{orchard_loop.pdf}. Include output/\path{data_disposition.txt} with every measurement ID and its status/reason, the source used for each cartridge identity, the calibration IDs applied to each capture, and the numeric duty-point mean and uncertainty values used for the decision. Keep this report concise; it is the audit trail for the figure, not a second report.

Perform spreadsheet corrections, data selection, curve construction and visual editing inside LabPlot. Reading the source files and writing the plain-text audit note can use a text editor. Do not compute the report in an external spreadsheet or script and then import completed results.

Materials are in \path{/home/user/CUESTraining/cues_r006}. Start with \path{READ_ME.txt} and reconcile the supplied source records. Create the output folder there. Save the three requested deliverables in that folder.

\par\endgroup

% s_007 instruction SHA-256: 95c2818a7f427cc08f7570c7ff91aae8e73a0b3937d3d70afc549c74aa8eb43d
\subsubsection{\texttt{s\_007} Curate Ferrybridge's reminder evidence}
\label{task:s-007}
\noindent\textit{Applications.} Zotero, Obsidian.\par\nopagebreak[4]
\begingroup
\small
\setlength{\parindent}{0pt}
\setlength{\parskip}{0.3em}
Prepare Ferrybridge's local reminder-evidence handoff using \path{/home/user/CUESTraining/cues_r007}. Read the packet and adapt \path{curation-procedure.md}. Do not use external lookups or accounts. Preserve the raw materials.

Import import/\path{analyst-exports.ris} into Zotero and reconcile the library using the supplied sources and authority records. Merge duplicate catalogue descriptions while retaining genuinely distinct versions, companion analyses, protocols and amendments. Correct issued metadata, preserve all incoming-record aliases in an identity ledger, and give every retained work a stable local source ID. Build a Ferrybridge Reminders project collection with Current evidence, Earlier versions, and Methods and amendments children; include project items explicitly in the parent. Add useful status/topic tags and a concise evidence note to every retained item, with source-section locators, version or dataset relationships, and relevant limitations. Corrections should remain traceable as their own records. No attachments or Zotero-to-Obsidian plugin are required.

Open the supplied ReviewVault in Obsidian and complete it in place. Reassess all four proposed claim notes; preserve each original proposal, add a scoped judgement using the procedure's vocabulary, and explain what evidence supports or limits it. Distinguish independent evidence from related analyses and observed associations from causal claims. For Value, show the defensible recurring cost per delivered message, disclose the setup charge, and explain what can or cannot be said about cost per extra attended appointment. Use stable source IDs shared with Zotero, source-section locators, and bidirectional source/claim links. Build source notes for every retained work, a source register/identity ledger mapping every incoming ID, and concept notes that explain the distinctions actually used in your reasoning. Complete Home navigation and an unresolved-questions note with concrete evidence gaps and implications for expansion. No community plugins, live URIs or external links are necessary; plain internal links suffice.

Leave the curated library saved in local Zotero. Under output/, save a RIS exchange export of the curated project and a short \path{handoff.md} explaining where the vault/library are, collection and tag conventions, consequential metadata/identity decisions, and remaining limits. The exchange file need not preserve collection structure; the local Zotero library and the editable vault are the primary deliverables.
\par\endgroup

% s_008 instruction SHA-256: 242b196ea68a127f5fc6bbbc82cba0faa925e690799688099e98c285bb75511f
\subsubsection{\texttt{s\_008} Resolve Tern Works' pump-enclosure clearances}
\label{task:s-008}
\noindent\textit{Applications.} Blender, Text editor.\par\nopagebreak[4]
\begingroup
\small
\setlength{\parindent}{0pt}
\setlength{\parskip}{0.3em}
Prepare Tern Works' Bay 4 pump-enclosure planning review in Blender. The supplied pack is at \path{/home/user/CUESTraining/cues_r008}. Import starter.obj with its MTL, preserve the measured KEEP\_ equipment, and replace the stale DRAFT\_ guard concept with the currently authorized enclosure. Reconcile the installed module history, module cards and field-level decision register; use their dimensions to lay out the guard, working front gate and service-clearance display. Model and organize the enclosure, visible hinge/latch blocks, closed gate, open-position ghost and clearance geometry through Blender UI editing. Do not generate the requested scene with a script. Give the assembly the specified materials, a simple floor, readable dimensional annotations and suitable lights/cameras. Produce an elevated front-east review render and an orthographic top clearance render following the review requirements in \path{04_decisions.txt}. This is a concept review, with simplified construction details. Save the editable scene as \path{/home/user/CUESTraining/cues_r008/output/TW41_review.blend} and the two PNGs as \path{TW41_assembly.png} and \path{TW41_clearance.png} in that output folder; use at least 1400 pixels on the long edge. Add \path{review_notes.txt} identifying the active module and accepted revisions, derived service-side location, any cutaway/visibility steps needed to reproduce the views, and any unresolved design issue. Keep named collections and both cameras in the native scene.
\par\endgroup

% s_009 instruction SHA-256: 5625cb87bd0e85f6f4f72aff9c87d11234f2759e2360995813270bce211a4b46
\subsubsection{\texttt{s\_009} Prepare Lantern Steps for chamber rehearsal}
\label{task:s-009}
\noindent\textit{Applications.} MuseScore Studio 4.6.\par\nopagebreak[4]
\begingroup
\small
\setlength{\parindent}{0pt}
\setlength{\parskip}{0.3em}
Prepare Lantern Steps for the Cedar Room youth chamber group using MuseScore. Open \path{/home/user/CUESTraining/cues_r009/Lantern_Steps_draft.musicxml} and follow \path{/home/user/CUESTraining/cues_r009/PRODUCTION_NOTES.txt} in full. Make the specified form, violin response, shared ending, phrasing, articulation and expression changes through MuseScore editing. Lay out the two-page full score and three one-page linked parts as specified, reviewing notation and exports visually. Save \path{/home/user/CUESTraining/cues_r009/finished/Lantern_Steps.mscz} and export \path{Lantern_Steps_score.pdf}, \path{Lantern_Steps_Flute.pdf}, \path{Lantern_Steps_Violin.pdf} and \path{Lantern_Steps_Cello.pdf} in that same finished folder. All musical requirements are written and visually checkable; no audio judgment is required.
\par\endgroup

% s_010 instruction SHA-256: 29e8da2bea12a6c47acc020ae374a04cb6d79915b49dd3643f537e4aa71b7cb7
\subsubsection{\texttt{s\_010} Resolve Northwick's damaged-folio supplier remedy}
\label{task:s-010}
\noindent\textit{Applications.} MailHub, Text editor.\par\nopagebreak[4]
\begingroup
\small
\setlength{\parindent}{0pt}
\setlength{\parskip}{0.3em}
You are Morgan Hale, Operations Coordinator at Northwick Editions. Resolve the NW-46 damaged folio case in MailHub using the correspondence and the records in \path{/home/user/CUESTraining/cues_r010}. Bring the supplier to a clear, authorized instruction that restores the required usable stock and requests the correct remedy for the paid invoice. Use your delegated authority and contact the relevant people where needed; keep the case current with incoming correspondence while you work. Send the final supplier instruction if the evidence supports a commitment within your authority, otherwise send the concrete issue for escalation without committing funds. Preserve the correspondence history. Replace or remove the obsolete full-remake draft so it cannot be sent accidentally. Leave one concise Finance handoff email as a draft to Nadia Bell, copied to Lea Moss, explaining the current stock, separate financial movements, evidence, and outstanding owners/actions. Do not treat an unacknowledged order or requested credit as complete. Organize the case with a MailHub label named NW-46 so the handoff and substantive case correspondence are easy to find. All participants are fictional and this is a simulated mailbox.
\par\endgroup

% s_011 instruction SHA-256: 307015d2015f78ce29e5c8cc1cc1e2db41ff0ca8e02756cd53cd882ea2047da6
\subsubsection{\texttt{s\_011} Prepare CedarWatch's Westmere renewal handoff}
\label{task:s-011}
\noindent\textit{Applications.} CloudCRM, Text editor.\par\nopagebreak[4]
\begingroup
\small
\setlength{\parindent}{0pt}
\setlength{\parskip}{0.3em}
You are Robin Hale at CedarWatch on 19 October 2026. Prepare the Westmere renewal handoff in CloudCRM using the four records in \path{/home/user/CUESTraining/cues_r011}. Read \path{01_operations_brief.txt} for our recording conventions and handoff requirements. Reconcile the existing Westmere account and contacts, the base renewal and separate Annex extension opportunities, the two service cases, and their next actions against the filed commercial, directory and service evidence. Correct relationships and stale information without deleting history or creating duplicate accounts, contacts or opportunities. Record annual subscription and first-invoice figures consistently; do not mistake filing, planning or budget acknowledgement for contract acceptance or technical release. Put actionable Finance and Service next steps with the specified owners and due dates into the opportunity nextStep fields and relevant case description. Publish the current "Westmere | 19 October handoff" Chatter post described in the operations brief and leave it visible. Keep work in CloudCRM; no external messages, attachments, reports, exports or local finished document are needed.
\par\endgroup

% s_012 instruction SHA-256: 01d352aad11b69b8a48280a75c512bbcc8a5705af41ecaf3ce79bdc6d3690d80
\subsubsection{\texttt{s\_012} Build Mend \& Lend's true-size kit return mat}
\label{task:s-012}
\noindent\textit{Applications.} LibreOffice Draw.\par\nopagebreak[4]
\begingroup
\small
\setlength{\parindent}{0pt}
\setlength{\parskip}{0.3em}
Use LibreOffice Draw to turn \path{/home/user/CUESTraining/cues_r012/return-mat-starter.svg} into Mend \& Lend's finished kit return mat. Follow \path{/home/user/CUESTraining/cues_r012/design-brief.txt} for the item geometry, content, layout and print requirements. Substantially edit the supplied vector parts: close the scissors, cap the seam ripper, tuck the tape tail, resize all five silhouettes, and arrange a readable set of item stations with a handling strip and scale check. Keep the working result editable in Draw. Save \path{/home/user/CUESTraining/cues_r012/Mend-and-Lend-return-mat.odg} and export the matching one-page A3 landscape PDF to \path{/home/user/CUESTraining/cues_r012/Mend-and-Lend-return-mat.pdf}. Inspect the page before handing it over; no external assets are needed.
\par\endgroup

% s_013 instruction SHA-256: 8fb078d3ec63b8e97704f8a14a8ad575102ca18e34d60cd7aa8ab16ffd244abd
\subsubsection{\texttt{s\_013} Open Vela Shift's assisted migration intake}
\label{task:s-013}
\noindent\textit{Applications.} CareerLink, Text editor.\par\nopagebreak[4]
\begingroup
\small
\setlength{\parindent}{0pt}
\setlength{\parskip}{0.3em}
You are Saira Bell, company admin for Vela Shift. Prepare our CareerLink page for the approved Vela Bridge assisted-migration opening. It is 14 September 2026 for this task. The three records in \path{/home/user/CUESTraining/cues_r013} explain the approved offer, operations handoff and old campaign copy; use their authority and the live page/conversation history to resolve differences.

In CareerLink, update only the company Overview and Specialties so they accurately describe the approved service alongside our existing software. Configure the Message button and inbox topics for the staffed intake routes. Create the single open Migration Support Coordinator role specified in the approved hiring authorization, with a complete, readable description covering terms, work, requirements and the application route. Reply as the company in Mae Porter --- Marlow Stage's existing conversation: address the actual situation in the thread, correct the obsolete claims, give their applicable Bridge total and request what is still needed for manual confirmation. Finally publish one company launch post explaining the approved offer and inquiry route in clear customer-facing prose. Preserve the historical post, existing conversation history and unrelated company details. No scheduling, attachments or external service actions are needed; the company website address is context only. Finish with the changes saved in CareerLink; no separate report is required.
\par\endgroup

% s_014 instruction SHA-256: b9eb1f31037cd28699abee272c8d3489545968b448a07485c85b36add5676d62
\subsubsection{\texttt{s\_014} Repair Common Ground's reusable facilitator handbook}
\label{task:s-014}
\noindent\textit{Applications.} LibreOffice Writer.\par\nopagebreak[4]
\begingroup
\small
\setlength{\parindent}{0pt}
\setlength{\parskip}{0.3em}
Prepare Common Ground's conversation-lab handbook for facilitators. Open \path{/home/user/CUESTraining/cues_r014/conversation_lab_draft.docx} in LibreOffice Writer and adapt the procedures in \path{writer_field_guide.txt} in the same folder. Preserve the supplied workshop prose, scenario, sequence, table labels and cover information; this is a publication repair, not a content rewrite. Remove the preparation-copy label and the note saying the contents was typed.

Make Prepare the session, Run the practice, Close and carry forward, and Observation sheet real level-one headings. Make the two named subsections under each of the first three chapters level-two headings. Keep the cover title and contents label out of the body outline. Set coherent reusable paragraph styles: body 11--12 pt, level-one headings 16--18 pt and level-two headings 12--14 pt, with headings kept with following text. Replace the typed contents with an updateable, linked table of contents showing both heading levels and their page numbers. Replace the two literal heading-and-page citations in the prose with live heading-text and page-reference fields pointing to Recover after a snag and Observation sheet; preserve the surrounding sentence wording.

Use Letter portrait pages for the cover, contents and three prose chapters, with margins of at least 0.7 inches. Keep the cover and contents on separate unnumbered pages. Begin Prepare the session on the next page, numbered 1, with subsequent page-number fields in the footer. Each of the three prose chapters should begin a fresh page; allow prose to flow within its chapter without stranded headings or unnecessary blank pages. Give the final Observation sheet its own Letter landscape page style and continue the body page-number sequence. It must occupy one detachable page, including its title, identification line, instructions, table and two closing prompts. Keep all four table columns in the supplied order. Give the Notice column 1.5--1.8 inches and each writing column at least 2.1 inches, with each of the four body rows at least 0.75 inches tall. Keep the header row readable, body labels intact, rows unsplit and both closing prompts clear of the table and footer. Do not shrink body or form text below 11 pt to force a fit.

Refresh contents and references after pagination, inspect the print layout, and save \path{conversation_lab.odt} plus a matching \path{conversation_lab.pdf} in \path{/home/user/CUESTraining/cues_r014/output/}. Retain the original draft and field guide.
\par\endgroup

% s_015 instruction SHA-256: a948abf10a0904a97a77a55b350271bf9c9f8631fd4f8592ff7c0c7c24e5abae
\subsubsection{\texttt{s\_015} Finish Rill Market's reusable cup circulation model}
\label{task:s-015}
\noindent\textit{Applications.} LibreOffice Calc, Text editor.\par\nopagebreak[4]
\begingroup
\small
\setlength{\parindent}{0pt}
\setlength{\parskip}{0.3em}
Finish Rill Market's cup circulation planning workbook in LibreOffice Calc. The files are in \path{/home/user/CUESTraining/cues_r015}. Start from \path{cup-circulation-starter.xlsx}, read \path{operations-spec.txt}, and adapt the procedure in \path{calc-reuse-guide.txt} to this workbook. Complete one formula-driven six-day model with the specified daily order, previous-day returns, clean and dirty stock carryforward, capacity constraints, and blank-versus-zero wash overrides. Wire the single case selector through the active drivers to all days and the summary. Preserve the supplied inputs and both tabs. Add usable numeric input validity, live shortage/reserve warnings, and accidental-edit protection while keeping the intended inputs editable. Include brief operating notes in the workbook. Exercise both cases and representative input changes, restore the original values, and leave Regular service selected. Save \path{/home/user/CUESTraining/cues_r015/cup-circulation.ods} and a short \path{/home/user/CUESTraining/cues_r015/handoff.txt} describing what you changed, the input changes tried, observed behavior and any remaining limitation. Follow the operating specification for all model definitions and handoff details.
\par\endgroup

% s_016 instruction SHA-256: ac2684ba6ddbc4d3e311151565e46bb2c3aad829056fec334ec74b3f1537f4ee
\subsubsection{\texttt{s\_016} Lodge Rowan Borough's temporary towpath diversion proposal}
\label{task:s-016}
\noindent\textit{Applications.} FormCraft, MailHub, Text editor.\par\nopagebreak[4]
\begingroup
\small
\setlength{\parindent}{0pt}
\setlength{\parskip}{0.3em}
Act as Rowan Ellis, the Rowan Borough case officer for RB-26-104. Read the three local records in \path{/home/user/CUESTraining/cues_r016} and the case correspondence in MailHub. Complete and submit the existing FormCraft temporary pedestrian diversion intake through its GUI, requesting the shortest defensible route under the supplied case conditions and within the delegated funding ceiling. Synthesize concise answers in your own words: justify the route against the alternatives, derive the length and funding request, connect remaining works to pre-opening checks, draft useful public notice wording, and explain operating and switch-back controls. The Orchard gate inspection is being completed while you prepare the case; check its final emailed result before this irreversible single submission and revise your draft where needed. Keep MailHub open with notifications enabled while working. Do not send email or create another application. The form has no saved draft: keep its page open while consulting the records and mailbox. Your deliverable is the one submitted proposal; no separate report is required.
\par\endgroup

% s_017 instruction SHA-256: 64f0556db54a1992fb88d6712235d13c8cccfb3005bcd7b772d0c7f9b8848ce8
\subsubsection{\texttt{s\_017} Pack Harth Fen's offline inspection field kit}
\label{task:s-017}
\noindent\textit{Applications.} Default file manager, Terminal, Local text editor.\par\nopagebreak[4]
\begingroup
\small
\setlength{\parindent}{0pt}
\setlength{\parskip}{0.3em}
Finish Harth Fen's offline field kit for the 14 September bank inspection walk, using the files in \path{/home/user/CUESTraining/cues_r017}. Read \path{START_HERE.md} and the records to determine the trip's route scope and the issued materials. A volunteer's unfinished copy is in working/field-kit. Repair and complete it, preserving the issued route content, matching tables/maps/cards and every required offline companion. Leave observation fields blank. Keep the original deliveries and records unchanged, and exclude out-of-scope routes, design previews, internal records and packing leftovers from the recipient kit.

Add a concise \path{INDEX.md} at the kit root with relative links to each included route card, its survey revision, useful open-notice context and brief offline opening/return instructions. Add \path{MANIFEST.csv} containing each other delivered file's relative path, byte size and SHA-256; omit the manifest itself. Package the kit as handoff/\path{Harth-Fen-2026-09-14.zip}, with one top-level Harth-Fen-2026-09-14 folder and no absolute paths. Keep the completed editable kit in working/field-kit.

Extract your ZIP to a separate check folder inside this task tree and check its inventory, relative links and route revision consistency. Put a brief handoff/\path{packing-note.md} outside the ZIP explaining the source basis, what you repaired in the unfinished copy, exclusions and the extraction checks performed. Work only inside \path{/home/user/CUESTraining/cues_r017}; no external transfer is needed.
\par\endgroup

% s_018 instruction SHA-256: fa568bf0893be98f4bdaca22dc9692690d6b30d479484ad01b493aca2a266da7
\subsubsection{\texttt{s\_018} Animate Meridian Gallery's grip--carry--release demonstration}
\label{task:s-018}
\noindent\textit{Applications.} Blender.\par\nopagebreak[4]
\begingroup
\small
\setlength{\parindent}{0pt}
\setlength{\parskip}{0.3em}
Create the silent eight-second gripper demonstration for Meridian Maker Gallery in Blender. Import the original separate meshes from \path{/home/user/CUESTraining/cues_r018/gripper_parts.obj} with its matching MTL, and follow \path{/home/user/CUESTraining/cues_r018/production_brief.txt} for the motion schedule, contact geometry, hierarchy, camera and handoff. The starting geometry is static; build the native animation yourself. Make the grasp, lifted transfer, set-down and release readable, with the orange parcel remaining on the destination after the gripper retracts. Save \path{/home/user/CUESTraining/cues_r018/output/grip_carry_release.blend} and the four 640 \(\times\) 400 camera-view PNG previews specified in the brief. Inspect the full sequence in the timeline and leave the saved scene at frame 1 in camera view. Create the output directory as needed. No rendered movie is required.
\par\endgroup

% s_019 instruction SHA-256: f635f3233395561cd9e0a329e94b84522c23b0b897b1a10740e0cd193a1556e2
\subsubsection{\texttt{s\_019} Explain Loopline's accessible episode release}
\label{task:s-019}
\noindent\textit{Applications.} Obsidian.\par\nopagebreak[4]
\begingroup
\small
\setlength{\parindent}{0pt}
\setlength{\parskip}{0.3em}
Finish Loopline's internal producer induction board in Obsidian. The local vault is \path{/home/user/CUESTraining/cues_r019/Loopline}; open Episode release.canvas and read the three Notes files. The board is currently a parking area, not an approved process. Turn it into a readable, non-hierarchical explanation a new producer can use during a twenty-minute induction.

Synthesize the process rather than pasting the notes. Show the normal route from recording/brief to a public episode, including guest approval tied to a particular cut, the parallel work that becomes possible after locking, the final-audio dependency of transcript review, and the common publication gate. Make it clear who owns the main work and which items the listener actually receives. Page-copy preparation must be visible without implying that it starts only after the audio is locked.

Show listener reports returning through the producer and distinguish a text-only correction from a change to spoken content. The former should visibly return only through the checks it needs; the latter must return to the appropriate upstream work and rejoin the release gate with replacement audio, transcript and approval. Make the transcript-review revision loop understandable too. Do not invent additional approval bodies, deadlines or episode data.

Choose a spatial organization in which native labeled groups, directed labeled connections and card hierarchy explain these relationships. Reserve a clear route for the correction loops so they do not obscure the normal handoffs. Connection labels should communicate the handoff or condition, not just say ``next.'' Use color consistently with a small legend, but keep meanings readable without relying on color alone. Avoid a single long chain or a departmental tree: parallel work, shared inputs and return routes should be apparent from placement. Aim for roughly 12--16 concise main cards; use supporting notes for detail. At a useful overview zoom, headings and the main route should remain readable; at working zoom, no text, labels or cards should overlap or be clipped, and edge endpoints must be unambiguous.

Save the finished editable board as Episode release.canvas in the same vault. Keep the three source notes intact and accessible from the board, using a clearly secondary reference area if helpful. Remove unfinished planning prompts. The board and its local vault content are the deliverable; no export or separate report is needed.
\par\endgroup

% s_020 instruction SHA-256: 91b2aefecf8915dd9c680feca1fef572a2c2ba8a2121f1b16e4c067818ce30fe
\subsubsection{\texttt{s\_020} Reconcile Morrow Voices' final recording day}
\label{task:s-020}
\noindent\textit{Applications.} Calendar, Text editor.\par\nopagebreak[4]
\begingroup
\small
\setlength{\parindent}{0pt}
\setlength{\parskip}{0.3em}
Please finish Ellis Vale's Morrow Voices recording-day calendar for Wednesday 21 October 2026. Start with \path{/home/user/CUESTraining/cues_r020/READ_ME.txt} and its three companion records, and read the existing Calendar event descriptions. Reconcile the sessions and logistics against the current evidence, then make the actual local Calendar usable under the supplied operating rules. Edit useful reservations, delete obsolete ones, and add missing confirmed work and protected intervals. Place active work as early as feasible, preserve fixed commitments, and time the walk to arrive at the courier appointment. Put concise source references in changed/new descriptions; use participant names in text and leave Guests empty. All times are UTC and the browser is configured to render UTC. Work only in the mock app, with no real messaging or account login. Leave the calendar ready as the operational handoff; no separate timetable is needed.
\par\endgroup

% s_021 instruction SHA-256: 9c7a61cd40fe3f634600e75ce8c62e2752a46785b31664c3a1930f3b104a046b
\subsubsection{\texttt{s\_021} Build Wren Library's self-service scan help}
\label{task:s-021}
\noindent\textit{Applications.} LibreOffice Impress.\par\nopagebreak[4]
\begingroup
\small
\setlength{\parindent}{0pt}
\setlength{\parskip}{0.3em}
Finish the self-service scan-station help presentation for Wren Library. Open \path{/home/user/CUESTraining/cues_r021/scan_station_draft.odp} and follow the complete commission and approved station content in \path{/home/user/CUESTraining/cues_r021/station_brief.txt}. Turn the eight-slide unfinished draft into a readable self-guided native Impress presentation with working local choice, outcome and return controls. Preserve the stated remedy limits and use the specified control sizes and placement. Save \path{/home/user/CUESTraining/cues_r021/scan_station_help.odp}, exercise all branches in its native slide show, and leave a brief factual \path{/home/user/CUESTraining/cues_r021/navigation_check.txt} with routes actually checked and any unverified behavior.
\par\endgroup

% s_022 instruction SHA-256: b15107b72050ff9b7429cb748e0074c01a5ed454e7ca82185ee9a81550a072e4
\subsubsection{\texttt{s\_022} Review the Branch Hints learning study}
\label{task:s-022}
\noindent\textit{Applications.} ReviewSphere, Local text editor.\par\nopagebreak[4]
\begingroup
\small
\setlength{\parindent}{0pt}
\setlength{\parskip}{0.3em}
You are Neris Calven, the assigned reviewer for Submission 22, ``Branch hints improve transferable learning in unfamiliar archive sorting,'' in the mock Ledger Learning Workshop 2027. Read the local package beginning at \path{/home/user/CUESTraining/cues_r022/00_reading_index.txt} and submit one complete official review in ReviewSphere. Assess the central learning claim against the manuscript, deposited plan, assessment specification, workshop counts and analysis memo. Use the venue guidance for review scope. Populate title, review, strengths, weaknesses, questions and limitations with a constructive, sourced assessment, showing the numerical comparisons that materially support your reasoning. Distinguish what existing evidence could resolve from what needs further research. Choose rating and confidence yourself. After submission, reload and confirm that your review remains visible and complete; edit that same review if needed rather than creating duplicates. The assignment is already seeded; do not change assignments, edit the manuscript or contact anyone. All research and identities are fictional. No external reading or separate output document is required.
\par\endgroup

% s_023 instruction SHA-256: 9ba4ae372bd676a8c5f47135809318dfceec1e16d90417a80991f429416e1481
\subsubsection{\texttt{s\_023} Finish Bracken House's local display kit request form}
\label{task:s-023}
\noindent\textit{Applications.} LibreOffice Writer, PDF viewer.\par\nopagebreak[4]
\begingroup
\small
\setlength{\parindent}{0pt}
\setlength{\parskip}{0.3em}
The Bracken House equipment desk needs a reusable local request form for its display kits. Open \path{/home/user/CUESTraining/cues_r023/display-kit-draft.docx} in LibreOffice Writer and use \path{/home/user/CUESTraining/cues_r023/response-specification.txt} to finish it. Replace the text placeholders with appropriate classic Writer form controls, including a mutually exclusive handover choice, independent equipment choices, a fixed venue list and useful text areas. Preserve the approved public content, arrange a readable one-page US Letter form, and make keyboard traversal follow the reading order. Save the editable blank form as \path{/home/user/CUESTraining/cues_r023/display-kit-request.odt} and export its blank fillable counterpart as \path{/home/user/CUESTraining/cues_r023/display-kit-request.pdf}. Exercise text entry, option independence, the exclusive choice and keyboard navigation in both formats; remove all temporary answers and reopen the final blank files. Leave the ODT ready for filling with Design Mode off. This is a local form: do not add submission actions, database connections, macros, calculations or passwords. The two blank files are the deliverables; no completed example is needed.
\par\endgroup

% s_024 instruction SHA-256: 97ab0eb87d606e70b2cd7c6b5be89e367b070e7a8dc212d5ff49150c5585ccd1
\subsubsection{\texttt{s\_024} Explain Mere Studio's spring-cart damping comparison}
\label{task:s-024}
\noindent\textit{Applications.} Shotcut.\par\nopagebreak[4]
\begingroup
\small
\setlength{\parindent}{0pt}
\setlength{\parskip}{0.3em}
Use Shotcut to complete the Mere Studio teaching video described below. Inputs are in \path{/home/user/CUESTraining/cues_r024/inputs/} and this brief is also in \path{/home/user/CUESTraining/cues_r024/production_brief.txt}. The timeline has not been started.

MERE STUDIO --- Spring-cart demonstration

We need a short silent teaching video for a public mechanics display. Two original schematic recordings show the same spring cart released from the same displacement with different damping. These are illustrative animations, not laboratory measurements. Use only the two supplied MP4s. Cobalt is Damping A; amber is Damping B. The physical cart width and the 100 MM bar are the same in both recordings; their framing is different.

Make an 18--24 second, 1280\(\times\)720, 30 fps comparison in Shotcut:

\textbullet{} A short opening introduces ``Same release, different damping'' and identifies the footage as an illustrative simulation.
\textbullet{} The central comparison places Damping A in the upper panel and Damping B in the lower panel, playing simultaneously at their original speed. Trim away unnecessary waiting, retain about half a second held before release, synchronize the visible release to within two output frames, and show at least ten seconds after release in both panels. The gray lamp turns green when the release jaw withdraws. Confirm alignment from motion as well as the lamp; do not synchronize the raw file starts.
\textbullet{} Match physical scale using the visible 100 MM bars: they should appear equally long to within roughly 5\%. Align the two vertical REST datums to the same screen x position to within about 8 pixels. Preserve proportions. Keep the moving carts, spring anchor, rest band and scale bars visible throughout. Crop unused source margins and arrange clean, distinct panels, with no footage spilling into the other panel. Panel labels must remain legible and not cover the moving cart or scale bars. Original corner run labels may be cropped away.
\textbullet{} Follow the main comparison with a synchronized half-speed replay of a 2--3 second interval that clearly shows the difference around the first passage through rest and subsequent reversal. Choose the interval by inspecting the footage. Enlarge/reframe this replay around the carts and rest bands; keep the two panels at equal physical scale and their rest datums aligned. The spring anchors and scale bars may fall outside this closer crop, but the cart travel during the chosen interval must remain visible. Label the replay ``Half speed''.
\textbullet{} End by holding an informative comparison image for approximately two seconds with one short, plain-language observation supported by what viewers just saw. Mention both overshoot and settling behavior, without inventing measured settling times or presenting a real engineering certification. Keep this readable at normal display size.

Simple cuts are suitable. No sound or music is needed. Do not stretch a clip independently to force correspondence: use synchronized trims for the main view and the same 0.5\(\times\) speed for both replay clips. No external assets are needed. Keep all video, panel placements, text and replay segments editable in the native project.

Save deliverables in \path{/home/user/CUESTraining/cues_r024/output/}:
  \path{damping_comparison.mlt} --- editable Shotcut project, using the supplied local input paths
  \path{damping_comparison.mp4} --- H.264/yuv420p, 1280\(\times\)720 at 30 fps, silent
  \path{handoff.txt} --- brief source in-points used for the synchronized main comparison, source interval used for each replay, and the visual checks you performed. A few lines suffice.

Review the exported picture, especially the release alignment, readable labels, equal cart scale, and whether moving parts stay inside their panels. Leave the native project open in Shotcut.

\par\endgroup

% s_025 instruction SHA-256: 8e48b9fa14ceb8c89738082fd02db512ff3123b2c825c0288a590074370b8289
\subsubsection{\texttt{s\_025} Reconstruct Waterpoint Campus's service interruptions}
\label{task:s-025}
\noindent\textit{Applications.} LibreOffice Calc, Text editor.\par\nopagebreak[4]
\begingroup
\small
\setlength{\parindent}{0pt}
\setlength{\parskip}{0.3em}
Prepare Waterpoint Campus Facilities' 14 May 2026 interruption report in LibreOffice Calc. The supplied files are in \path{/home/user/CUESTraining/cues_r025}. Read \path{READ-ME.txt}, then use \path{Waterpoint-records.xlsx} to reconstruct the interruption episodes that overlap 08:00--16:00. Preserve the three original source sheets.

Add an auditable analysis sheet with one row per included episode. Show the episode ID, station and location, zone, outlet count, original start, authorized release if present, status at the cutoff, in-window unavailable minutes and unavailable outlet-minutes. Include the contributing Call IDs and the release Visit ID (or an explicit note that no release exists in the export). Use visible spreadsheet formulas for the two duration/impact measures, with clearly labeled window boundaries. Grouping may be done manually; the reasoning must remain traceable. Briefly account for source records excluded solely by the reporting window.

On a separate report sheet, build a native, editable Calc pivot using that episode table as its source. Use zone as rows and cutoff status as columns. Report episode count, sum of unavailable minutes, and sum of unavailable outlet-minutes, with clear labels and grand totals. Use ordinary count/sum aggregations. Do not count call rows or rely on distinct count. Beside the pivot, state which zone has the greatest unavailable outlet-minutes, whether open episodes contribute to that finding, and one limitation of this service measure. Cite the supporting pivot figures.

Refresh the pivot after finishing the analysis and check its grand totals against the episode table. Note the check briefly. Save the complete editable workbook as \path{/home/user/CUESTraining/cues_r025/Waterpoint-review.ods} and export the report sheet only to \path{/home/user/CUESTraining/cues_r025/Waterpoint-review.pdf}, legible on one landscape page. Leave the ODS open on the report sheet.
\par\endgroup

% s_026 instruction SHA-256: 75493fb61febe6403a43a2da3aac9500523cdc49ebcec59e1961de5de461b7b4
\subsubsection{\texttt{s\_026} Adjudicate Northbank Weekly's evening lending feature}
\label{task:s-026}
\noindent\textit{Applications.} LibreOffice Writer, Local text editor.\par\nopagebreak[4]
\begingroup
\small
\setlength{\parindent}{0pt}
\setlength{\parskip}{0.3em}
You are the features subeditor for Northbank Weekly. In \path{/home/user/CUESTraining/cues_r026}, open \path{after_dark_review.docx} in LibreOffice Writer and read \path{desk_brief.txt} and \path{reporting_packet.txt}. Complete the editorial review for the 20 June issue: adjudicate the inherited tracked proposals against the reporting, repair other material inaccuracies or unsupported implications, and preserve the readable local feature and reported scene. Do not treat either reviewer as automatically authoritative. Resolve the inherited revisions individually or in sensible related groups; keep your own additions and rewrites as pending native tracked changes for the editor. Add three or four concise anchored comments explaining consequential judgments with source headings/dates. Save one editable Writer document at \path{/home/user/CUESTraining/cues_r026/output/after_dark_editorial_review.odt}, with your new revisions and comments still reviewable. Do not create a clean copy, a decision spreadsheet or publication PDF, and do not send or publish the article.
\par\endgroup

% s_027 instruction SHA-256: 1601a518153c7122af2d8da0eb2e3b84a5aaeab9c44370969153670c6ae84986
\subsubsection{\texttt{s\_027} Assess Larch Materials' pulsed-air drying trial}
\label{task:s-027}
\noindent\textit{Applications.} W\&B mock, Local text editor.\par\nopagebreak[4]
\begingroup
\small
\setlength{\parindent}{0pt}
\setlength{\parskip}{0.3em}
In the existing W\&B mock, finish the unfinished Pilot-oven discussion report in larch-lab/liner-drying. Read \path{/home/user/CUESTraining/cues_r027/experiment-brief.md} and \path{bench-ledger.md} for the engineering question, metric definitions and executed conditions. Establish which runs support direct comparisons, examine the time histories and sampled qualification behavior, and recommend what the next pilot experiment should establish. Curate run tags, append assessment notes, and persist workspace visibility for your main comparison while retaining all records. Compose the native report with numerical evidence, limitations and at least three useful line charts, each with explicitly selected runs and nearby series identification; choose the charts and comparisons yourself. Keep context evidence distinct. The report should be useful to the process engineer without opening these instructions. Work only with supplied fictional records; no new experiment execution, external account or export is requested.
\par\endgroup

% s_028 instruction SHA-256: ad56efa31f33d27dd207289dd586fadb78aa537832d6694f8f25b5630563bc30
\subsubsection{\texttt{s\_028} Book Alder Family Centre's listening outing}
\label{task:s-028}
\noindent\textit{Applications.} Eventix, Local text editor.\par\nopagebreak[4]
\begingroup
\small
\setlength{\parindent}{0pt}
\setlength{\parskip}{0.3em}
Arrange Alder Family Centre's six-person listening outing in the local Eventix mock. Read \path{/home/user/CUESTraining/cues_r028/outing-request.md}, \path{programme-cards.md}, \path{venue-access.md} and \path{mock-payment.md}. Compare the authored events and inspect visual seat availability; choose one event and six seats satisfying the request, complete one mock order using the supplied synthetic payment details, then save \path{/home/user/CUESTraining/cues_r028/handoff.md} with the requested practical handoff and leave Eventix on its confirmation page. Browse and reason before making the first seat selection so ordinary checkout fits the ten-minute timer. The browser must already be configured to local UTC and this task must begin in a fresh Eventix tab with no prior cart timer; these are setup preconditions. This is a local mock purchase only; no real service or money is involved.
\par\endgroup

% s_029 instruction SHA-256: fc30d4a5e25a7cbbf70183e4500e0f1be2715d3bff0fb7c5eff7e5efc6248322
\subsubsection{\texttt{s\_029} Illustrate Brook Vale's three rain pathways}
\label{task:s-029}
\noindent\textit{Applications.} LibreOffice Draw, PDF viewer.\par\nopagebreak[4]
\begingroup
\small
\setlength{\parindent}{0pt}
\setlength{\parskip}{0.3em}
Finish Brook Vale Community Garden's rain comparison poster in LibreOffice Draw. Read \path{/home/user/CUESTraining/cues_r029/production-brief.txt} and use \path{/home/user/CUESTraining/cues_r029/rain-studies.svg} as the editable unfinished artwork. Build the three coordinated cross-sections so their geometry, water layers and directional marks communicate the approved runoff, infiltration and temporary-storage stories. Keep the final source genuinely editable as native shapes and text, with each illustration grouped for reuse; do not deliver a flattened picture. Import or convert the SVG into editable Draw shapes as needed and check that individual objects can be selected before proceeding. Follow the brief's composition, diagram and handoff requirements. Create \path{/home/user/CUESTraining/cues_r029/output/} and save \path{rain-display.odg}, the one-page \path{rain-display.pdf} and \path{handoff.txt} there. Inspect both the native page and PDF before finishing.
\par\endgroup

% s_030 instruction SHA-256: c29b9f6558ed389f7b3b5b69d94e2c470dd53c9f94ce54693dad278cca363200
\subsubsection{\texttt{s\_030} Prepare Mereford's individualized home-practice letters}
\label{task:s-030}
\noindent\textit{Applications.} LibreOffice Writer, LibreOffice Calc.\par\nopagebreak[4]
\begingroup
\small
\setlength{\parindent}{0pt}
\setlength{\parskip}{0.3em}
Prepare Measure Together's home-practice letters using the files in \path{/home/user/CUESTraining/cues_r030}. Read \path{production_request.txt}, \path{teaching_bank.txt}, \path{diagnostic_responses.txt} and \path{learner_roster.csv}. The \path{home_practice_master.docx} is an unfinished editable starting letter, with plain-text placeholders.

Interpret each learner's responses using the supplied prerequisite and completeness rules. Build a local Calc data source that retains all learners and records the chosen issue/hold state, a concise internal evidence reason, and the individualized teaching content needed for the letters. Follow the consent and incomplete-observation rules. For each issued letter use one appropriate activity from the bank and supportive feedback grounded in that learner's responses, with no score or public internal labels.

Convert the starting letter into one reusable native Writer merge master with genuine database-backed fields. Suppress optional address lines and the kit paragraph using native conditional/address features. Bind to the saved local data, retain any connector needed to reopen the merge, and produce a single combined editable ODT and PDF in LearnerID order. Follow every requirement in \path{production_request.txt}, including one A4 page per recipient, at least 11 pt body text, at least 35 mm clear response space, the correct practice and session details, and consistent recipient data. Inspect the full run and regenerate from the master if anything needs correction.

Save final files together in \path{/home/user/CUESTraining/cues_r030/output/}: \path{home_practice_master.odt}, \path{learners.ods}, \path{home_practice_letters.odt}, \path{home_practice_letters.pdf}, any required local database connector, and \path{handoff.txt}. The handoff should identify held IDs and reasons and explain the local connection and repeat-run procedure. Do not email, send, or physically print anything.
\par\endgroup

% Generated by docs/experiments/scienceboard-synthetic-appendix/build_tex.py.
% Instructions as received in eval_cache_scienceboard_qwen_sb_syn50v15_60m (identical
% across candidates), runner suffix removed. Titles and applications from task.json.
\section{ScienceBoard Synthetic Training Tasks}
\label{sec:appendix-scienceboard-synthetic-tasks}

We use 48 synthetic tasks to provide construction experience for
\methodname{} on ScienceBoard. They cover the benchmark's six scientific
applications, with 10 for Celestia, 7 for ChimeraX, 9 for GRASS GIS, 10 for KAlgebra, 7 for Lean and 5 for TeXstudio. Each task includes an instruction, an initial
application state and, where needed, input materials. The tasks were authored
and reviewed as described in Section~\ref{sec:setup}. Of the 50 authored
tasks, we exclude \texttt{sb\_chimerax\_006} and \texttt{sb\_grassgis\_010}
because their initialization fails before the task agent starts. The tasks
have no evaluators, so attempts are recorded without success labels.
Table~\ref{tab:scienceboard-synthetic-tasks} lists the tasks, and
Section~\ref{sec:scienceboard-synthetic-instructions} gives the full
instructions as the task agent received them.

\begingroup
\small
\setlength{\tabcolsep}{4pt}
\renewcommand{\arraystretch}{1.12}
\begin{longtable}{@{}p{0.2\textwidth}p{0.6\textwidth}p{0.13\textwidth}@{}}
\caption{\textbf{ScienceBoard synthetic training tasks.} The 48 tasks span six
  scientific applications and give the meta-agent trajectories to study,
  without ground-truth success labels.}
\label{tab:scienceboard-synthetic-tasks}\\
\toprule
\textbf{ID} & \textbf{Task} & \textbf{Application} \\
\midrule
\endfirsthead
\multicolumn{3}{l}{\textit{Table~\ref{tab:scienceboard-synthetic-tasks} continued}}\\
\toprule
\textbf{ID} & \textbf{Task} & \textbf{Application} \\
\midrule
\endhead
\midrule
\multicolumn{3}{r}{\textit{Continued on the next page}}\\
\endfoot
\bottomrule
\endlastfoot
\hyperref[task:sb-celestia-001]{\texttt{sb\_celestia\_001}} & One Moon, two viewpoints. & Celestia \\
\hyperref[task:sb-celestia-002]{\texttt{sb\_celestia\_002}} & A fixed coastline under changing daylight. & Celestia \\
\hyperref[task:sb-celestia-003]{\texttt{sb\_celestia\_003}} & Return to a ring-shadow observing point. & Celestia \\
\hyperref[task:sb-celestia-004]{\texttt{sb\_celestia\_004}} & Enlarge Mars by magnification and by travel. & Celestia \\
\hyperref[task:sb-celestia-005]{\texttt{sb\_celestia\_005}} & Compare the surface speeds of two slow-turning planets. & Celestia \\
\hyperref[task:sb-celestia-006]{\texttt{sb\_celestia\_006}} & Date the light received from Neptune. & Celestia \\
\hyperref[task:sb-celestia-007]{\texttt{sb\_celestia\_007}} & View the Summer Triangle from another star. & Celestia \\
\hyperref[task:sb-celestia-008]{\texttt{sb\_celestia\_008}} & Distinguish orbital direction from rotating-frame motion. & Celestia \\
\hyperref[task:sb-celestia-009]{\texttt{sb\_celestia\_009}} & Estimate the selection bias of a magnitude-limited stellar census. & Celestia \\
\hyperref[task:sb-celestia-010]{\texttt{sb\_celestia\_010}} & Find the latitude where Capella stops setting. & Celestia \\
\hyperref[task:sb-chimerax-001]{\texttt{sb\_chimerax\_001}} & Separate hinge motion from scaffold displacement. & ChimeraX \\
\hyperref[task:sb-chimerax-002]{\texttt{sb\_chimerax\_002}} & Compare two alternate oxygen positions. & ChimeraX \\
\hyperref[task:sb-chimerax-003]{\texttt{sb\_chimerax\_003}} & Screen high-residual sites on a shared scale. & ChimeraX \\
\hyperref[task:sb-chimerax-004]{\texttt{sb\_chimerax\_004}} & Map contacts across two toy edges. & ChimeraX \\
\hyperref[task:sb-chimerax-005]{\texttt{sb\_chimerax\_005}} & Find the closest sampled approach of two moving markers. & ChimeraX \\
\hyperref[task:sb-chimerax-007]{\texttt{sb\_chimerax\_007}} & Measure the tilt and probe clearance of two fitted panels. & ChimeraX \\
\hyperref[task:sb-chimerax-008]{\texttt{sb\_chimerax\_008}} & Rotate one linker bond while keeping its anchor fixed. & ChimeraX \\
\hyperref[task:sb-grassgis-001]{\texttt{sb\_grassgis\_001}} & Meadow access with a wet-ground exclusion. & GRASS GIS \\
\hyperref[task:sb-grassgis-002]{\texttt{sb\_grassgis\_002}} & Local height contrast without bridging survey gaps. & GRASS GIS \\
\hyperref[task:sb-grassgis-003]{\texttt{sb\_grassgis\_003}} & Map the light sensors needing a repeat visit. & GRASS GIS \\
\hyperref[task:sb-grassgis-004]{\texttt{sb\_grassgis\_004}} & Coarsen a sprinkler map without changing what its values mean. & GRASS GIS \\
\hyperref[task:sb-grassgis-005]{\texttt{sb\_grassgis\_005}} & Recover slope and downhill direction from a centimetre DEM. & GRASS GIS \\
\hyperref[task:sb-grassgis-006]{\texttt{sb\_grassgis\_006}} & Map a cart's twenty-second service reach. & GRASS GIS \\
\hyperref[task:sb-grassgis-007]{\texttt{sb\_grassgis\_007}} & Retain connected habitat patches large enough for a refuge trial. & GRASS GIS \\
\hyperref[task:sb-grassgis-008]{\texttt{sb\_grassgis\_008}} & Account for cut and fill only where both surveys exist. & GRASS GIS \\
\hyperref[task:sb-grassgis-009]{\texttt{sb\_grassgis\_009}} & Map redundant sightlines across a teaching plot. & GRASS GIS \\
\hyperref[task:sb-kalgebra-001]{\texttt{sb\_kalgebra\_001}} & Minimize coating cost for an open channel. & KAlgebra \\
\hyperref[task:sb-kalgebra-002]{\texttt{sb\_kalgebra\_002}} & Overlay a delayed actuator pulse. & KAlgebra \\
\hyperref[task:sb-kalgebra-003]{\texttt{sb\_kalgebra\_003}} & Orient a corrugated-sheet surface. & KAlgebra \\
\hyperref[task:sb-kalgebra-004]{\texttt{sb\_kalgebra\_004}} & Recover inputs from an ambiguous calibration. & KAlgebra \\
\hyperref[task:sb-kalgebra-005]{\texttt{sb\_kalgebra\_005}} & Separate travel along a guide from transverse drift. & KAlgebra \\
\hyperref[task:sb-kalgebra-006]{\texttt{sb\_kalgebra\_006}} & Find the first successful dilution cycle. & KAlgebra \\
\hyperref[task:sb-kalgebra-007]{\texttt{sb\_kalgebra\_007}} & Account for recording bias in bead energy. & KAlgebra \\
\hyperref[task:sb-kalgebra-008]{\texttt{sb\_kalgebra\_008}} & Construct a sideways reflector from its focus. & KAlgebra \\
\hyperref[task:sb-kalgebra-009]{\texttt{sb\_kalgebra\_009}} & Fit a colorimeter calibration with a free baseline. & KAlgebra \\
\hyperref[task:sb-kalgebra-010]{\texttt{sb\_kalgebra\_010}} & Compare interior and transition errors in harmonic reconstruction. & KAlgebra \\
\hyperref[task:sb-lean-001]{\texttt{sb\_lean\_001}} & Prove a weighted energy bound and characterize equality. & Lean \\
\hyperref[task:sb-lean-002]{\texttt{sb\_lean\_002}} & Prove rollback for shifts and reflections. & Lean \\
\hyperref[task:sb-lean-003]{\texttt{sb\_lean\_003}} & Characterize the outputs of two compatible projections. & Lean \\
\hyperref[task:sb-lean-004]{\texttt{sb\_lean\_004}} & Show that count mixing preserves the greatest common divisor. & Lean \\
\hyperref[task:sb-lean-005]{\texttt{sb\_lean\_005}} & Account for route costs when every outlet is expanded. & Lean \\
\hyperref[task:sb-lean-006]{\texttt{sb\_lean\_006}} & Split an integer scan without skipped or repeated ticks. & Lean \\
\hyperref[task:sb-lean-007]{\texttt{sb\_lean\_007}} & Bound the accumulated error of a disturbed iteration. & Lean \\
\hyperref[task:sb-texstudio-001]{\texttt{sb\_texstudio\_001}} & Two coupled carts: calculate and typeset the normal modes. & TeXstudio \\
\hyperref[task:sb-texstudio-002]{\texttt{sb\_texstudio\_002}} & Show the torsion of a toy four-carbon chain. & TeXstudio \\
\hyperref[task:sb-texstudio-003]{\texttt{sb\_texstudio\_003}} & Report cylinder density with shared calibration uncertainty. & TeXstudio \\
\hyperref[task:sb-texstudio-004]{\texttt{sb\_texstudio\_004}} & Make an exchange note survive renumbering. & TeXstudio \\
\hyperref[task:sb-texstudio-005]{\texttt{sb\_texstudio\_005}} & Repair the citations in a thermometer lesson. & TeXstudio \\
\end{longtable}
\endgroup

\subsection{Task Instructions}
\label{sec:scienceboard-synthetic-instructions}

The runner appends the same execution reminder as for the OSWorld V2
synthetic tasks (Section~\ref{sec:synthetic-task-instructions}).

% sb_celestia_001 instruction SHA-256: 3653ab5596dd5de23ff45fc26e6641d78f12c329210b5142859688c85f10ae33
\subsubsection{\texttt{sb\_celestia\_001} One Moon, two viewpoints}
\label{task:sb-celestia-001}
\noindent\textit{Applications.} Celestia.\par\nopagebreak[4]
\begingroup
\small
\setlength{\parindent}{0pt}
\setlength{\parskip}{0.3em}
Use Celestia to prepare a two-panel illustration of how the Moon's apparent phase depends on the observer's position. Pause time and show the Moon in two side-by-side views at that same instant. In the left panel, make it clearly crescent: less than half of the disk illuminated. In the right panel, make it clearly gibbous: more than half illuminated, but not full. Change the viewpoints rather than advancing time. Keep the entire lunar disk inside each panel at approximately the same apparent diameter. Leave the comparison open and explain in one sentence why the illuminated fractions differ.
\par\endgroup

% sb_celestia_002 instruction SHA-256: b8e04ebb6942f1f752568d985f74cacb0de0b9b799687b92feea43867438eb36
\subsubsection{\texttt{sb\_celestia\_002} A fixed coastline under changing daylight}
\label{task:sb-celestia-002}
\noindent\textit{Applications.} Celestia.\par\nopagebreak[4]
\begingroup
\small
\setlength{\parindent}{0pt}
\setlength{\parskip}{0.3em}
In Celestia, set up a running day/night view of Earth in which the Australian mainland stays in the middle of the disk as simulation time advances. Show the whole Earth disk, turn clouds off, and keep the day/night shading clearly visible. Leave time running forward at 1000$\times$. The coastline must stay in the same position on screen without further camera adjustments while its illumination changes.
\par\endgroup

% sb_celestia_003 instruction SHA-256: e19f3cbf608451578427b036811c6ae22f0c3e18c8342d8b111981a8ec7b50d8
\subsubsection{\texttt{sb\_celestia\_003} Return to a ring-shadow observing point}
\label{task:sb-celestia-003}
\noindent\textit{Applications.} Celestia.\par\nopagebreak[4]
\begingroup
\small
\setlength{\parindent}{0pt}
\setlength{\parskip}{0.3em}
Create a reusable observing point in Celestia for studying Saturn's shadow on its rings. Set simulation time to 2021-01-01 00:00:00 UTC and pause it. Compose an oblique view with the whole planet and the outer edges of the rings inside the window, making the dark shadow cast by Saturn onto the rings clearly visible. Save this observer state as a native bookmark named 'Ring-shadow observing point', with the saved simulation time. To check that it can reproduce the observation, go to Neptune and change simulation time to 2031-01-01 00:00:00 UTC, then use the bookmark to return. Leave the restored Saturn view visible and paused at the saved 2021 time, with the same framing and shadow geometry.
\par\endgroup

% sb_celestia_004 instruction SHA-256: 59a1a7a172c4710e66abfbf0ff2cfc8d9feaf8b4efccc6e603308e5fb5893da8
\subsubsection{\texttt{sb\_celestia\_004} Enlarge Mars by magnification and by travel}
\label{task:sb-celestia-004}
\noindent\textit{Applications.} Celestia.\par\nopagebreak[4]
\begingroup
\small
\setlength{\parindent}{0pt}
\setlength{\parskip}{0.3em}
Use Celestia to compare two ways of making Mars appear larger. Pause time and prepare one centered, mostly sunlit view of Mars with a field of view (FoV) of 45 degrees. Adjust the observer distance so the whole disk is about one quarter of the graphics viewport's height; this is your baseline. First, keep the observer at that position and change only FoV until the disk is about half the viewport height. Next, restore FoV to 45 degrees and change only the distance to Mars to obtain a half-height disk again. After establishing the baseline, keep simulation time, viewing direction, and window dimensions fixed, with Mars centered throughout. Report the displayed FoV and Mars distance, including units, for the baseline and each of the two enlarged views. Use these readings to explain briefly how the two enlargement procedures differ. Leave the final 45-degree view open with the whole disk and the FoV and distance readouts visible.
\par\endgroup

% sb_celestia_005 instruction SHA-256: a6fab291c5f9ef288cc250c512d03b5664eb6056101ff60da98fb59c6ffd1d91
\subsubsection{\texttt{sb\_celestia\_005} Compare the surface speeds of two slow-turning planets}
\label{task:sb-celestia-005}
\noindent\textit{Applications.} Celestia.\par\nopagebreak[4]
\begingroup
\small
\setlength{\parindent}{0pt}
\setlength{\parskip}{0.3em}
Use Celestia's built-in object information to compare the rotational surface speeds at the equators of Mercury and Venus. Treat each planet as a sphere with Celestia's displayed radius, and use its displayed sidereal rotation period. Compare speed magnitudes, ignoring the direction of rotation and any motion of clouds. Report the radius and rotation-period readings for both planets with their units, then calculate the dimensionless ratio \path{v_Venus} / \path{v_Mercury} to two significant figures. Show the relation you used and identify which planet has the faster equatorial surface despite the difference in their sizes. Base the comparison on Celestia's displayed values. Leave the faster planet selected and in view, with its radius and rotation period visible in the object information.
\par\endgroup

% sb_celestia_006 instruction SHA-256: c46f1433ebe8bf23e498475736dda3934609769b2802de6f8969c4b75dade5b2
\subsubsection{\texttt{sb\_celestia\_006} Date the light received from Neptune}
\label{task:sb-celestia-006}
\noindent\textit{Applications.} Celestia.\par\nopagebreak[4]
\begingroup
\small
\setlength{\parindent}{0pt}
\setlength{\parskip}{0.3em}
Use Celestia to work out when light arriving near Earth just after midnight left Neptune. Set simulation time to 2025-01-01 00:10:00 UTC and pause it. Go to Earth and use that arrival viewpoint as the observing position. Select Neptune without travelling to it, and read Celestia's light-travel delay from the observer to Neptune. Treat that displayed delay as constant for this exercise. Report the displayed delay with its units and the corresponding light-emission date and time in UTC, rounded to the nearest minute. Briefly explain the relation between the reception and emission times. Set the simulation clock to the emission time you reported and leave Celestia paused with Neptune selected and the UTC date/time readout visible.
\par\endgroup

% sb_celestia_007 instruction SHA-256: 038d23db0b3db4cbeb316e6ca70912a0c51d162d9f6b4f99112289729672bb8d
\subsubsection{\texttt{sb\_celestia\_007} View the Summer Triangle from another star}
\label{task:sb-celestia-007}
\noindent\textit{Applications.} Celestia.\par\nopagebreak[4]
\begingroup
\small
\setlength{\parindent}{0pt}
\setlength{\parskip}{0.3em}
Use Celestia to investigate whether the Summer Triangle keeps its apparent shape when the observer moves between stars. Its three members are Vega, Deneb, and Altair. Pause time and use Go To Sol to establish the first observing position. Record the displayed distances from there to all three triangle stars, including units. Prepare two equal-sized views at this same paused instant: one at the Sol arrival position and one at the arrival position after Go To Capella. Show all three triangle stars with readable names in both views, using the same field of view. Keep Vega centered in both panels, and orient the Deneb-to-Vega direction the same way on screen in both views. Framing changes must rotate the camera without moving either observing position. Keep constellation diagrams and boundaries off. Compare the triangle's shape, allowing for an overall rotation or uniform enlargement: does its relative spacing stay the same? Support your answer with a specific comparison of one named star pair's separation relative to another pair's separation, then use the three distance readings to explain your observation. A qualitative spacing comparison is sufficient. Leave the two views visible and time paused.
\par\endgroup

% sb_celestia_008 instruction SHA-256: 6dc9825cb7321e2374e146645eec4107aee02edf6d6ef93b1a44d8ab0a626d9b
\subsubsection{\texttt{sb\_celestia\_008} Distinguish orbital direction from rotating-frame motion}
\label{task:sb-celestia-008}
\noindent\textit{Applications.} Celestia.\par\nopagebreak[4]
\begingroup
\small
\setlength{\parindent}{0pt}
\setlength{\parskip}{0.3em}
Use Celestia to investigate whether a moon's apparent motion around Mars can be mistaken for a retrograde orbit. Start at 2025-03-01 00:00:00 UTC with time paused. Prepare two equal side-by-side views looking down on Mars's north pole, with Mars centered and the complete orbits of Phobos and Deimos in view. Use matching initial observer distances and fields of view. Show both moon names, both orbit paths, and Mars's planetographic grid. In the left panel, make the observer travel with Mars while its camera axes remain fixed relative to distant stars. In the right panel, make the observer rotate with Mars so its longitude grid stays fixed on screen. Do not track either moon. Run the shared simulation time forward at 10000$\times$ and watch at least 24 simulated hours without adjusting either camera. For each moon, report its clockwise or counterclockwise motion in each panel, and decide whether its actual orbit is prograde or retrograde relative to Mars's spin. Explain the observations briefly in terms of the moons' angular motion relative to the rotating planet. Add a visible object marker only to each moon whose apparent clockwise/counterclockwise sense changes between the two panels. Leave both views visible and running forward at 10000$\times$, with the moon names, orbit paths, and grid still shown.
\par\endgroup

% sb_celestia_009 instruction SHA-256: 45d658ca24c364b8fd48eb039c4888b1401fe4fcc90a727f973a2a56f4370f34
\subsubsection{\texttt{sb\_celestia\_009} Estimate the selection bias of a magnitude-limited stellar census}
\label{task:sb-celestia-009}
\noindent\textit{Applications.} Celestia.\par\nopagebreak[4]
\begingroup
\small
\setlength{\parindent}{0pt}
\setlength{\parskip}{0.3em}
Use Celestia to compare two idealized stellar populations whose members have the same absolute magnitude as Altair or Deneb, respectively. A hypothetical all-sky survey detects a star when its apparent magnitude is at most 6.0. Assume the two populations have equal, uniform number densities; ignore extinction and Galactic boundaries, and treat the catalog magnitudes as fixed. Pause time and use Go To Sol to establish the observing position. Keep that position throughout the task. In the Star Browser, obtain Altair's and Deneb's distance, apparent magnitude, and absolute magnitude; report all six readings with distance units. Identify which prototype appears brighter from this position. Using the absolute magnitudes, calculate each population's maximum detection distance in parsecs and the expected detected-number ratio \path{N_Deneb-like} / \path{N_Altair-like}. Give distances and the ratio to two significant figures, using unrounded intermediate calculations, and state the relations you used. Briefly explain why the ordering by apparent brightness here need not match the ordering by expected survey detections. Leave the Star Browser visible in Brightest mode, sorted by absolute magnitude from brighter to fainter, with the prototype representing the more numerous detected population selected. Keep time paused and the observer at the Sol arrival position.
\par\endgroup

% sb_celestia_010 instruction SHA-256: b9d4b8acf7cedfa4535ebe868c15995eaa077d51157127d7f2ba88a605a6d34b
\subsubsection{\texttt{sb\_celestia\_010} Find the latitude where Capella stops setting}
\label{task:sb-celestia-010}
\noindent\textit{Applications.} Celestia.\par\nopagebreak[4]
\begingroup
\small
\setlength{\parindent}{0pt}
\setlength{\parskip}{0.3em}
Use Celestia to determine how far north an observer must be for Capella to become circumpolar: never below the ideal horizon during a daily circuit. Compare two observing sites on Earth's surface, both at longitude 0 degrees, at latitudes 30 degrees N and 60 degrees N. At each site, inspect Capella throughout the simulated interval from 2025-01-01 00:00:00 UTC to 2025-01-02 00:00:00 UTC, keeping the observer fixed to that surface location. You may turn the camera in place. Show the horizontal coordinate grid and turn off atmospheres and clouds. Use the grid's altitude-zero circle as the ideal horizon; ignore refraction and terrain. For each site, report whether Capella crosses below that horizon or stays above it throughout the circuit. At 60 degrees N, estimate its lowest altitude to the nearest 5 degrees. Use that estimate to infer, also to the nearest 5 degrees, the northern latitude where Capella's daily circle would just graze the ideal horizon, and briefly explain the geometric relation used. Leave Celestia paused near Capella's lowest altitude during the specified interval at the 60-degree-N site, with Capella selected and named, the northern horizon and horizontal grid readable, and the observer still fixed to Earth.
\par\endgroup

% sb_chimerax_001 instruction SHA-256: 155199739d0130366100cef1105527a6faa7f2a2c005c3cb80dd09fdbef548f4
\subsubsection{\texttt{sb\_chimerax\_001} Separate hinge motion from scaffold displacement}
\label{task:sb-chimerax-001}
\noindent\textit{Applications.} ChimeraX.\par\nopagebreak[4]
\begingroup
\small
\setlength{\parindent}{0pt}
\setlength{\parskip}{0.3em}
The two open models, \path{support_reference.pdb} and \path{support_moving.pdb}, are synthetic C-alpha point scaffolds for a geometry exercise, not real proteins. In both, chain A residues 1-7 form a fixed support; chain B residues 1-5 form a lever whose tip is the CA atom of B:5. Superpose the moving model onto the reference using all seven corresponding chain-A CA atoms only. Apply one rigid transformation to the whole moving model, preserving its internal geometry. Then measure the distance between the two B:5 CA tips in this shared support frame. Report that displacement to the nearest 0.01 \AA{} and briefly explain why the lever atoms were excluded from the fit. Leave both full scaffolds visible in distinct colors, with the two tip atoms shown as spheres and a readable distance monitor between them. Save the comparison as \path{/home/user/ScienceBoardSynthetic/sb_chimerax_001/hinge_comparison.cxs}.
\par\endgroup

% sb_chimerax_002 instruction SHA-256: 18d180c6797f368a66a5bbded6c72627135097cc159018af2af48b2a1703ded0
\subsubsection{\texttt{sb\_chimerax\_002} Compare two alternate oxygen positions}
\label{task:sb-chimerax-002}
\noindent\textit{Applications.} ChimeraX.\par\nopagebreak[4]
\begingroup
\small
\setlength{\parindent}{0pt}
\setlength{\parskip}{0.3em}
The open \path{alternate_site.pdb} contains invented serine and water geometry for an alternate-location exercise, not an experimental structure. For serine A:7, compare alternate locations A and B of OG with the fixed water oxygen W:1@O. In ChimeraX, report each location's occupancy and its OG-to-water-oxygen distance to the nearest 0.01 \AA{}, then state whether the higher-occupancy location is also the closer one. Leave the closer OG location active in the original model with a readable distance monitor to the water oxygen. Preserve both stored alternate locations and the original coordinates.
\par\endgroup

% sb_chimerax_003 instruction SHA-256: 77abd1617c51c066a660e3189dfd71b5f86d4e1db4b09dcb2a970ef2b27d2c1e
\subsubsection{\texttt{sb\_chimerax\_003} Screen high-residual sites on a shared scale}
\label{task:sb-chimerax-003}
\noindent\textit{Applications.} ChimeraX.\par\nopagebreak[4]
\begingroup
\small
\setlength{\parindent}{0pt}
\setlength{\parskip}{0.3em}
The open \path{array_a.pdb} and \path{array_b.pdb} are invented point arrays for a data-screening exercise, not molecules or experimental structures. Each carbon marker represents one numbered site. Its PDB B-factor stores a synthetic residual magnitude in arbitrary units, and its occupancy stores a sampling-coverage fraction. In ChimeraX, prepare a comparison with all sites shown as spheres and colored by residual using the same continuous scale in both arrays: blue at 0, white at 6, and red at 12, with linear color interpolation between these values. Add a readable matching color key. Select exactly the sites across both arrays whose residual is at least 8 and whose coverage is at least 0.80; high residual alone does not qualify a poorly sampled site. Leave both complete arrays visible with this selection highlighted, preserving their coordinates and stored values.
\par\endgroup

% sb_chimerax_004 instruction SHA-256: cb601130276b095346505c907ac1b086bf5b13feb3e22f6c51289357ee576161
\subsubsection{\texttt{sb\_chimerax\_004} Map contacts across two toy edges}
\label{task:sb-chimerax-004}
\noindent\textit{Applications.} ChimeraX.\par\nopagebreak[4]
\begingroup
\small
\setlength{\parindent}{0pt}
\setlength{\parskip}{0.3em}
The open \path{edge_markers.pdb} contains two invented rows of carbon markers, chains A and B, for a geometry exercise; these are not molecules or experimental observations. Each numbered residue is one site. In ChimeraX, find every cross-chain contact using a center-to-center distance of at most 3.50 \AA{} as the sole criterion. Exclude pairs within the same chain and do not use van der Waals overlap as the cutoff. Show all markers as spheres, distinguish the chains by color, and display only the qualifying cross-chain contacts as dashed pseudobonds with readable distance labels. Select all participating sites and leave both complete rows visible without changing coordinates. Report the number of contact pairs and the number of distinct participating sites in each chain; a site with several partners counts only once in its chain's site total.
\par\endgroup

% sb_chimerax_005 instruction SHA-256: e7d9523649a75f1e0d42a76403843c2fc707126bda47b2d7d18d4d5c54506556
\subsubsection{\texttt{sb\_chimerax\_005} Find the closest sampled approach of two moving markers}
\label{task:sb-chimerax-005}
\noindent\textit{Applications.} ChimeraX.\par\nopagebreak[4]
\begingroup
\small
\setlength{\parindent}{0pt}
\setlength{\parskip}{0.3em}
In ChimeraX, open \path{/home/user/ScienceBoardSynthetic/sb_chimerax_005/gate_sweep.pdb} as one trajectory with nine coordinate sets. This is invented point geometry, not a molecule, experimental structure, or physical dynamics simulation. The C atoms of chain A residue 1 and chain B residue 1 are the two moving tips; the other four markers are fixed reference points. Inspect the tip-to-tip separation across all nine sampled frames. Report the frame with the smallest sampled separation, its labelled time, and its separation to the nearest 0.01 \AA{}. Also report every sampled frame with separation strictly below 4.50 \AA{}. Frame 1 is labelled 0 ps and successive saved frames are 2 ps apart; do not infer an unsampled minimum or a continuous threshold-crossing time. Export all six markers from only the minimum-separation frame to \path{/home/user/ScienceBoardSynthetic/sb_chimerax_005/closest_snapshot.pdb}. Leave the full trajectory loaded and paused at that frame, with all six markers visible, the frame slider visible, and a readable distance monitor between the two tips. Preserve all input coordinates.
\par\endgroup

% sb_chimerax_007 instruction SHA-256: a0710b2ebe9666e2999908ea7b5ce7a5179a18a457115550084d6fffc3645a7d
\subsubsection{\texttt{sb\_chimerax\_007} Measure the tilt and probe clearance of two fitted panels}
\label{task:sb-chimerax-007}
\noindent\textit{Applications.} ChimeraX.\par\nopagebreak[4]
\begingroup
\small
\setlength{\parindent}{0pt}
\setlength{\parskip}{0.3em}
The open \path{panel_markers.pdb} contains invented point geometry, not molecules or experimental observations. Chains A and B each contain nine carbon markers sampling a slightly nonplanar panel; P:1@C is a separate probe. In ChimeraX, create one native fitted plane for each panel using all nine of that chain's markers with equal weight. Use orthogonal least-squares fitting, minimizing squared perpendicular distances, and exclude the probe from both fits. Report the smaller angle between the two fitted planes in degrees to one decimal place. Measure the unsigned perpendicular distance from P:1@C to each fitted plane, report both to the nearest 0.01 \AA{}, and identify the nearer plane. Treat the planes as infinite for these measurements; the displayed disk rims are not panel boundaries. Leave both fitted plane disks and all 19 markers visible in a useful oblique view, with each panel and its plane sharing a distinct color and the probe visually distinct. Keep the Axes/Planes/Centroids tool open and preserve the input coordinates.
\par\endgroup

% sb_chimerax_008 instruction SHA-256: c13554707acbf18908e05fc30d2678e0c9576cabcdb2527bc558444ba02d56a0
\subsubsection{\texttt{sb\_chimerax\_008} Rotate one linker bond while keeping its anchor fixed}
\label{task:sb-chimerax-008}
\noindent\textit{Applications.} ChimeraX.\par\nopagebreak[4]
\begingroup
\small
\setlength{\parindent}{0pt}
\setlength{\parskip}{0.3em}
The open \path{torsion_linker.pdb} is an invented seven-carbon heavy-atom linkage for a geometry exercise, not an experimental or energy-minimized molecule. Its single residue A:1 contains the bonded chain C1-C2-C3-C4-C5-C6-C7. In ChimeraX, measure the starting C1-to-C7 distance, then set the ordered C2-C3-C4-C5 dihedral to +75.0 degrees by rotating about C3-C4. Keep C1, C2, and C3 at their original coordinates; rotate the entire C4-C7 side together about that bond axis, preserving all bond lengths, bond angles, and other internal torsions. Keep exactly the seven supplied atoms and the six supplied bonds. Report the starting and final C1-to-C7 distances to the nearest 0.01 \AA{} and state whether this rotation lengthened or shortened the end-to-end separation. Save the edited linkage in its original model coordinate frame as \path{/home/user/ScienceBoardSynthetic/sb_chimerax_008/linker_75.pdb}. Leave the edited linkage fully visible in ball-and-stick style with readable atom-name labels and a C1-to-C7 distance monitor. Keep Build Structure's Adjust Torsions section open with C3-C4 active, C2 shown as Fixed, C5 shown as Moving, and the final signed angle displayed.
\par\endgroup

% sb_grassgis_001 instruction SHA-256: 920ea27746ad6d57a4a6a7d92e68f7eef89920c31af5632a7571c8d3add8b586
\subsubsection{\texttt{sb\_grassgis\_001} Meadow access with a wet-ground exclusion}
\label{task:sb-grassgis-001}
\noindent\textit{Applications.} GRASS GIS.\par\nopagebreak[4]
\begingroup
\small
\setlength{\parindent}{0pt}
\setlength{\parskip}{0.3em}
In this fictional meadow study, equipment can reach ground within 35 metres of the \path{service_track} centreline, but must stay inside meadow and outside \path{wet_ground}. Using the three existing vector maps, create \path{accessible_meadow} containing all qualifying ground, including any disconnected pieces. Use planar distances in the project CRS, with rounded buffer corners and end caps. Save the result in the current work mapset and display it with the meadow outline and service track. Report the total qualifying area in hectares, rounded to two decimal places. The project uses EPSG:32631 (WGS 84 / UTM zone 31N), with coordinates in metres; all geometry is invented toy data.
\par\endgroup

% sb_grassgis_002 instruction SHA-256: 075b7d42ca1bf91c945f2de2ea454f03e7a22cdf3c4652c1fd4055baebef8cae
\subsubsection{\texttt{sb\_grassgis\_002} Local height contrast without bridging survey gaps}
\label{task:sb-grassgis-002}
\noindent\textit{Applications.} GRASS GIS.\par\nopagebreak[4]
\begingroup
\small
\setlength{\parindent}{0pt}
\setlength{\parskip}{0.3em}
The existing \path{pad_height} raster is an invented survey of a small test pad. Create a floating-point raster named \path{neighbor_relief} in the current work mapset: for each cell, subtract the arithmetic mean of its eight immediately surrounding cells (four edge neighbors and four diagonal neighbors) from that cell's height. Exclude the center cell from the mean. Evaluate a cell only when its height and all eight neighboring heights are known; otherwise write NULL, including at the outer boundary. A valid zero contrast must remain zero. Keep the input's full extent, alignment, and 2 m cell size. Heights are in metres above an arbitrary local datum, and the resulting signed contrast is in metres; positive contrast means locally raised ground. The horizontal CRS is EPSG:32631, with coordinates in metres. Save \path{neighbor_relief} as a native raster and leave it displayed in GRASS GIS.
\par\endgroup

% sb_grassgis_003 instruction SHA-256: c83f2e54ef816142bda2ca6852bacf093c3187a2c5c6253388f34fb890a3f298
\subsubsection{\texttt{sb\_grassgis\_003} Map the light sensors needing a repeat visit}
\label{task:sb-grassgis-003}
\noindent\textit{Applications.} GRASS GIS.\par\nopagebreak[4]
\begingroup
\small
\setlength{\parindent}{0pt}
\setlength{\parskip}{0.3em}
Import \path{/home/user/ScienceBoardSynthetic/sb_grassgis_003/station_counts.csv} into GRASS GIS as 2D station points, then create a native point vector named \path{revisit_stations} containing exactly the stations with qc = valid, a strictly positive baseline count, and a repeat count at least 20\% lower than the baseline count. Include the 20\% boundary. Both pulse counts cover equal 60-second intervals; zero is a recorded value. The CSV lists northing before easting: use \path{easting_m} for x and \path{northing_m} for y. Coordinates are metres in the current EPSG:32631 (WGS 84 / UTM zone 31N) project, and all data are invented. Preserve the selected stations' original coordinates, station identifiers, counts, and quality flags in the output. Save \path{revisit_stations} in the current work mapset and leave it displayed with all its points in view.
\par\endgroup

% sb_grassgis_004 instruction SHA-256: d1baf26e16d0507f74989d706181fa6b0a9cff57cf0cc027e7a2485880d2e130
\subsubsection{\texttt{sb\_grassgis\_004} Coarsen a sprinkler map without changing what its values mean}
\label{task:sb-grassgis-004}
\noindent\textit{Applications.} GRASS GIS.\par\nopagebreak[4]
\begingroup
\small
\setlength{\parindent}{0pt}
\setlength{\parskip}{0.3em}
The existing \path{wetting_depth} and \path{surface_code} rasters describe an invented sprinkler plot on matching 2 m grids. Create native rasters \path{wetting_depth_6m} and \path{surface_code_6m} on a common 6 m square-cell grid covering exactly the original footprint: west 500001, east 500025, south 5100007, north 5100025. Do not expand or shift the footprint. Each \path{wetting_depth} value is water depth in millimetres averaged over its source cell, not a per-cell volume. The coarser depth must represent the mean depth over each target cell and preserve the total represented water volume across the plot, without rounding depths to integers. Each \path{surface_code} value is a nominal class: 2 = sand, 5 = turf, 9 = paving. For each target cell, retain the code of the surface occupying the greatest area within it; do not blend the codes. There are no missing cells or tied dominant classes. Save both results in the current work mapset and leave \path{surface_code_6m} displayed with the entire plot in view. The project uses EPSG:32631 (WGS 84 / UTM zone 31N), with horizontal coordinates in metres; use planar cell areas.
\par\endgroup

% sb_grassgis_005 instruction SHA-256: 91b578659288e69d2b149e23a0da48afaec233c1b24835d4d0b4dad8321ae7ec
\subsubsection{\texttt{sb\_grassgis\_005} Recover slope and downhill direction from a centimetre DEM}
\label{task:sb-grassgis-005}
\noindent\textit{Applications.} GRASS GIS.\par\nopagebreak[4]
\begingroup
\small
\setlength{\parindent}{0pt}
\setlength{\parskip}{0.3em}
The existing \path{training_height_cm} raster is an invented earthwork surface, with elevations in centimetres above an arbitrary local datum. The horizontal CRS is EPSG:32631 (WGS 84 / UTM zone 31N), with coordinates and 2 m cell sizes in metres. Create two native floating-point rasters in the current work mapset using GRASS GIS's Horn 3-by-3 terrain-derivative method: \path{slope_percent}, containing slope as percent rise, and \path{downhill_azimuth}, containing the direction of steepest descent in degrees clockwise from north. Account for the different vertical and horizontal units. Azimuths must lie in [0,360), with north = 0, east = 90, south = 180, and west = 270; use -9999 for exactly flat cells, whose slope must be zero. Preserve the input's full extent, alignment, and 2 m resolution. Leave the outermost cell ring NULL in both outputs, without estimating edge derivatives. Save both rasters and leave \path{slope_percent} displayed with the entire plot in view.
\par\endgroup

% sb_grassgis_006 instruction SHA-256: 7a8f815335f42ae63baafa6a1325c6f457e60de7b96296f55d521d915ef1becd
\subsubsection{\texttt{sb\_grassgis\_006} Map a cart's twenty-second service reach}
\label{task:sb-grassgis-006}
\noindent\textit{Applications.} GRASS GIS.\par\nopagebreak[4]
\begingroup
\small
\setlength{\parindent}{0pt}
\setlength{\parskip}{0.3em}
The existing \path{cell_time} and depot rasters describe an invented cart yard on matching 2 m square-cell grids. Create a native raster named \path{reachable_20s} marking every cell whose centre can be reached from the centre of the sole non-NULL depot cell with a minimum accumulated travel time of at most 20 seconds. Set qualifying cells to 1 and all other cells to NULL, including barriers and disconnected ground. The depot starts at elapsed time zero and must be included. Each \path{cell_time} value is the time in seconds to traverse one whole cell orthogonally, not seconds per metre. Use the standard eight-neighbor movement model: an edge-neighbor move costs the arithmetic mean of its two endpoint values, and a diagonal move costs that mean times sqrt(2). A move is allowed whenever both endpoint cells are non-NULL, including diagonals past barrier corners; do not use knight's moves. NULL \path{cell_time} cells are impassable, and travel must remain within the input footprint. Minimize travel time rather than geometric distance, and apply the inclusive 20-second limit without rounding accumulated times. Preserve the input extent, alignment, and resolution. Save \path{reachable_20s} in the current work mapset and leave it displayed with the entire yard in view. The project uses EPSG:32631 (WGS 84 / UTM zone 31N), with horizontal coordinates in metres; all data are fictional.
\par\endgroup

% sb_grassgis_007 instruction SHA-256: c0c9257bb64b9964c16552cfb1b9fecdac4df18ef7685ea5d7b2472ad5aa5aa6
\subsubsection{\texttt{sb\_grassgis\_007} Retain connected habitat patches large enough for a refuge trial}
\label{task:sb-grassgis-007}
\noindent\textit{Applications.} GRASS GIS.\par\nopagebreak[4]
\begingroup
\small
\setlength{\parindent}{0pt}
\setlength{\parskip}{0.3em}
The existing \path{land_cover} raster represents an invented refuge trial plot on a 2 m square-cell grid. Codes 3 (grass) and 7 (shrubs) both count as habitat; code 0 is bare ground, code 8 is paving, and NULL means unsurveyed. Create a native integer raster named \path{refuge_patches} containing only habitat patches with a planar area of at least 48 square metres, including exactly 48. Treat grass and shrubs as one habitat class when deciding connectivity: cells belong to the same patch if linked through habitat cells sharing an edge. Corner contact alone does not connect patches. Give every retained patch its own positive integer ID, with the same ID throughout that patch; the actual ID numbers are your choice. All other cells must be NULL. Count only habitat cells toward patch area, preserve holes and gaps, and remove small patches without expanding or merging them. Keep the input's full extent, alignment, and 2 m resolution. Save \path{refuge_patches} in the current work mapset and leave it displayed with the entire plot in view. The project uses EPSG:32631 (WGS 84 / UTM zone 31N), with coordinates in metres. All data and the refuge criterion are fictional teaching material.
\par\endgroup

% sb_grassgis_008 instruction SHA-256: 2416223ee708a93d4e060997638f3b81f2a5af8413a3202063695dc524cec49c
\subsubsection{\texttt{sb\_grassgis\_008} Account for cut and fill only where both surveys exist}
\label{task:sb-grassgis-008}
\noindent\textit{Applications.} GRASS GIS.\par\nopagebreak[4]
\begingroup
\small
\setlength{\parindent}{0pt}
\setlength{\parskip}{0.3em}
The existing \path{before_height} and \path{after_height} rasters describe an invented earthwork trial. Both give cell-average ground height in metres above the same arbitrary local datum on matching 2 m square-cell grids. Create a native floating-point raster named \path{change_m} containing after minus before, retaining a value only where both heights are known. Keep valid zero change as zero and all other cells NULL; preserve the input extent, alignment, and resolution. Then create a native area vector named \path{block_budget} with the same three polygons and category IDs as \path{work_blocks}. For each block, add numeric attributes \path{surveyed_m2} (the area with both heights known), \path{cut_m3} (positive volume removed where ground became lower), \path{fill_m3} (positive volume added where ground became higher), and \path{net_m3} (signed fill minus cut). Integrate height change over planar cell area, treating each cell average as constant across its cell. The block edges follow grid edges, so include whole cells only. Use the same jointly surveyed cells for all four attributes, including unchanged cells in \path{surveyed_m2}; do not fill survey gaps or extrapolate to unsurveyed parts of a block. Ignore soil bulking and compaction, and keep full calculation precision in the stored attributes. Save both results in the current work mapset and leave \path{change_m} displayed with block outlines and the \path{block_budget} attribute table open. The horizontal CRS is EPSG:32631 (WGS 84 / UTM zone 31N), with coordinates in metres. All data are fictional teaching material.
\par\endgroup

% sb_grassgis_009 instruction SHA-256: ba40cc297b6fa4db0ed21e0761569771b3c6dc3a1d7f46a3eb4608bd31b95f54
\subsubsection{\texttt{sb\_grassgis\_009} Map redundant sightlines across a teaching plot}
\label{task:sb-grassgis-009}
\noindent\textit{Applications.} GRASS GIS.\par\nopagebreak[4]
\begingroup
\small
\setlength{\parindent}{0pt}
\setlength{\parskip}{0.3em}
The existing \path{terrain_m} raster and \path{observation_posts} vector describe a fictional visibility trial. Create a native integer raster named \path{sightline_coverage} classifying every terrain cell: 0 = visible from neither post, 1 = from A only, 2 = from B only, and 3 = from both. Use each post's \path{eye_m} attribute as its eye height above local ground, and test a target 0.75 m above local ground at every terrain cell centre. Use GRASS GIS's native r.viewshed terrain model, looking in all directions with no distance limit; ignore Earth curvature and atmospheric refraction. Terrain is the only obstruction. Keep the DEM's complete extent, alignment, and 4 m resolution; all terrain heights are known, so keep class 0 as zero rather than NULL. Report the percentage of the full plot's planar area visible from both posts, to one decimal place. Save \path{sightline_coverage} in the current work mapset and leave it displayed with the observation posts overlaid and the coverage classes distinguishable. The horizontal CRS is EPSG:32631 (WGS 84 / UTM zone 31N), with coordinates in metres; terrain heights are metres above an arbitrary local datum. All inputs are invented teaching data.
\par\endgroup

% sb_kalgebra_001 instruction SHA-256: aa18dcc3d9bf23a5da80d66b33f9a90337b782c63ae989ca923a3bd6c9c796ad
\subsubsection{\texttt{sb\_kalgebra\_001} Minimize coating cost for an open channel}
\label{task:sb-kalgebra-001}
\noindent\textit{Applications.} KAlgebra.\par\nopagebreak[4]
\begingroup
\small
\setlength{\parindent}{0pt}
\setlength{\parskip}{0.3em}
An idealized open-top channel has a rectangular internal cross-section with area 3.6 cm$^2$. Its width w and wall height h are both positive. For each unit length of channel, coating the bottom costs three times as much per unit area as coating either side wall. Ignore wall thickness and end walls. Use KAlgebra to express the coating cost as a function of width alone, normalized by the side-wall coating price per unit area and by channel length. Find the width and wall height that minimize this cost, and report both to the nearest 0.01 cm with a brief explanation of why your choice is a minimum. Keep the numerical dimension calculations in the calculator history. Leave the 2D Graph tab visible with the cost curve and its positive-width minimum clearly in view.
\par\endgroup

% sb_kalgebra_002 instruction SHA-256: 9ddc797fd72b9bf4676721946529df3a4b9a2ce86092211cc5cdd58b17f4c3b5
\subsubsection{\texttt{sb\_kalgebra\_002} Overlay a delayed actuator pulse}
\label{task:sb-kalgebra-002}
\noindent\textit{Applications.} KAlgebra.\par\nopagebreak[4]
\begingroup
\small
\setlength{\parindent}{0pt}
\setlength{\parskip}{0.3em}
Use KAlgebra to plot two command profiles for a toy actuator model. The horizontal coordinate x is time in seconds and the vertical coordinate is command level in arbitrary units. The first profile is zero through x = 1, rises linearly to level 4 at x = 3, stays at level 4 through x = 5, falls linearly to zero at x = 8, and stays zero thereafter. The second profile has exactly the same shape, delayed by 1.5 seconds and with every command level multiplied by 0.7. Represent each complete profile as one piecewise function, including its zero portions. Leave the 2D Graph tab visible with both profiles enabled in different colors and the viewport set to x from -1 to 11 and y from -0.5 to 4.5.
\par\endgroup

% sb_kalgebra_003 instruction SHA-256: aac6b43dec898b2ecbb61ee71288019a5ec9f6f71536bfc232e9c82c0a0aaa16
\subsubsection{\texttt{sb\_kalgebra\_003} Orient a corrugated-sheet surface}
\label{task:sb-kalgebra-003}
\noindent\textit{Applications.} KAlgebra.\par\nopagebreak[4]
\begingroup
\small
\setlength{\parindent}{0pt}
\setlength{\parskip}{0.3em}
Use KAlgebra to model an idealized corrugated sheet as a height surface z = f(x,y), with x, y and z measured in centimetres. The corrugation is sinusoidal about z = 0, with amplitude 0.6 cm. Its straight, parallel crest lines run in the direction (3,4) in the xy plane, and the perpendicular distance between neighboring crest lines is 4 cm. The sheet passes through the origin, and its height initially increases when moving from the origin along positive x. Construct and plot the height function in 3D. Leave the 3D Graph tab visible in Lines style at an oblique angle that clearly shows the parallel crests and troughs and their height variation. State the height function you used.
\par\endgroup

% sb_kalgebra_004 instruction SHA-256: 892334325484232c0781f8717e75dc61ecc1a6a5dd8b11dc12b5cd5e79c59779
\subsubsection{\texttt{sb\_kalgebra\_004} Recover inputs from an ambiguous calibration}
\label{task:sb-kalgebra-004}
\noindent\textit{Applications.} KAlgebra.\par\nopagebreak[4]
\begingroup
\small
\setlength{\parindent}{0pt}
\setlength{\parskip}{0.3em}
A toy instrument uses the calibration r = ((x-1)*(x-5))\textasciicircum{}(1/2), where x is restricted to the interval [0,6] and the exponent denotes the nonnegative real square root. Use KAlgebra to find every admissible input x that produces the reading r = 2. Report the inputs to three decimal places, identify exactly which parts of [0,6] yield a real calibration output, and state whether this reading determines x uniquely. Keep numerical substitution checks for every returned input, using unrounded values, in the Calculator history. Leave the 2D Graph tab visible with the calibration curve plotted in a viewport of x from 0 to 6 and y from 0 to 3, so the real-domain branches can be inspected. Both x and r are dimensionless.
\par\endgroup

% sb_kalgebra_005 instruction SHA-256: b57169ac8d8955b5e26b358ed82fee65d90d3e2d0d85c72b470ae270c09de725
\subsubsection{\texttt{sb\_kalgebra\_005} Separate travel along a guide from transverse drift}
\label{task:sb-kalgebra-005}
\noindent\textit{Applications.} KAlgebra.\par\nopagebreak[4]
\begingroup
\small
\setlength{\parindent}{0pt}
\setlength{\parskip}{0.3em}
An invented motion-tracking sample has instantaneous velocity v = (4, -1, 3) mm/s in fixed orthogonal x, y, z axes. Its intended travel line has direction d = (2, 3, 1); d is dimensionless and is not a unit vector. Use KAlgebra to decompose v into a velocity vector parallel to this line and a perpendicular drift velocity vector whose sum is v. Report both vectors in x, y, z order and the magnitude of the drift velocity, all to three decimal places in mm/s. Keep the calculations in the Calculator history, including checks with unrounded values that the two vectors reconstruct v and that the drift vector has zero scalar product with d. Vector notation or component-by-component arithmetic is acceptable. Leave the Calculator tab visible with the decomposition and checks available in its history.
\par\endgroup

% sb_kalgebra_006 instruction SHA-256: f73f1613989faf9271848ed5a6c66924f47eda8036696820d406b9e56a148406
\subsubsection{\texttt{sb\_kalgebra\_006} Find the first successful dilution cycle}
\label{task:sb-kalgebra-006}
\noindent\textit{Applications.} KAlgebra.\par\nopagebreak[4]
\begingroup
\small
\setlength{\parindent}{0pt}
\setlength{\parskip}{0.3em}
In an invented mixing experiment, a well-mixed vessel initially holds 12 L of water containing 18 mg of a conserved tracer. Each cycle drains 3 L, refills with tracer-free water to 12 L, and then adds 1.2 mg of tracer in negligible volume, mixing fully before the next cycle. No other tracer loss occurs. Use KAlgebra to find the smallest whole number of completed cycles after which the tracer concentration is strictly below 0.55 mg/L. Cycle 0 means the initial vessel; evaluate concentration only after complete cycles, including the added dose. State the recurrence or equivalent expression you used, and report the cycle number and the concentrations after that cycle and the immediately preceding cycle. Give concentrations to three decimal places, but use unrounded values to decide the strict inequality. Keep the calculations and both threshold comparisons in the Calculator history, and leave the Calculator tab visible.
\par\endgroup

% sb_kalgebra_007 instruction SHA-256: deee88435da3acbb9bfce3a93746d08a1dbcc7e481538aef117e3437613f999e
\subsubsection{\texttt{sb\_kalgebra\_007} Account for recording bias in bead energy}
\label{task:sb-kalgebra-007}
\noindent\textit{Applications.} KAlgebra.\par\nopagebreak[4]
\begingroup
\small
\setlength{\parindent}{0pt}
\setlength{\parskip}{0.3em}
An invented bead stream has four possible speeds: 0.5, 1.0, 1.5, and 2.0 m/s. Their incoming population proportions are given by the relative weights 7, 9, 5, and 3, respectively. Every bead has mass 4.0 g. A detector records an incoming bead with probability 0.10, 0.35, 0.80, or 0.95, respectively, conditional on its speed; recording does not change the bead's speed. Treat all supplied values as exact for this toy model. Use KAlgebra to calculate the expected translational kinetic energy of a randomly selected recorded bead. Use K = m*v\textasciicircum{}2/2 with consistent units. Report the expected energy in millijoules to three significant figures and briefly explain how you weighted the four speed classes after recording. Keep the weighting and energy calculations in the Calculator history, and leave the Calculator tab visible.
\par\endgroup

% sb_kalgebra_008 instruction SHA-256: c68ec6ba254427fd25162d8a996ed87c9073b922983e17d5f8ab8f231b312805
\subsubsection{\texttt{sb\_kalgebra\_008} Construct a sideways reflector from its focus}
\label{task:sb-kalgebra-008}
\noindent\textit{Applications.} KAlgebra.\par\nopagebreak[4]
\begingroup
\small
\setlength{\parindent}{0pt}
\setlength{\parskip}{0.3em}
In a toy two-dimensional optical geometry, a reflector cross-section consists of all points whose Euclidean distance from the focus F = (-2, 1) cm equals their perpendicular distance from the vertical directrix x = 4 cm. Use KAlgebra to construct this locus as one native 2D graph giving x as a function of y, with x horizontal and y vertical. Include both the upper and lower parts of the reflector. Overlay the directrix as a second graph in a different color. Report the vertex coordinates and the vertex-to-focus distance in centimetres, and briefly explain how the equal-distance condition determines which way the reflector opens. Leave the 2D Graph tab visible with the viewport set to x from -8 to 5 and y from -9 to 11.
\par\endgroup

% sb_kalgebra_009 instruction SHA-256: 2bbd751f9495a94ab4f5616d6310b7faf2949c9b66ceb61e6c2c474118676f77
\subsubsection{\texttt{sb\_kalgebra\_009} Fit a colorimeter calibration with a free baseline}
\label{task:sb-kalgebra-009}
\noindent\textit{Applications.} KAlgebra.\par\nopagebreak[4]
\begingroup
\small
\setlength{\parindent}{0pt}
\setlength{\parskip}{0.3em}
Use KAlgebra to calibrate an invented colorimeter from five teaching measurements. The standard concentrations c are 0, 2, 4, 6, and 8 mmol/L; their corresponding dimensionless absorbances A are 0.035, 0.210, 0.401, 0.562, and 0.758. Treat the standard concentrations as exact. Fit the affine model A = a + b*c by ordinary least squares, minimizing the sum of squared absorbance residuals with equal weight for all five standards. Fit the intercept a as well as the slope b. Then use this fitted model to estimate the concentration of an unknown sample whose absorbance is 0.487. Report the fitted intercept and slope to four decimal places, with units where applicable, and the unknown concentration to three significant figures in mmol/L. Use unrounded fitted coefficients for the concentration calculation. Keep the fitting and concentration calculations in the Calculator history, and leave the Calculator tab visible.
\par\endgroup

% sb_kalgebra_010 instruction SHA-256: 074e4933880961d20ec2b11e882d3aa5f8a7ae5e98a19d8da0659fe40f16c751
\subsubsection{\texttt{sb\_kalgebra\_010} Compare interior and transition errors in harmonic reconstruction}
\label{task:sb-kalgebra-010}
\noindent\textit{Applications.} KAlgebra.\par\nopagebreak[4]
\begingroup
\small
\setlength{\parindent}{0pt}
\setlength{\parskip}{0.3em}
Use KAlgebra to investigate a toy square-wave reconstruction. The ideal voltage has period T = 2.4 s: it is +1.5 V for 0 $<$ t $<$ T/2 and -1.5 V for T/2 $<$ t $<$ T, repeated periodically. An N-term reconstruction adds the first N odd sine harmonics. For k = 1,...,N, harmonic k has frequency (2*k-1)/T Hz, amplitude 4*1.5/(pi*(2*k-1)) V, and zero phase. Construct the reconstructions for N = 4 and N = 12. For each reconstruction, calculate the signed percentage error relative to +1.5 V at two locations: the fixed interior time t = T/4, and the near-transition time t = T/(4*N), using that reconstruction's own N. Define this error as 100*(reconstructed voltage - 1.5)/1.5. Report all four signed errors to one decimal place, identifying the reconstruction and sampling location, and briefly compare how the interior and near-transition errors change when more harmonics are included. Use unrounded intermediate values and keep the reconstruction evaluations and error calculations in the Calculator history. Leave the 2D Graph tab visible with both reconstructions and the +1.5 V reference line enabled in three distinct colors. Use the horizontal coordinate x for time in seconds and set the viewport to x from 0 to 1.2 and y from -0.2 to 2.0 V.
\par\endgroup

% sb_lean_001 instruction SHA-256: 68c97a9f3a3537befc1e8575dec85310a72abc5c09cfb7c2d00e0b536cf0b87c
\subsubsection{\texttt{sb\_lean\_001} Prove a weighted energy bound and characterize equality}
\label{task:sb-lean-001}
\noindent\textit{Applications.} Lean.\par\nopagebreak[4]
\begingroup
\small
\setlength{\parindent}{0pt}
\setlength{\parskip}{0.3em}
Complete the theorem \path{ScienceBoardWeightedEnergy.weighted_energy_bound} in \path{/home/user/sci/Sci/sb_lean_001.lean}. For arbitrary real readings x and y and strictly positive weights a and b, prove that (a*x + b*y)\textasciicircum{}2 is at most (a+b)*(a*x\textasciicircum{}2 + b*y\textasciicircum{}2), and that equality holds if and only if x = y. Preserve the theorem name, parameters, hypotheses, and conclusion. You may add proved helper lemmas and use Mathlib tactics. Replace the unfinished proof with a proof containing no sorry, admit, or new axioms. Save the file and leave it open in VS Code with Lean reporting no errors or unfinished-proof warnings.
\par\endgroup

% sb_lean_002 instruction SHA-256: 1527f66c297b54f2cdddd6e24948ea1527a4277aa793d0e3fc25f29e7256d395
\subsubsection{\texttt{sb\_lean\_002} Prove rollback for shifts and reflections}
\label{task:sb-lean-002}
\noindent\textit{Applications.} Lean.\par\nopagebreak[4]
\begingroup
\small
\setlength{\parindent}{0pt}
\setlength{\parskip}{0.3em}
Complete \path{ScienceBoardRollback.run_rewind} in \path{/home/user/sci/Sci/sb_lean_002.lean}. The supplied command language moves an integer position using shifts and reflections through zero. Prove that executing any program and then its rewind restores the starting position. Keep the supplied definitions and theorem statement unchanged. You may add proved helper lemmas and use Mathlib tactics. Replace the unfinished proof with a proof containing no sorry, admit, or new axioms. Save the file and leave it open in VS Code with Lean reporting no errors or unfinished-proof warnings.
\par\endgroup

% sb_lean_003 instruction SHA-256: 1b01eb87889303152f3a0b9564a6ccae4cc3e5da9f9346e8c6fb8ab1264a0c7b
\subsubsection{\texttt{sb\_lean\_003} Characterize the outputs of two compatible projections}
\label{task:sb-lean-003}
\noindent\textit{Applications.} Lean.\par\nopagebreak[4]
\begingroup
\small
\setlength{\parindent}{0pt}
\setlength{\parskip}{0.3em}
Complete \path{ScienceBoardJointStability.range_settle} in \path{/home/user/sci/Sci/sb_lean_003.lean}. The functions p and q act on an arbitrary type, each is idempotent, and they commute. The supplied settle function applies q and then p. Prove that the set of all outputs of settle is exactly the set of states fixed by both p and q. Keep the supplied definition and theorem statement unchanged. You may add proved helper lemmas and use Mathlib tactics. Replace the unfinished proof with a proof containing no sorry, admit, or new axioms. Save the file and leave it open in VS Code with Lean reporting no errors or unfinished-proof warnings.
\par\endgroup

% sb_lean_004 instruction SHA-256: 8c1723a9524f4e27f3861ad1d546b993027f4934ba4df3bdff3b35809c9bf3f3
\subsubsection{\texttt{sb\_lean\_004} Show that count mixing preserves the greatest common divisor}
\label{task:sb-lean-004}
\noindent\textit{Applications.} Lean.\par\nopagebreak[4]
\begingroup
\small
\setlength{\parindent}{0pt}
\setlength{\parskip}{0.3em}
Complete \path{ScienceBoardCounterMix.gcd_mixed_counts} in \path{/home/user/sci/Sci/sb_lean_004.lean}. A toy counter system mixes two natural-number counts a and b into 2*a + 3*b and 3*a + 5*b. Prove that the greatest common divisor of the mixed counts is exactly the greatest common divisor of the original counts, for every a and b, including zero. Keep the theorem name, parameters, and conclusion unchanged. You may add proved helper lemmas and use Mathlib tactics. Replace the unfinished proof with a proof containing no sorry, admit, or new axioms. Save the file and leave it open in VS Code with Lean reporting no errors or unfinished-proof warnings.
\par\endgroup

% sb_lean_005 instruction SHA-256: f01291423ef83a3335978d43ebaf8ec2b3de320b5da503f28610d659d430c6ac
\subsubsection{\texttt{sb\_lean\_005} Account for route costs when every outlet is expanded}
\label{task:sb-lean-005}
\noindent\textit{Applications.} Lean.\par\nopagebreak[4]
\begingroup
\small
\setlength{\parindent}{0pt}
\setlength{\parskip}{0.3em}
Complete \path{ScienceBoardRouteExpansion.totalRouteCost_replaceOutlets} in \path{/home/user/sci/Sci/sb_lean_005.lean}. The supplied toy routing tree has terminal outlets, links with natural-number costs, and zero-cost forks. totalRouteCost sums the link costs along every root-to-outlet path, counting shared links once for each outlet reached. replaceOutlets replaces each original outlet with one copy of the same replacement tree, without expanding the inserted outlets again. Prove the stated formula for the resulting total route cost for all tree shapes and all link costs, including zero. Keep the supplied datatype, definitions, and theorem statement unchanged. You may add proved helper lemmas and use Mathlib tactics. Replace the unfinished proof with a proof containing no sorry, admit, or new axioms. Save the file and leave it open in VS Code with Lean reporting no errors or unfinished-proof warnings.
\par\endgroup

% sb_lean_006 instruction SHA-256: 9130741625fbede3948c43a6b973bfbbe879471407aaea644958915eabd71d56
\subsubsection{\texttt{sb\_lean\_006} Split an integer scan without skipped or repeated ticks}
\label{task:sb-lean-006}
\noindent\textit{Applications.} Lean.\par\nopagebreak[4]
\begingroup
\small
\setlength{\parindent}{0pt}
\setlength{\parskip}{0.3em}
Complete \path{ScienceBoardScanWindows.exact_scan_split_iff} in \path{/home/user/sci/Sci/sb_lean_006.lean}. A toy scanner visits every integer tick in a closed window, including both endpoints. The first scan runs from lo to leftEnd; the second runs from rightStart to hi. Both inner endpoints lie inside the target window from lo to hi. Prove that the two scans together cover the target with no tick repeated if and only if rightStart = leftEnd + 1. The statement expresses complete coverage by a union equality and absence of repeats by an empty intersection. It must hold for arbitrary integer endpoints, including negative ones. Keep the supplied definition, theorem name, parameters, hypotheses, and conclusion unchanged. You may add proved helper lemmas and use Mathlib tactics. Replace the unfinished proof with a proof containing no sorry, admit, or new axioms. Save the file and leave it open in VS Code with Lean reporting no errors or unfinished-proof warnings.
\par\endgroup

% sb_lean_007 instruction SHA-256: 8049fec09b7e2d7caef8c0fe1cbacf6074845434ddbba79603374504386e6ef6
\subsubsection{\texttt{sb\_lean\_007} Bound the accumulated error of a disturbed iteration}
\label{task:sb-lean-007}
\noindent\textit{Applications.} Lean.\par\nopagebreak[4]
\begingroup
\small
\setlength{\parindent}{0pt}
\setlength{\parskip}{0.3em}
Complete \path{ScienceBoardDisturbedIteration.disturbed_iterate_bound} in \path{/home/user/sci/Sci/sb_lean_007.lean}. The toy scalar process starts at x0 and applies f, then adds the signed disturbance noise n in the transition from step n to step n + 1. The ideal update contracts distances by a factor q with 0 $\leq$ q $<$ 1 and fixes p; every disturbance has magnitude at most $\varepsilon$. Prove the stated finite-step error bound around p for every natural-number step, including step zero. Disturbances may change magnitude and sign at every step, and q = 0 and $\varepsilon$ = 0 are allowed. Keep the supplied definition, theorem name, parameters, hypotheses, and conclusion unchanged. You may add proved helper lemmas and use Mathlib tactics. Replace the unfinished proof with a proof containing no sorry, admit, or new axioms. Save the file and leave it open in VS Code with Lean reporting no errors or unfinished-proof warnings.
\par\endgroup

% sb_texstudio_001 instruction SHA-256: e3f6bfdacbafcee79261f0ffa6d7fa5b5d68f89ee93b2e08eec21af5cfb17038
\subsubsection{\texttt{sb\_texstudio\_001} Two coupled carts: calculate and typeset the normal modes}
\label{task:sb-texstudio-001}
\noindent\textit{Applications.} TeXstudio, KAlgebra.\par\nopagebreak[4]
\begingroup
\small
\setlength{\parindent}{0pt}
\setlength{\parskip}{0.3em}
Complete the short normal-mode note in \path{/home/user/ScienceBoardSynthetic/sb_texstudio_001/mode_note.tex} for the educational two-cart model described there. Use KAlgebra for the numerical work; \path{cart_parameters.kal} in the same folder contains only the raw parameter assignments and can be loaded through Calculator $>$ Load Script. From the supplied equations of motion, derive the characteristic equation in z = omega squared, then calculate the two positive natural frequencies f in Hz and the signed amplitude ratio A2/A1 for each mode. Use the second equation of motion to check both modes with unrounded values, keeping the calculations and checks in KAlgebra's calculator history. Replace the note's pending text and table cells with the characteristic equation, a brief explanation of the conversion from z to f, and the results ordered by increasing frequency. Report frequencies to three significant figures and ratios to three decimal places; explain what the ratio signs imply about the carts' relative motion and briefly state the consistency-check outcome. Save the completed source and its compiled PDF as \path{mode_note.tex} and \path{mode_note.pdf} in the same folder. Build the note in TeXstudio without compilation errors and leave the PDF visible, with KAlgebra still open.
\par\endgroup

% sb_texstudio_002 instruction SHA-256: 63a034356bd340f64a5b35ae945516995ab2a5f9d262c469847d2a3611009d19
\subsubsection{\texttt{sb\_texstudio\_002} Show the torsion of a toy four-carbon chain}
\label{task:sb-texstudio-002}
\noindent\textit{Applications.} TeXstudio, ChimeraX.\par\nopagebreak[4]
\begingroup
\small
\setlength{\parindent}{0pt}
\setlength{\parskip}{0.3em}
Complete the single figure and caption in \path{/home/user/ScienceBoardSynthetic/sb_texstudio_002/torsion_note.tex} using the original toy four-carbon chain already open in ChimeraX from \path{four_carbon.pdb}. The model is educational geometry with hydrogens omitted, not an experimental structure. In ChimeraX, measure the signed C1-C2-C3-C4 torsion in that atom order without changing the coordinates. Prepare one ball-and-stick view with all four atom names readable and all three bonds visible; choose an oblique view that makes the chain's three-dimensional arrangement clear. Export the graphics view as \path{chain_view.png} in the same folder and replace the TeX figure placeholder with that image, preserving its aspect ratio. Replace the pending caption with a concise description that identifies the toy model, reports the measured signed torsion to the nearest degree, and says which ideal reference it is closest to: eclipsed (0 degrees), gauche (+60 or -60 degrees), or anti (180 degrees, equivalent to -180 degrees). Compare angles cyclically. Save \path{torsion_note.tex} and build \path{torsion_note.pdf} in TeXstudio without compilation errors. Keep the note to one page and leave the PDF visible with the image and caption readable.
\par\endgroup

% sb_texstudio_003 instruction SHA-256: f6fcb55aee5de9a0ec40fdaa8900fdec37a6bac47c22cbb5b1181fe5d183033e
\subsubsection{\texttt{sb\_texstudio\_003} Report cylinder density with shared calibration uncertainty}
\label{task:sb-texstudio-003}
\noindent\textit{Applications.} TeXstudio, KAlgebra.\par\nopagebreak[4]
\begingroup
\small
\setlength{\parindent}{0pt}
\setlength{\parskip}{0.3em}
Complete the one-page measurement note in \path{/home/user/ScienceBoardSynthetic/sb_texstudio_003/density_note.tex} using its invented cylinder measurements and stated uncertainty assumptions. Preserve the raw data and assumptions. Replace the pending content with a coherent density model, the uncertainty expression for each mean length, and the first-order relative standard-uncertainty formula for density. Fill the summary table with each dimension's mean, sample standard deviation, and combined standard uncertainty of its mean, all in millimetres. Give means to three decimal places and both standard deviations and standard uncertainties to two significant figures. Use unrounded intermediate values for propagation. Report density as a value plus or minus its standard uncertainty in kg/m\textasciicircum{}3, with uncertainty to two significant figures and density rounded to the same decimal place; label this as a standard uncertainty (k = 1). State the unit conversion, identify which of mass, mean diameter, or mean height has the largest propagated uncertainty contribution, and explain briefly why averaging does not reduce the calibration terms. KAlgebra is optional for numerical checks: \path{cylinder_readings.kal} contains only the raw assignments (lengths and b values in mm; mass and um in g) and can be loaded with Calculator $>$ Load Script. The task may be completed without KAlgebra; no calculator history or export is required. Save \path{density_note.tex} and build \path{density_note.pdf} in the same folder using TeXstudio without compilation errors. Keep the completed note to one readable page and leave its PDF visible.
\par\endgroup

% sb_texstudio_004 instruction SHA-256: 63be180f3f28333e616924317ac96ee8c67b989362e2bc0a6f11fe04b9e5d2e7
\subsubsection{\texttt{sb\_texstudio\_004} Make an exchange note survive renumbering}
\label{task:sb-texstudio-004}
\noindent\textit{Applications.} TeXstudio.\par\nopagebreak[4]
\begingroup
\small
\setlength{\parindent}{0pt}
\setlength{\parskip}{0.3em}
Edit \path{/home/user/ScienceBoardSynthetic/sb_texstudio_004/exchange_note.tex} in TeXstudio. Reorganize the note into this section order: Model assumptions, Balance equations, Conservation check. Typeset the two compartment balances together as a two-line display aligned at their equals signs, with one shared automatic equation number. Keep the total-tracer conservation equation separately numbered. Replace every manually typed section and equation number in the prose with a label-based LaTeX cross-reference to its intended target; references to the pair of balances should use their shared number. Use ordinary automatic numbering, without fixed tags or counter overrides. Preserve the scientific content, equations, section titles, and variable definitions, changing reference wording where needed. Save \path{exchange_note.tex} and build \path{exchange_note.pdf} in the same folder, with no compilation errors or unresolved references. Leave the compiled PDF visible in TeXstudio.
\par\endgroup

% sb_texstudio_005 instruction SHA-256: 177da423f4453518e92800855b65d9c29acf32946e508fe0ba9bcd1d8ff06a08
\subsubsection{\texttt{sb\_texstudio\_005} Repair the citations in a thermometer lesson}
\label{task:sb-texstudio-005}
\noindent\textit{Applications.} TeXstudio.\par\nopagebreak[4]
\begingroup
\small
\setlength{\parindent}{0pt}
\setlength{\parskip}{0.3em}
Repair the citations and bibliography in \path{/home/user/ScienceBoardSynthetic/sb_texstudio_005/thermometer_note.tex} using TeXstudio and the supplied \path{teaching_sources.bib} in the same folder. All references are fictional teaching materials. Replace the four manually typed bracketed citations with key-based LaTeX citations to the works named in their surrounding sentences. Replace the draft reference list with a bibliography generated by BibTeX from \path{teaching_sources.bib}. Use numeric citations and number the bibliography in order of first citation; repeated citations to the same work must reuse its number. Include only works cited in the lesson, while retaining all records and metadata in the .bib file. Preserve the lesson text, equation, and fictional-materials notice. Save \path{thermometer_note.tex} and build \path{thermometer_note.pdf} in the same folder without compilation errors or unresolved citations. Leave the compiled PDF visible in TeXstudio.
\par\endgroup

\clearpage

\end{document}